\documentclass[10pt,twocolumn,letterpaper]{article}
\usepackage{titling}
\usepackage[pagenumbers]{cvpr} %

\usepackage[accsupp]{axessibility}
\usepackage{amsmath}
\usepackage{amssymb}
\usepackage{booktabs}
\usepackage{bbding}
\usepackage{graphicx}
\usepackage{animate}
\usepackage{grffile}
\usepackage{siunitx} %
\usepackage{enumitem}
\usepackage[table, usenames, dvipsnames]{xcolor}

\definecolor{mycyan}{rgb}{0,0.76,0.75}

\usepackage[pagebackref,breaklinks,colorlinks]{hyperref}

\usepackage[capitalize]{cleveref}
\crefname{section}{Sec.}{Secs.}
\Crefname{section}{Section}{Sections}
\Crefname{table}{Table}{Tables}
\crefname{table}{Tab.}{Tabs.}

\usepackage{times}
\usepackage{epsfig}
\usepackage{amsmath}
\usepackage{amssymb}
\usepackage{booktabs}
\usepackage{makecell}
\usepackage{multirow}
\usepackage{tabularx}
\usepackage{bbm}
\usepackage{caption} 
\usepackage{subcaption}
\usepackage{animate}
\usepackage{pifont}     %

\newcommand{\bc}{\mathbf{c}}

\newcommand{\bg}{\mathbf{g}}

\newcommand{\bo}{\mathbf{o}}
\newcommand{\bp}{\mathbf{p}}

\newcommand{\br}{\mathbf{r}}\newcommand{\bR}{\mathbf{R}}

\newcommand{\bt}{\mathbf{t}}

\newcommand{\cL}{\mathcal{L}}

\newcommand{\figref}[1]{Fig.~\ref{#1}}
\newcommand{\secref}[1]{Section~\ref{#1}}

\newcommand{\eqnref}[1]{Eq.~\eqref{#1}}
\newcommand{\tabref}[1]{Table~\ref{#1}}

\makeatletter
\DeclareRobustCommand\onedot{\futurelet\@let@token\@onedot}
\def\@onedot{\ifx\@let@token.\else.\null\fi\xspace}
\def\eg{e.g\onedot} 
\def\ie{i.e\onedot} 
 
\def\etc{etc\onedot}

\def\etal{et~al\onedot}

\makeatother

\newcommand{\boldparagraph}[1]{\vspace{0.1em}\noindent{\bf #1} }

\definecolor{darkgreen}{rgb}{0,0.7,0}

\newcommand{\ours}{NICE-SLAM}

\newcommand{\red}[1]{\noindent{\color{red}{#1}}}

\newcolumntype{P}[1]{>{\centering\arraybackslash}m{#1}}

\def\@fnsymbol#1{\ensuremath{\ifcase#1\or *\or \dagger\or \ddagger\or
   \mathsection\or \mathparagraph\or \|\or **\or \dagger\dagger
   \or \ddagger\ddagger \else\@ctrerr\fi}}
\newcommand{\ssymbol}[1]{^{\@fnsymbol{#1}}}

\begin{document}

\title{NICE-SLAM: Neural Implicit Scalable Encoding for SLAM}

\author{
Zihan Zhu$^{1,2 *}$ \qquad Songyou Peng$^{2,4}$\thanks{Equal contribution.}\qquad Viktor Larsson$^{3}$ \qquad Weiwei Xu$^{1}$ \qquad Hujun Bao$^{1}$\\
Zhaopeng Cui$^{1}$\thanks{Corresponding author.} \qquad Martin R. Oswald$^{2,5}$ \qquad Marc Pollefeys$^{2,6}$\vspace{0.5em}\\
$^{1}$State Key Lab of CAD\&CG, Zhejiang University\qquad $^{2}$ETH Zurich\qquad
$^{3}$Lund University\\$^{4}$MPI for Intelligent Systems, T\"ubingen \qquad
$^{5}$University of Amsterdam\qquad $^{6}$Microsoft
}

\maketitle

\begin{abstract}

Neural implicit representations have recently shown encouraging results in various domains, including promising progress in simultaneous localization and mapping (SLAM). Nevertheless, existing methods produce over-smoothed scene reconstructions and have difficulty scaling up to large scenes. 
These limitations are mainly due to their simple fully-connected network architecture that does not incorporate local information in the observations.
In this paper, we present NICE-SLAM, a dense SLAM system that incorporates multi-level local information by introducing a hierarchical scene representation. Optimizing this representation with pre-trained geometric priors enables detailed reconstruction on large indoor scenes. Compared to recent neural implicit SLAM systems, our approach is more scalable, efficient, and robust. Experiments on five challenging datasets demonstrate competitive results of NICE-SLAM in both mapping and tracking quality. Project page: \url{https://pengsongyou.github.io/nice-slam}.

\end{abstract}

\section{Introduction}  \label{sec:intro}
Dense visual Simultaneous Localization and Mapping (SLAM) is a fundamental problem in 3D computer vision with many applications in autonomous driving, indoor robotics, mixed reality, \etc. 
In order to make a SLAM system truly useful for real-world applications, the following properties are essential.
First, we desire the SLAM system to be real-time.
Next, the system should have the ability to make reasonable predictions for regions without observations.
Moreover, the system should be able to scale up to large scenes.
Last but not least, it is crucial to be robust to noisy or missing observations.

In the scope of real-time dense visual SLAM system, many methods have been introduced for RGB-D cameras in the past years. 
Traditional dense visual SLAM systems~\cite{Schoeps2019CVPR,whelan2015elasticfusion,newcombe2011dtam,Whelan:etal:RSSRGBD2012} fulfil the real-time requirement and can be used in large-scale scenes, but they are unable to make plausible geometry estimation for unobserved regions.
On the other hand, learning-based SLAM approaches~\cite{bloesch2018codeslam,zhi2019scenecode,czarnowski2020deepfactors,Sucar20203DV} attain a certain level of predictive power since they typically train on task-specific datasets. 
Moreover, learning-based methods tend to better deal with noises and outliers.
However, these methods are typically only working in small scenes with multiple objects.
Recently, Sucar~\etal~\cite{imap} applied a neural implicit representation in the real-time dense SLAM system (called iMAP), and they showed decent tracking and mapping results for room-sized datasets.
Nevertheless, when scaling up to larger scenes, e.g., an apartment consisting of multiple rooms, significant performance drops are observed in both the dense reconstruction and camera tracking accuracy.

\begin{figure}[!t]
  \centering
  \small
  \setlength{\tabcolsep}{0.3em}
  \begin{tabular}
  {
  >{\centering\arraybackslash}m{0.46\textwidth}}
    \includegraphics[width=\linewidth]{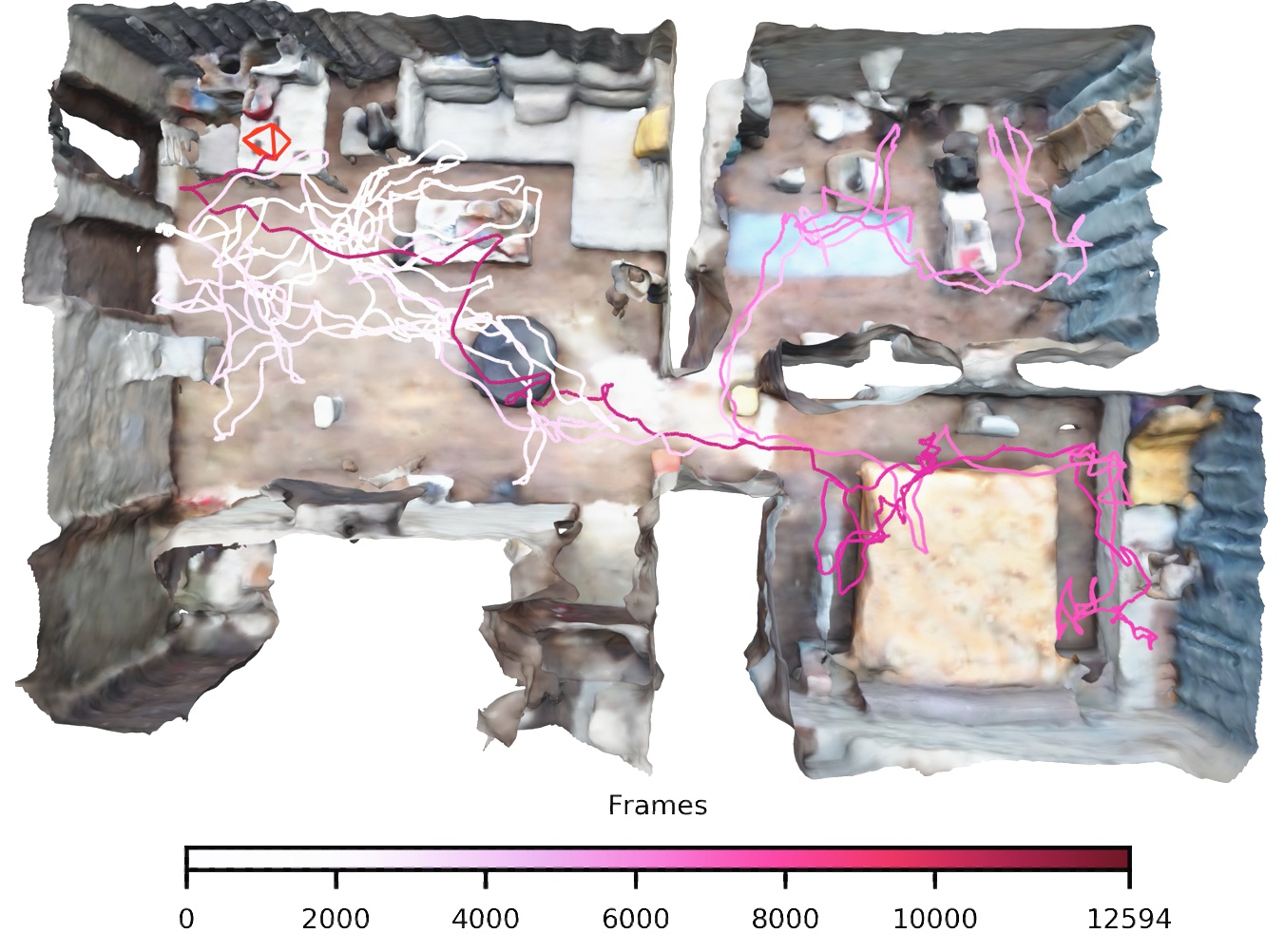} \\[-4pt]
  \end{tabular} 
  \caption{\textbf{Multi-room Apartment 3D Reconstruction using \ours{}}. A hierarchical feature grid jointly encodes geometry and color information and is used for both mapping and tracking. We depict the final mesh and camera tracking trajectory.}
  \label{fig:teaser}
\end{figure}

The key limiting factor of iMAP~\cite{imap} stems from its use of a single multi-layer perceptron (MLP) to represent the entire scene, which can only be updated globally with every new, potentially partial RGB-D observations.
In contrast, recent works~\cite{Peng2020ECCV,Sun2021CVPR} demonstrate that establishing multi-level grid-based features can help to preserve geometric details and enable reconstructing complex scenes, but these are offline methods without real-time capability.

In this work, we seek to combine the strengths of hierarchical scene representations with those of neural implicit representations for the task of dense {RGB-D} SLAM.
To this end, we introduce \ours{}, a dense {RGB-D} SLAM system that can be applied to large-scale scenes while preserving the predictive ability.
Our key idea is to represent the scene geometry and appearance with hierarchical feature grids and incorporate the inductive biases of neural implicit decoders pretrained at different spatial resolutions.
With the rendered depth and color images from the occupancy and color decoder outputs, we can optimize the features grids only within the viewing frustum by minimizing the re-rendering losses.
We perform extensive evaluations on a wide variety of indoor {RGB-D} sequences and demonstrate the scalability and predictive ability of our method.
Overall, we make the following contributions:
\begin{itemize}[itemsep=0.1pt,topsep=3pt,leftmargin=*]
    \item We present NICE-SLAM, a dense {RGB-D} SLAM system that is real-time capable, scalable, predictive, and robust to various challenging scenarios. 
    \item The core of NICE-SLAM is a hierarchical, grid-based neural implicit encoding. In contrast to global neural scene encodings, this representation allows for local updates, which is a prerequisite for large-scale approaches. %
    
    \item We conduct extensive evaluations on various datasets which demonstrate competitive performance in both mapping and tracking. 
\end{itemize}
The code is available at~\url{https://github.com/cvg/nice-slam}.

\section{Related Work}  \label{sec:related}
\boldparagraph{Dense Visual SLAM.}
Most modern methods for visual SLAM follow the overall architecture introduced in the seminal work by Klein et al.~\cite{klein2009parallel}, decomposing the task into mapping and tracking.
The map representations can be generally divided into two categories: view-centric and world-centric. 
The first anchors 3D geometry to specific keyframes, often represented as depth maps in the dense setting.
One of the early examples of this category was DTAM~\cite{newcombe2011dtam}. 
Because of its simplicity, DTAM has been widely adapted in many recent learning-based SLAM systems. 
For example, \cite{ummenhofer2017demon,zhou2018deeptam} regress both depth and pose updates. 
DeepV2D~\cite{teed2019deepv2d} similarly alternates between regressing depth and pose estimation but uses test-time optimization. 
BA-Net~\cite{tang2018ba} and DeepFactors~\cite{czarnowski2020deepfactors} simplify the optimization problem by using a set of basis depth maps.
There are also some methods, \eg,~CodeSLAM~\cite{bloesch2018codeslam}, SceneCode~\cite{zhi2019scenecode} and NodeSLAM~\cite{Sucar20203DV}, which optimize a latent representation that decodes into the keyframe or object depth maps.
DROID-SLAM~\cite{teed2021droid} uses regressed optical flow to define geometrical residuals for its refinement.
TANDEM~\cite{Koestler2021CORL} combines multi-view stereo with DSO~\cite{Engel2017direct} for a real-time dense SLAM system.
On the other hand, the world-centric map representation anchors the 3D geometry in uniform world coordinates, and can be further divided into surfels~\cite{whelan2015elasticfusion,schops2019bad} and voxel grids, typically storing occupancies or TSDF values~\cite{curless1996volumetric}.
Voxel grids have been used extensively in RGB-D SLAM, \eg, KinectFusion\cite{newcombe2011kinectfusion} among other works~\cite{bylow2013real,niessner2013real,InfiniTAM_ECCV_2016,dai2017bundlefusion}.
In our proposed pipeline we also adopt the voxel-grid representation. 
In contrast to previous SLAM approaches, we store implicit latent codes of the geometry and directly optimize them during mapping. 
This richer representation allows us to achieve more accurate geometry at lower grid resolutions. 

\boldparagraph{Neural Implicit Representations.}
Recently, neural implicit representations demonstrated promising results for object geometry representation~\cite{mescheder2019occupancy,Park2019CVPR,Chen2019CVPR,Xu2019NIPS,niemeyer2020differentiable,Yariv2020NEURIPS,oechsle2021unisurf,Wang2021NEURIPS,Yariv2021NEURIPS,Saito2019ICCV,Liu2020CVPR,Peng2021NEURIPS}, scene completion~\cite{Peng2020ECCV,Jiang2020CVPR,Chabra2020ECCV}, novel view synthesis~\cite{mildenhall2020nerf,Zhang2020ARXIV,Martin2021CVPR,Reiser2021ICCV} and also generative modelling~\cite{Schwarz2020NEURIPS,Chan2021CVPR, Niemeyer2021ARXIV,Niemeyer2021CVPR}. 
A few recent papers~\cite{Murez2020ECCV,Sun2021CVPR,Choe2021ICCV,Bovzivc2021ARXIV,Azinovic2021ARXIV,yan2021continual,Weder2021CVPR} attempt to predict scene-level geometry with RGB-(D) inputs, but they all assume given camera poses.
Another set of works~\cite{Wang2021ARXIV,Yen2020ARXIV,Lin2021ARXIV} tackle the problem of camera pose optimization, but they need a rather long optimization process, which is not suitable for real-time applications.

The most related work to our method is iMAP~\cite{imap}.
Given an RGB-D sequence, they introduce a real-time dense SLAM system that uses a single multi-layer perceptron (MLP) to compactly represent the entire scene. 
Nevertheless, due to the limited model capacity of a single MLP, iMAP fails to produce detailed scene geometry and accurate camera tracking, especially for larger scenes.
In contrast, we provide a scalable solution akin to iMAP, that combines learnable latent embeddings with a pretrained continuous implicit decoder.
In this way, our method can reconstruct complex geometry and predict detailed textures for larger indoor scenes, 
while maintaining much less computation and faster convergence.
Notably, the works~\cite{Jiang2020CVPR,Peng2020ECCV} also combine traditional grid structures with learned feature representations for scalability, but neither of them is real-time capable.
Moreover, DI-Fusion~\cite{huang2021di} also optimizes a feature grid given an RGB-D sequence, but their reconstruction often contain holes and their camera tracking is not robust for the pure surface rendering loss.

\section{Method}  \label{sec:method}

\begin{figure*}[htbp]
  \centering
  \small
  \includegraphics[width=\linewidth]{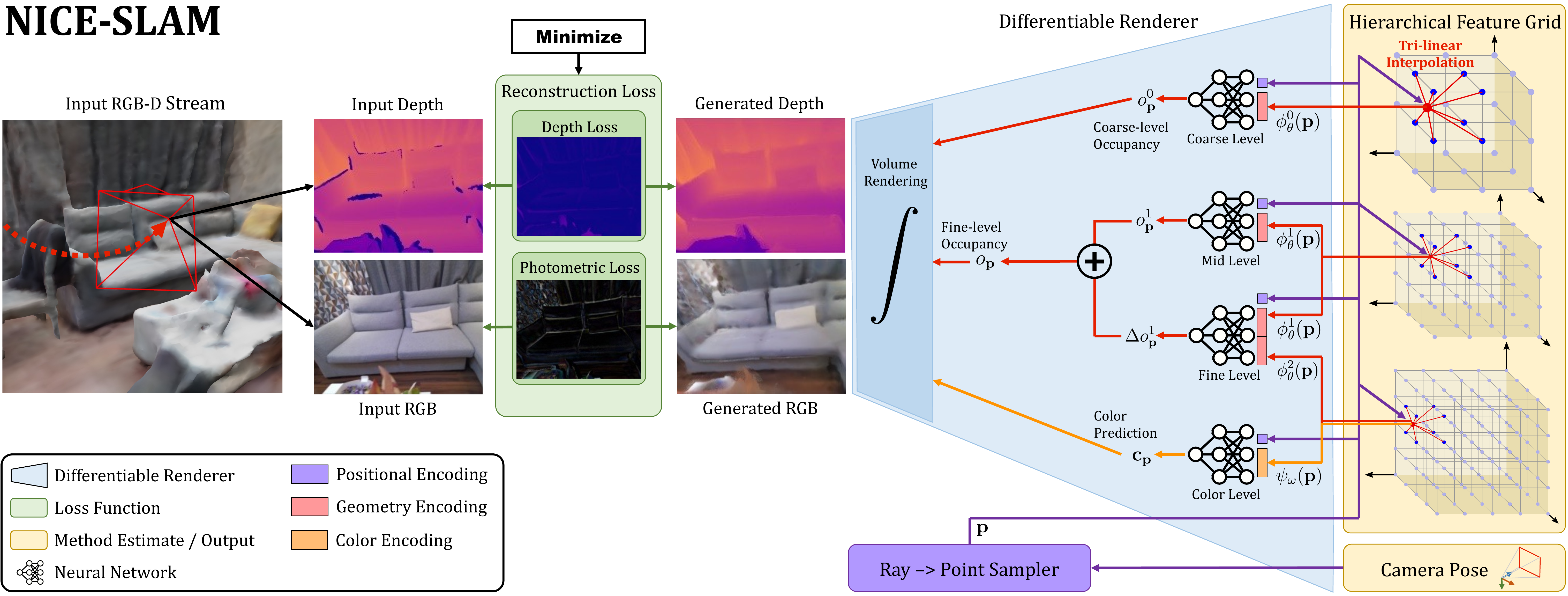}\\[2pt]
  \caption{\textbf{System Overview.} Our method takes an RGB-D image stream as input and outputs both the camera pose as well as a learned scene representation in form of a hierarchical feature grid.
  From \textit{right-to-left}, our pipeline can be interpreted as a generative model which renders depth and color images from a given scene representation and camera pose. At test time we estimate both the scene representation and camera pose by solving the inverse problem via backpropagating the image and depth reconstruction loss through a differentiable renderer (\textit{left-to-right}). Both entities are estimated within an alternating optimization: \textbf{Mapping:} The backpropagation only updates the hierarchical scene representation; \textbf{Tracking:} The backpropagation only updates the camera pose. For better readability we joined the fine-scale grid for geometry encoding with the equally-sized color grid and show them as one grid with two attributes (red and orange).
  }
  \label{fig:system_overview}
\end{figure*}

We provide an overview of our method in~\figref{fig:system_overview}.
We represent the scene geometry and appearance using four feature grids and their corresponding decoders (Sec.~\ref{sec:hierarchical}).
We trace the viewing rays for every pixel using the estimated camera calibration.
By sampling points along a viewing ray and querying the network, we can render both depth and color values of this ray (Sec.~\ref{sec:rendering}).
By minimizing the re-rendering losses for depth and color, we are able to optimize both the camera pose and the scene geometry in an alternating fashion (Sec.~\ref{sec:joint_opt}) for selected keyframes (Sec.~\ref{sec:keyframe_select}).

\subsection{Hierarchical Scene Representation}  \label{sec:hierarchical}
We now introduce our hierarchical scene representation that combines multi-level grid features with pre-trained decoders for occupancy predictions. 
The geometry is encoded into three feature grids $\phi^{l}_\theta$ and their corresponding MLP decoders $f^{l}$, where $l\in\{0, 1, 2\}$ is referred to coarse, mid and fine-level scene details. 
In addition, we also have a single feature grid $\psi_\omega$ and decoder $\bg_\omega$ to model the scene appearance.
Here $\theta$ and $\omega$ indicate the optimizable parameters for geometry and color, i.e.,~the features in the grid and the weights in the color decoder.

\boldparagraph{Mid-\&Fine-level Geometric Representation.}
The observed scene geometry is represented in the mid- and fine-level feature grids.
In the reconstruction process we use these two grids in a coarse-to-fine approach where the geometry is first reconstructed by optimizing the mid-level feature grid, followed by a refinement using the fine-level. 
In the implementation we use voxel grids with side lengths of 32cm and 16cm respectively, except for TUM RGB-D~\cite{sturm2012benchmark} we use 16cm and 8cm.
For the mid-level, the features are directly decoded into occupancy values using the associated MLP $f^1$. For any point $\bp \in \mathbb{R}^3$, we get the occupancy as
\begin{equation}
  o_\bp^1 = f^1(\bp, \phi^1_\theta(\bp)),
  \label{eq:mid_occ}
\end{equation}
where $\phi^1_\theta(\bp)$ denotes that the feature grid is tri-linearly interpolated at the point $\bp$.
The relatively low-resolution allow us to efficiently optimize the grid features to fit the observations.
To capture smaller high-frequency details in the scene geometry we add in the fine-level features in a residual manner. 
In particular, the fine-level feature decoder takes as input both the corresponding mid-level feature and the fine-level feature and outputs an offset from the mid-level occupancy, i.e.,
\begin{equation}
  \Delta o_\bp^1 = f^2(\bp, \phi^1_\theta(\bp), \phi^2_\theta(\bp)),
  \label{eq:high_occ}
\end{equation}
where the final occupancy for a point is given by
\begin{equation}
  o_\bp = o_\bp^1 + \Delta o_\bp^1.
  \label{eq:occ_final}
\end{equation}
Note that we fix the pre-trained decoders $f^1$ and $f^2$, and only optimize the feature grids $\phi^1_\theta$ and $\phi^2_\theta$ throughout the entire optimization process. 
We demonstrate that this helps to stabilize the optimization and learn consistent geometry.

\boldparagraph{Coarse-level Geometric Representation.}
The coarse-level feature grid aims to capture the high-level geometry of the scene (e.g.,~walls, floor, etc), and is optimized independently from the mid- and fine-level. The goal of the coarse-grid is to be able to predict approximate occupancy values outside of the observed geometry (which is encoded in the mid/fine-levels), even when each coarse voxel has only been partially observed. For this reason we use a very low resolution, with a side-length of 2m in the implementation.
Similarly to the mid-level grid, we decode directly into occupancy values by interpolating the features and passing through the MLP $f^0$, i.e.,
\begin{equation}
o_\bp^0 = f^0(\bp, \phi^0_\theta(\bp)).
\label{eq:occ_coarse}
\end{equation}
During tracking, the coarse-level occupancy values are only used for predicting previously unobserved scene parts. %
This \textit{forecasted} geometry allows us to track even when a large portion of the current image is previously unseen.

\boldparagraph{Pre-training Feature Decoders.}
In our framework we use three different fixed MLPs to decode the grid features into occupancy values. The coarse and mid-level decoders are pre-trained as part of ConvONet~\cite{Peng2020ECCV} which consists of a CNN encoder and an MLP decoder. We train both the encoder/decoder using the binary cross-entropy loss between the predicted and the ground-truth value, same as in \cite{Peng2020ECCV}. After training, we only use the decoder MLP, as we will directly optimize the features to fit the observations in our reconstruction pipeline. In this way the pre-trained decoder can leverage resolution-specific priors learned from the training set, when decoding our optimized features.

The same strategy is used to pre-train the fine-level decoder, except that we simply concatenate the
feature $\phi^1_\theta(\bp)$ from the mid-level together with the fine-level feature $\phi^2_\theta(\bp)$ before inputting to the decoder.

\boldparagraph{Color Representation.}
While we are mainly interested in the scene geometry, we also encode the color information allowing us to render RGB images which provides additional signals for tracking.
To encode the color in the scene, we apply another feature grid $\psi_\omega$ and decoder $\bg_\omega$:
\begin{equation}
\bc_\bp = \bg_\omega(\bp, \psi_\omega(\bp)),
\label{eq:color_pred}
\end{equation}
where $\omega$ indicates learnable parameters during optimization. 
Different from the geometry that has strong prior knowledge,
we empirically found that jointly optimizing the color features $\psi_\omega$  and decoder $\bg_\omega$ improves the tracking performance  (c.f.~\tabref{tab:ab_color_BA}). 
Note that, similarly to iMAP~\cite{imap}, this can lead to forgetting problems and the color is only consistent locally. 
If we want to visualize the color for the entire scene, it can be optimized globally as a post-processing step.

\boldparagraph{Network Design.}
For all MLP decoders, we use a hidden feature dimension of 32 and 5 fully-connected blocks. 
Except for the coarse-level geometric representation, we apply a learnable Gaussian positional encoding~\cite{Tancik2020NEURIPS,imap} to $\bp$ before serving as input to MLP decoders. 
We observe this allows discovery of high frequency details for both geometry and appearance.

\subsection{Depth and Color Rendering}  \label{sec:rendering}
Inspired by the recent success of volume rendering in NeRF~\cite{mildenhall2020nerf}, we propose to also use a differentiable rendering process which integrates the predicted occupancy and colors from our 
scene representation in~\secref{sec:hierarchical}. 

Given camera intrinsic parameters and current camera pose, we can calculate the viewing direction $\br$ of a pixel coordinate.
We first sample along this ray $N_\text{strat}$ points for stratified sampling, and also uniformly sample $N_\text{imp}$ points near to the depth\footnote{We empirically define the sampling interval as $\pm 0.05D$, where $D$ is the depth value of the current ray.}. 
In total we sample $N = N_\text{strat} + N_\text{imp}$ points for each ray.
More formally, let $\bp_i = \bo + d_i\br, i\in\{1, \cdots, N\}$ denote the sampling points on the ray $\br$ given the camera origin $\bo$, and $d_i$ corresponds to the depth value of $\bp_i$ along this ray.
For every point $\bp_i$, we can calculate their coarse-level occupancy probability $o^0_{\bp_i}$, fine-level occupancy probability $o_{\bp_i}$, and color value $\bc_{\bp_i}$ using~\eqnref{eq:occ_coarse}, \eqnref{eq:occ_final}, and \eqnref{eq:color_pred}.
Similar to~\cite{oechsle2021unisurf}, we model the ray termination probability at point $\bp_i$ as $w_i^c = o^0_{\bp_i}\prod_{j=1}^{i-1} (1-o^0_{\bp_j})$ for coarse level, and $w_i^f = o_{\bp_i}\prod_{j=1}^{i-1} (1-o_{\bp_j})$ for fine level.

Finally for each ray, the depth at both coarse and fine level, and color can be rendered as:
\begin{equation}
\label{eq:rendering}
  \hat{D}^c = \sum_{i=1}^N w_i^c d_i, \quad \hat{D}^f = \sum_{i=1}^N w_i^f d_i, \quad \hat{I} = \sum_{i=1}^N w_i^f \textbf{c}_i.
\end{equation}
Moreover, we also calculate depth variances along the ray:
\begin{equation}
\label{eq:depth_variance}
\hat{D}^c_{var} = \sum_{i=1}^N w_i^c (\hat{D}^c - d_i)^2\\ \quad \hat{D}^f_{var} = \sum_{i=1}^N w_i^f (\hat{D}^f - d_i)^2.
\end{equation}

\subsection{Mapping and Tracking}  \label{sec:joint_opt}
In this section, we provide details on the optimization of the scene geometry $\theta$ and appearance $\omega$ parameters of our hierarchical scene representation, and of the camera poses.

\boldparagraph{Mapping.}
To optimize the scene representation mentioned in~\secref{sec:hierarchical}, we uniformly sample total $M$ pixels from the current frame and the selected keyframes. 
Next, we perform optimization in a staged fashion to minimize the geometric and photometric losses.  

The geometric loss is simply an $L_1$ loss between the observations and predicted depths at coarse or fine level:
\begin{equation}
    \cL_{g}^{l} = \frac{1}{M}\sum_{m=1}^M {\left|D_m - \hat{D}^l_m\right|},\quad l\in\{c, f\}.
    \label{eq:geo_loss}
\end{equation}
The photometric loss is also an $L_1$ loss between the rendered and observed color values for $M$ sampled pixel:
\begin{equation}
    \cL_p = \frac{1}{M}\sum_{m=1}^M\left|I_m - \hat{I}_m\right|~.
    \label{eq:photo_loss}
\end{equation}
At the first stage, we optimize only the mid-level feature grid $\phi^1_\theta$ using the geometric loss $\cL_{g}^f$ in~\eqnref{eq:geo_loss}.
Next, we jointly optimize both the mid and fine-level $\phi^1_\theta, \phi^2_\theta$ features with the same fine-level depth loss $\cL_{g}^{f}$.
Finally, we conduct a local bundle adjustment (BA) to jointly optimize feature grids at all levels, the color decoder, as well as the camera extrinsic parameters $\{\bR_{i}, \bt_{i} \}$ of $K$ selected keyframes: %
\begin{equation}
\label{eq:geo_opt_loss}
    \min \limits_{\theta, \omega, \{\bR_{i}, \bt_{i} \}}(\cL_g^c + \cL_g^f + \lambda_p \cL_p)~,
\end{equation}
where $\lambda_p$ is the loss weighting factor.

This multi-stage optimization scheme leads to better convergence as the higher-resolution appearance and fine-level features can rely on the already refined geometry coming from mid-level feature grid. 

Note that we parallelize our system in three threads to speed up the optimization process: one thread for coarse-level mapping, one for mid-\&fine-level geometric and color optimization, and another one for camera tracking.

\boldparagraph{Camera Tracking.}
In addition to optimizing the scene representation, we also run in parallel camera tracking to optimize the camera poses of the current frame, i.e., rotation and translation $\{\bR, \bt \}$.
To this end, we sample $M_t$ pixels in the current frame and apply the same photometric loss in~\eqnref{eq:photo_loss} but use a modified geometric loss:
\begin{equation}
    \cL_{g\_var} = \frac{1}{M_t}\sum_{m=1}^{M_t} \frac{\left|D_m - \hat{D}^c_m\right|}{\sqrt{\hat{D}^c_{var}}} + \frac{\left|D_m - \hat{D}^f_m\right|}{\sqrt{\hat{D}^f_{var}}}.
\end{equation}
The modified loss down-weights less certain regions in the reconstructed geometry~\cite{imap,Yang2020CVPR}, e.g., object edges.
The camera tracking is finally formulated as the following minimization problem:
\begin{equation}
    \min \limits_{\bR, \bt} ~(\cL_{g\_var} + \lambda_{pt} \cL_p)~.
    \label{eq:tracking}
\end{equation}

The coarse feature grid is able to perform short-range predictions of the scene geometry. This extrapolated geometry provides a meaningful signal for the tracking as the camera moves into previously unobserved areas. Making it more robust to sudden frame loss or fast camera movement. 
We provide experiments in the supplementary material.

\boldparagraph{Robustness to Dynamic Objects.}
To make the optimization more robust to dynamic objects during tracking, we filter pixels with large depth/color re-rendering loss. In particular, we remove any pixel from the optimization where the loss~\eqnref{eq:tracking} is larger than $10\times$ the median loss value of all pixels in the current frame. \figref{fig:dynamic} shows an example where a dynamic object is ignored since it is not present in the rendered RGB and depth image.
Note that for this task, we only optimize the scene representation during the mapping. Jointly optimizing camera parameters and scene representations under dynamic environments is non-trivial, and we consider it as an interesting future direction.

\subsection{Keyframe Selection}  \label{sec:keyframe_select}
Similar to other SLAM systems, we continuously optimize our hierarchical scene representation with a set of selected keyframes.
We maintain a global keyframe list in the same spirit of iMAP~\cite{imap}, where we incrementally add new keyframes based on the information gain. 
However, in contrast to iMAP~\cite{imap}, we only include keyframes which have visual overlap with the current frame when optimizing the scene geometry. This is possible since we are able to make local updates to our grid-based representation, and we do not suffer from the same forgetting problems as~\cite{imap}.
This keyframe selection strategy not only ensures the geometry outside of the current view remains static, but also results in a very efficient optimization problem as we only optimize the necessary parameters each time.
In practice, we first randomly sample pixels and back-project the corresponding depths using the optimized camera pose. 
Then, we project the point cloud to every keyframe in the global keyframe list. From those keyframes that have points projected onto, we randomly select $K-2$ frames. 
In addition, we also include the most recent keyframe and the current frame in the scene representation optimization, forming a total number of $K$ active frames.
Please refer to~\secref{sec:ablation_study} for an ablation study on the keyframe selection strategy.

\section{Experiments} \label{sec:experiments}
We evaluate our SLAM framework on a wide variety of datasets, both real and synthetic, of varying size and complexity.
We also conduct a comprehensive ablation study that supports our design choices.

\subsection{Experimental Setup}

\boldparagraph{Datasets.} 
We consider 5 versatile datasets: Replica~\cite{replica19arxiv}, ScanNet~\cite{dai2017scannet}, TUM RGB-D dataset~\cite{sturm2012benchmark}, Co-Fusion dataset~\cite{runz2017co}, as well as a self-captured large apartment with multiple rooms.
We follow the same pre-processing step for TUM RGB-D as in ~\cite{teed2021tangent}.

\boldparagraph{Baselines.}
We compare to TSDF-Fusion~\cite{curless1996volumetric} with our camera poses with a voxel grid resolution of $256^3$ (results of higher resolutions are reported in the supp. material),
DI-Fusion~\cite{huang2021di} using their official implementation\footnote{\url{https://github.com/huangjh-pub/di-fusion}}, as well as our faithful iMAP~\cite{imap} re-implementation: iMAP$^*$.
Our re-implementation has similar performance as the original iMAP in both scene reconstruction and camera tracking.

\boldparagraph{Metrics.} 
We use both 2D and 3D metrics to evaluate the scene geometry.
For the 2D metric, we evaluate the L1 loss on 1000 randomly-sampled depth maps from both reconstructed and ground truth meshes. 
For fair comparison, we apply the bilateral solver~\cite{barron2016fast} to DI-Fusion~\cite{huang2021di} and TSDF-Fusion to fill depth holes before calculating the average L1 loss. 
For 3D metrics, we follow~\cite{imap} and consider \textit{Accuracy} [cm], \textit{Completion} [cm], and \textit{Completion Ratio} [$<$ 5cm \%], except that we remove unseen regions that are not inside any camera's viewing frustum.
As for the evaluation of camera tracking, we use ATE RMSE~\cite{sturm2012benchmark}.
If not specified otherwise, by default we report the average results of 5 runs.

\boldparagraph{Implementation Details.} 
We run our SLAM system on a desktop PC with a 3.80GHz Intel i7-10700K CPU and an NVIDIA RTX 3090 GPU.
In all our experiments, we use the number of sampling points on a ray $N_\text{strat}=32$ and $N_\text{imp}=16$, photometric loss weighting $\lambda_p=0.2$ and $\lambda_{pt}=0.5$.
For small-scale synthetic datasets (Replica and Co-Fusion), we select $K=5$ keyframes and sample $M=1000$ and $M_t=200$ pixels. 
For large-scale real datasets (ScanNet and our self-captured scene), we use $K=10$, $M=5000$, $M_t=1000$. 
As for the challenging TUM RGB-D dataset, we use $K=10$, $M=5000$, $M_t=5000$. 
For our re-implementation iMAP$^*$, we follow all the hyperparameters mentioned in~\cite{imap} except that we set the number of sampling pixels to 5000 since it leads to better performance in both reconstruction and tracking.

\subsection{Evaluation of Mapping and Tracking}

\begin{figure}[tbp]
  \centering
  \scriptsize
  \setlength{\tabcolsep}{0.5pt}
  \newcommand{\sz}{0.17}  %

  \begin{tabular}{lccccc}

    &\multicolumn{2}{c}{\tt room-2} & \multicolumn{2}{c}{\tt office-2} \\
    \makecell{\rotatebox{90}{iMAP$^*$~\cite{imap}}}  &
    \makecell{\includegraphics[height=\sz\linewidth]{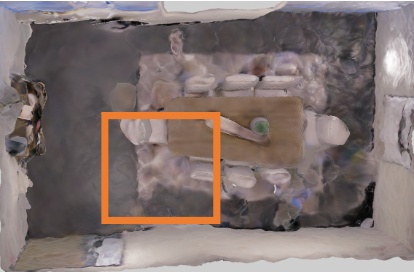}} & 
    \makecell{\includegraphics[height=\sz\linewidth]{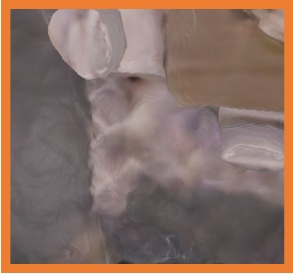}} &
    \makecell{\includegraphics[height=\sz\linewidth]{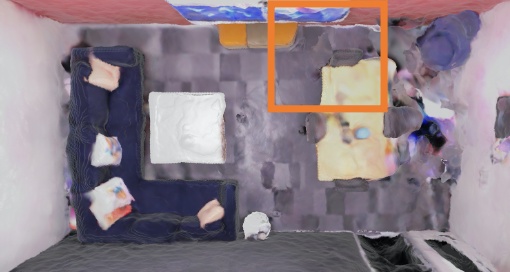}} &
    \makecell{\includegraphics[height=\sz\linewidth]{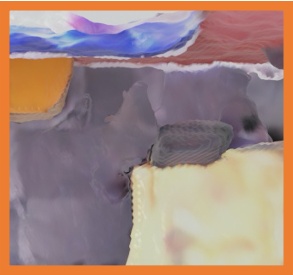}} \\
    \makecell{\rotatebox{90}{\ours{}}}  &
    \makecell{\includegraphics[height=\sz\linewidth]{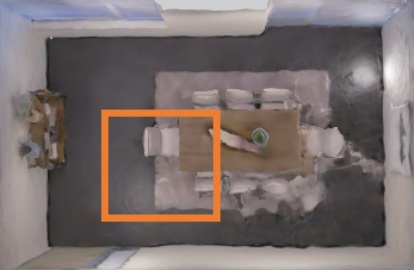}} & 
    \makecell{\includegraphics[height=\sz\linewidth]{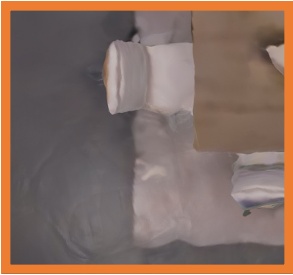}} &
    \makecell{\includegraphics[height=\sz\linewidth]{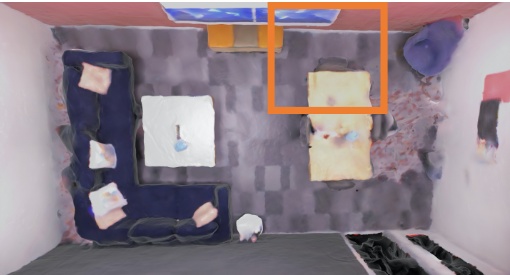}} &
    \makecell{\includegraphics[height=\sz\linewidth]{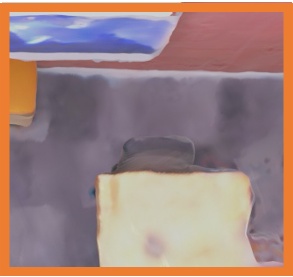}} \\
    \makecell{\rotatebox{90}{GT} } &
    \makecell{\includegraphics[height=\sz\linewidth]{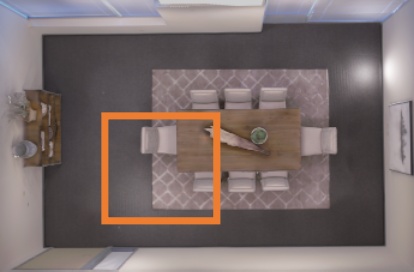}} & 
    \makecell{\includegraphics[height=\sz\linewidth]{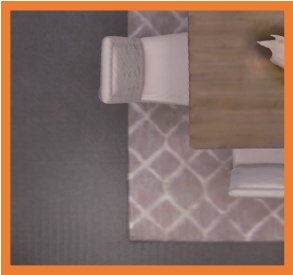}} &
    \makecell{\includegraphics[height=\sz\linewidth]{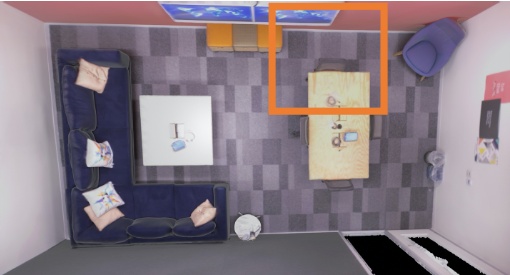}} &
    \makecell{\includegraphics[height=\sz\linewidth]{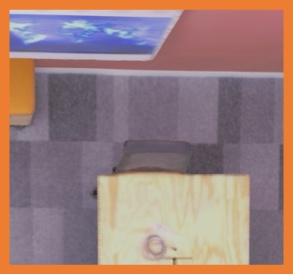}} \\
    
  \end{tabular}
  \caption{\textbf{Reconstruction Results on the Replica Dataset~\cite{replica19arxiv}.}  iMAP$^*$ refers to our iMAP re-implementation.
  }
  \label{fig:replica_zoom}
\end{figure}

\boldparagraph{Evaluation on Replica~\cite{replica19arxiv}.}
To evaluate on Replica~\cite{replica19arxiv}, we use the same rendered RGB-D sequence provided by the authors of iMAP. With the hierarchical scene representation, our method is able to reconstruct the geometry precisely within limited iterations. As shown in Table~\ref{tab:replica}, \ours{} significantly outperforms baseline methods on almost all metrics, while keeping a reasonable memory consumption.
Qualitatively, we can see from~\figref{fig:replica_zoom} that our method produces sharper geometry and less artifacts.

\begin{table}[tb!]
  \centering
     \resizebox{\linewidth}{!}{
    \begin{tabular}{lccccc}
      \toprule
      &\multicolumn{1}{c}{TSDF-Fusion~\cite{curless1996volumetric}} & {iMAP$^*$}~\cite{imap} & DI-Fusion~\cite{huang2021di}&  \textbf{NICE-SLAM}\\
         \midrule
        {Mem. (MB) $\downarrow$} &   67.10 & \bf 1.04 & 3.78 & 12.02   \\
        \midrule
        {\bf Depth L1} $\downarrow$ & 7.57 & 7.64 & 23.33 & \textbf{3.53}\\
        {\bf Acc.} $\downarrow$ &  \textbf{1.60} & 6.95 & 19.40 & 2.85\\
        {\bf Comp.} $\downarrow$  & 3.49 & 5.33 & 10.19 & \textbf{3.00}   \\
        {\bf Comp. Ratio} $\uparrow$  & 86.08 & 66.60 & 72.96 & \textbf{89.33}\\
       \bottomrule
    \end{tabular}}%
    \caption{\textbf{Reconstruction Results for the Replica Dataset~\cite{replica19arxiv} (average over 8 scenes).} 
    iMAP$^*$ indicates our re-implementation of iMAP.
    TSDF-Fusion uses camera poses from~\ours.
    Detailed metrics for each scene can be found in the supp. material.
    }
    \label{tab:replica}
\end{table}

\boldparagraph{Evaluation on TUM RGB-D~\cite{sturm2012benchmark}.}
We also evaluate the camera tracking performance on the small-scale TUM RGB-D dataset.
As shown in Table~\ref{tab:tum_rmse}, our method outperforms iMAP and DI-Fusion even though ours is by design more suitable for large scenes.
As can be noticed, the state-of-the-art approaches for tracking (e.g. BAD-SLAM~\cite{schops2019bad}, %
ORB-SLAM2~\cite{Mur-Artal:etal:TRO2017}) still outperform the methods based on implicit scene representations (iMAP~\cite{imap} and ours).
Nevertheless, our method significantly reduces the gap between these two categories, while retaining the representational advantages of implicit representations. 
\begin{table}[!tb]
  \centering
  \footnotesize
  \setlength{\tabcolsep}{0.7em}
  \resizebox{\linewidth}{!}{
    \begin{tabular}{lccc}
      \toprule
         & \tt{fr1/desk} &  \tt{fr2/xyz} &  \tt{fr3/office} \\
         \midrule
        
        {iMAP}~\cite{imap}      & 4.9 & 2.0 & 5.8  \\
        {iMAP$^*$}~\cite{imap} & 7.2 & 2.1  & 9.0 \\
        {DI-Fusion~\cite{huang2021di}} & 4.4 & 2.3 & 15.6 \\
        {\bf \ours{}}           & 2.7 & 1.8 & 3.0 \\
      \midrule
        {BAD-SLAM}\cite{schops2019bad} & 1.7  & 1.1  & 1.7 \\
        {Kintinuous}\cite{Whelan:etal:RSSRGBD2012} & 3.7  &  2.9  & 3.0 \\
        {ORB-SLAM2}\cite{Mur-Artal:etal:TRO2017} & \bf 1.6  & \bf 0.4  & \bf 1.0 \\
       \bottomrule
    \end{tabular}}
    \vspace{2pt}
    \caption{\textbf{Camera Tracking Results on TUM RGB-D~\cite{sturm2012benchmark}.}
    ATE RMSE [cm] ($\downarrow$) is used as the evaluation metric.
    \ours{} reduces the gap between SLAM methods with neural implicit representations and traditional approaches. 
    We report the best out of 5 runs for all methods in this table. 
    The numbers for iMAP, BAD-SLAM, Kintinuous, and ORB-SLAM2 are taken from~\cite{imap}.
    }
    \label{tab:tum_rmse}
\end{table}

\begin{figure*}[htbp]
  \centering
  \footnotesize
  \setlength{\tabcolsep}{1.5pt}
  \newcommand{\sz}{0.19}
  \begin{tabular}{ccccc}
    TSDF-Fusion w/ our pose &
    iMAP$^*$~\cite{imap} & DI-Fusion~\cite{huang2021di} & \ours{} & ScanNet Mesh\\
    \includegraphics[width=\sz\linewidth]{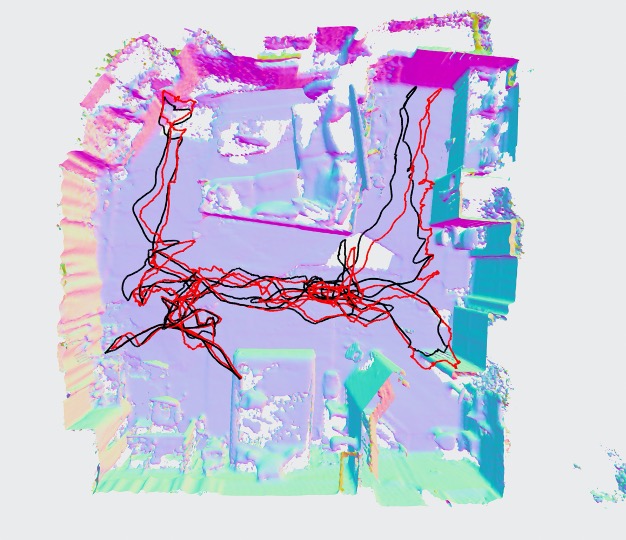} &
    \includegraphics[width=\sz\linewidth]{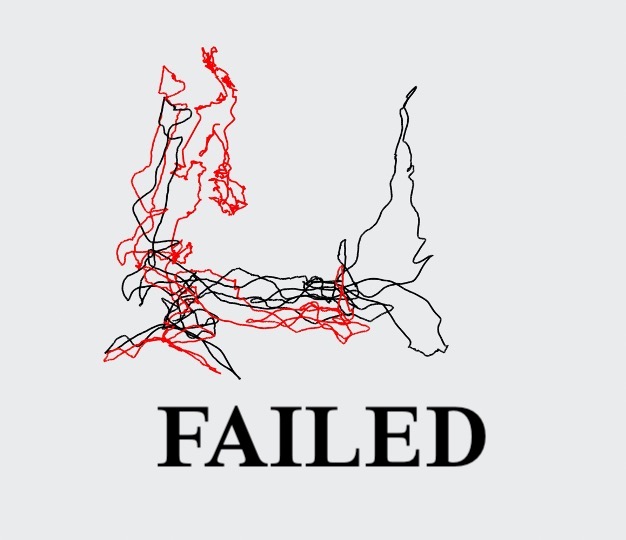} &
    \includegraphics[width=\sz\linewidth]{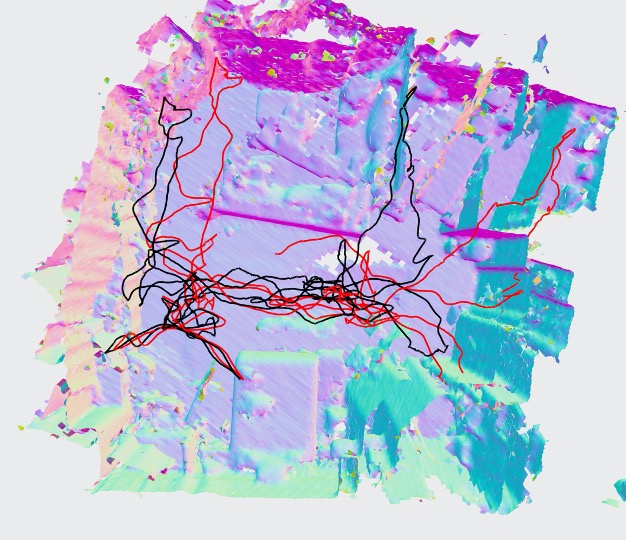} &
    \includegraphics[width=\sz\linewidth]{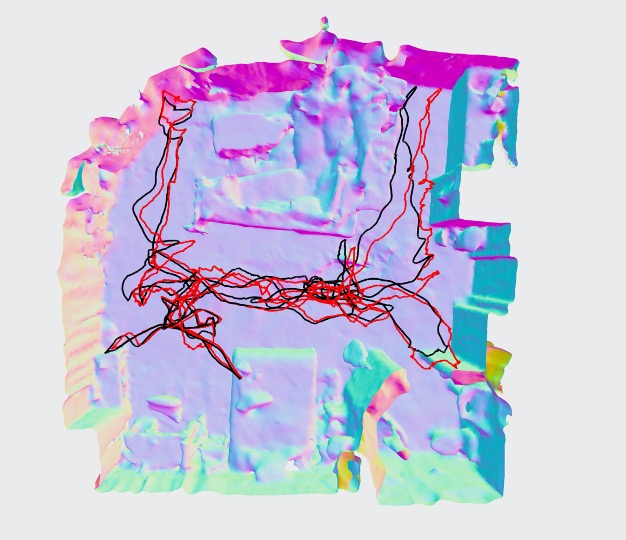} &
    \includegraphics[width=\sz\linewidth]{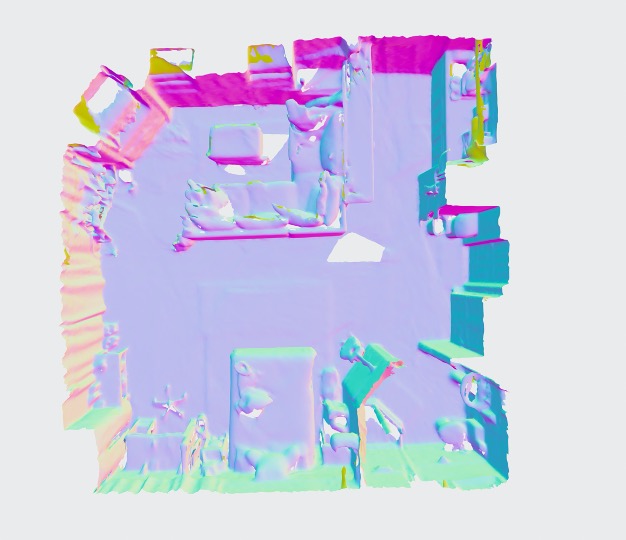} \\
    \includegraphics[width=\sz\linewidth]{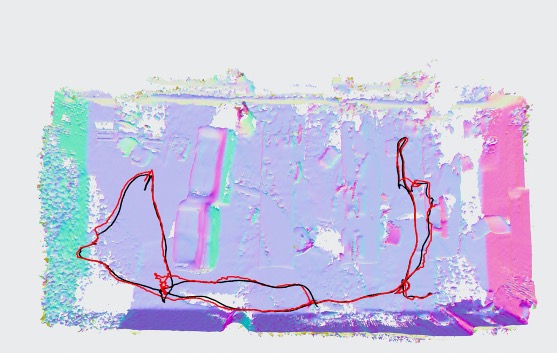} &
    \includegraphics[width=\sz\linewidth]{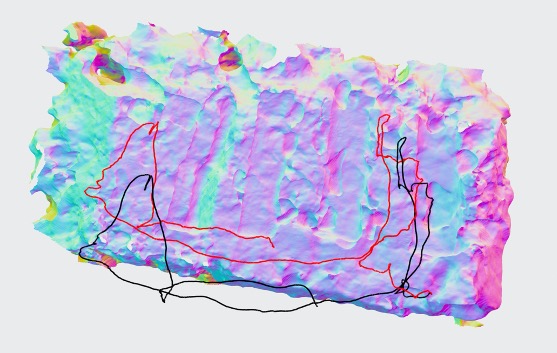} &
    \includegraphics[width=\sz\linewidth]{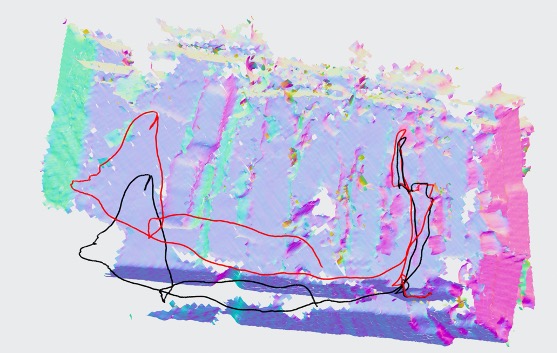} &
    \includegraphics[width=\sz\linewidth]{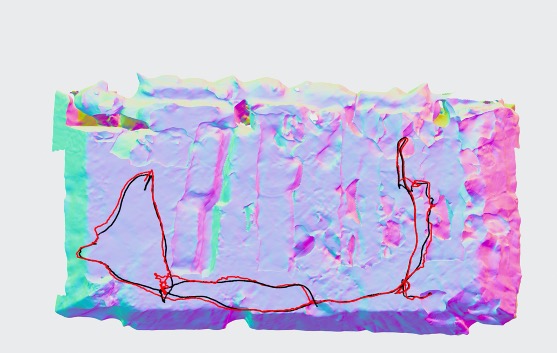} &
    \includegraphics[width=\sz\linewidth]{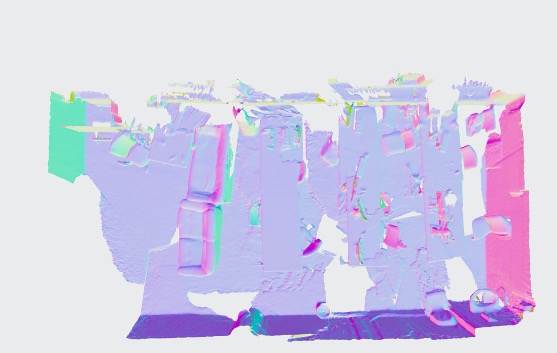} \\
    \includegraphics[width=\sz\linewidth]{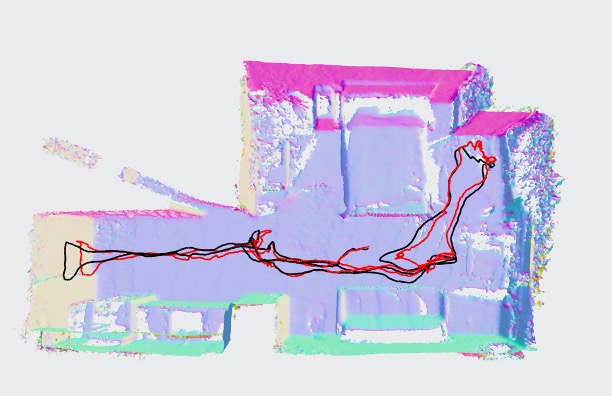} &
    \includegraphics[width=\sz\linewidth]{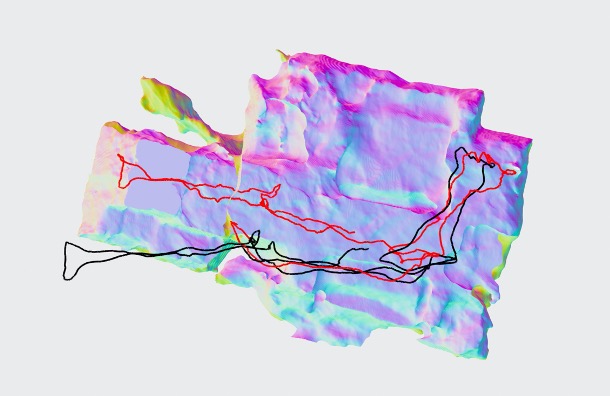} &
    \includegraphics[width=\sz\linewidth]{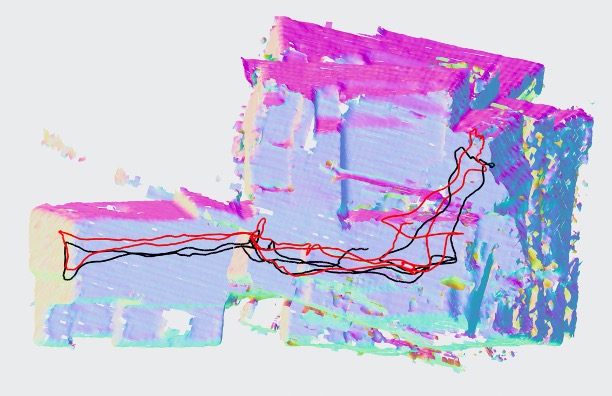} &
    \includegraphics[width=\sz\linewidth]{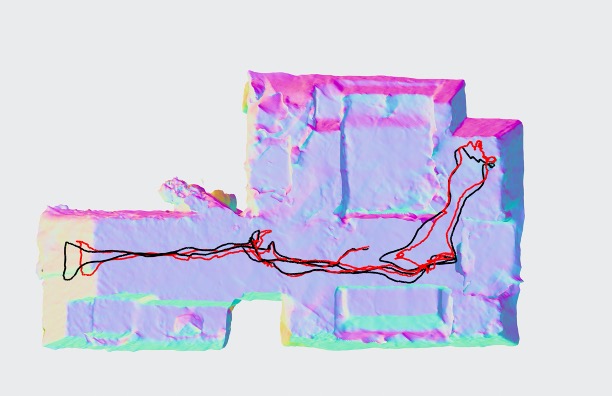} &
    \includegraphics[width=\sz\linewidth]{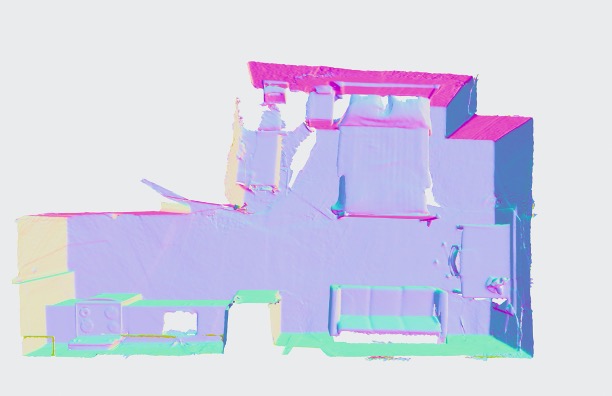} \\
  \end{tabular} 
  \caption{\textbf{3D Reconstruction and Tracking on ScanNet~\cite{dai2017scannet}}.
  The black trajectory is from ScanNet~\cite{dai2017scannet}, the red trajectory is the methods' tracking result.
  We tried various hyperparameters for iMAP$^*$ and present the best results which are mostly inferior.
  }
  \label{fig:scannet2}
\end{figure*}

\boldparagraph{Evaluation on ScanNet~\cite{dai2017scannet}.}
We select multiple large scenes from ScanNet \cite{dai2017scannet} to benchmark the scalability of different methods. 
For the geometry shown in~\figref{fig:scannet2}, we can clearly notice that \ours{} produces sharper and more detailed geometry over TSDF-Fusion, DI-Fusion and iMAP$^*$. 
In terms of tracking, as can be observed, iMAP$^*$ and DI-Fusion either completely fails or introduces large drifting, while our method successfully reconstructs the entire scene. 
Quantitatively speaking, our tracking results are also significantly more accurate than both DI-Fusion and iMAP$^*$ as shown in~\tabref{tab:scannet}.

\begin{figure*}[htbp]
  \centering
  \footnotesize
  \setlength{\tabcolsep}{1.5pt}
  \newcommand{\sz}{0.24}
  \newcommand{\szd}{0.246}
  \begin{tabular}{cccc}
    iMAP$^*$~\cite{imap} & 
    DI-Fusion~\cite{huang2021di} &
    \ours{} &
    Redwood~\cite{redwood} \\
    \includegraphics[width=\sz\linewidth]{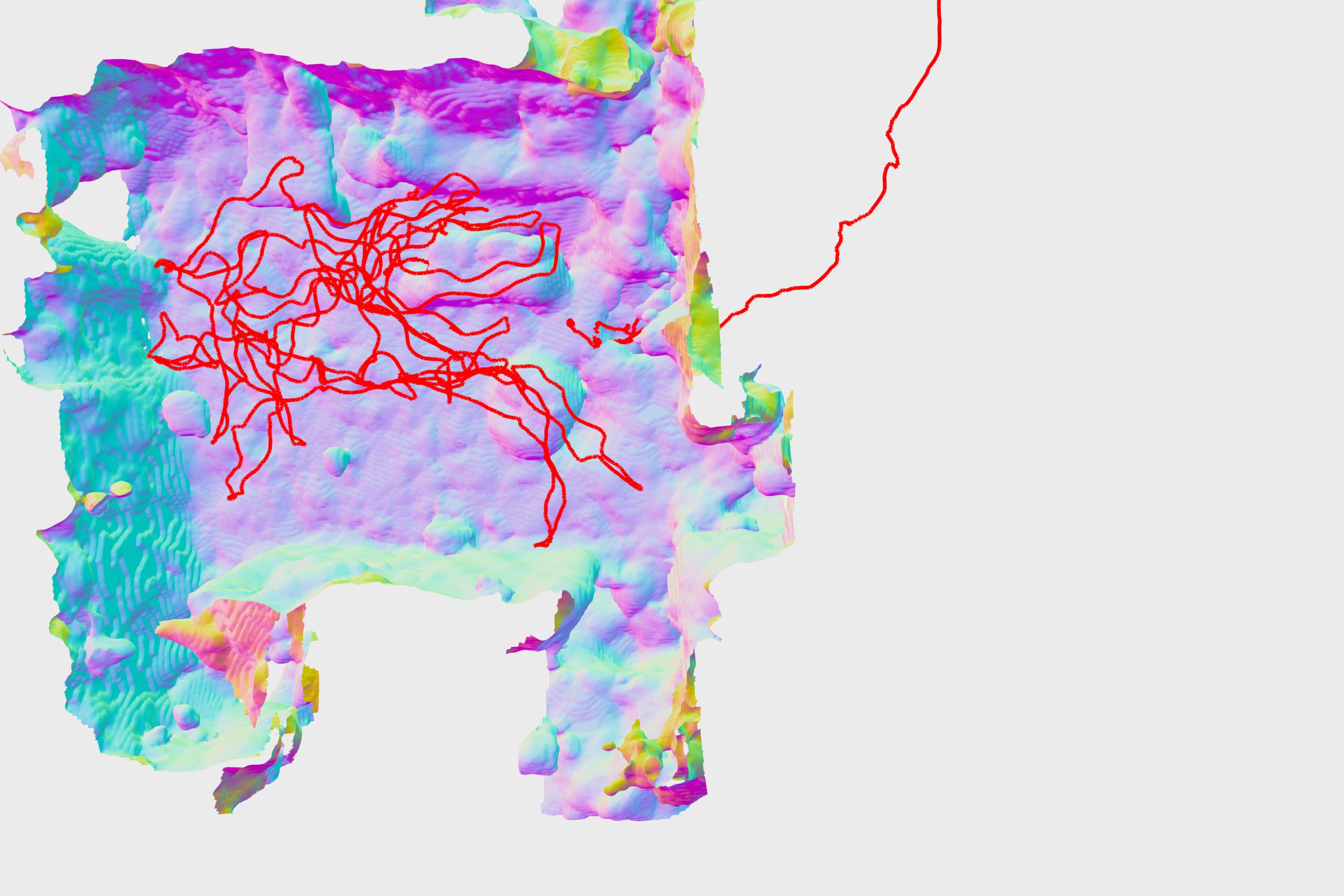}  &
    \includegraphics[width=\szd\linewidth]{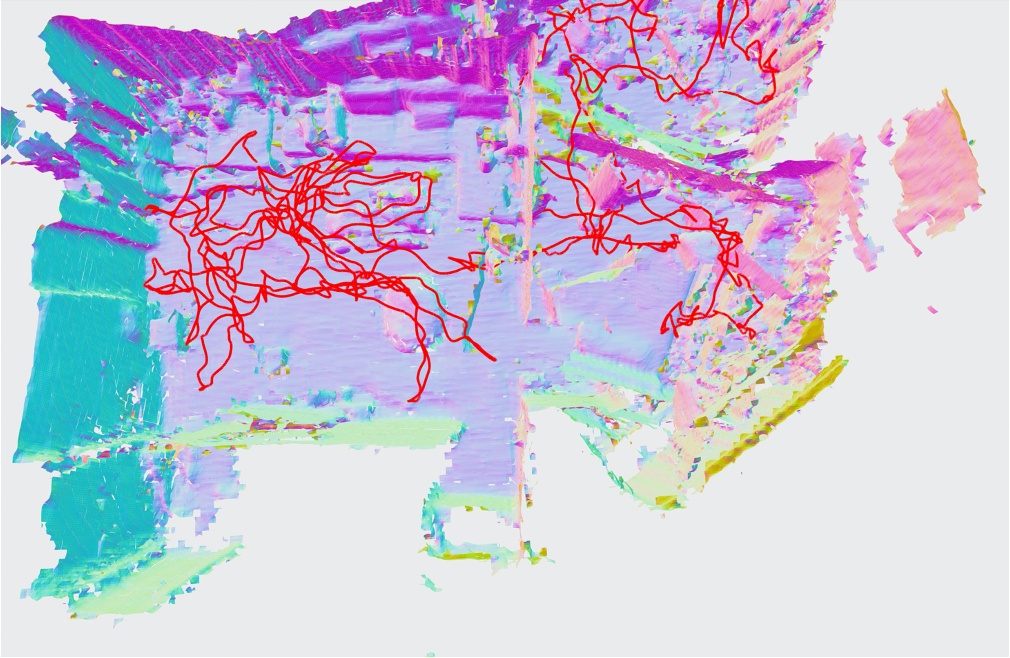}  &
    \includegraphics[width=\sz\linewidth]{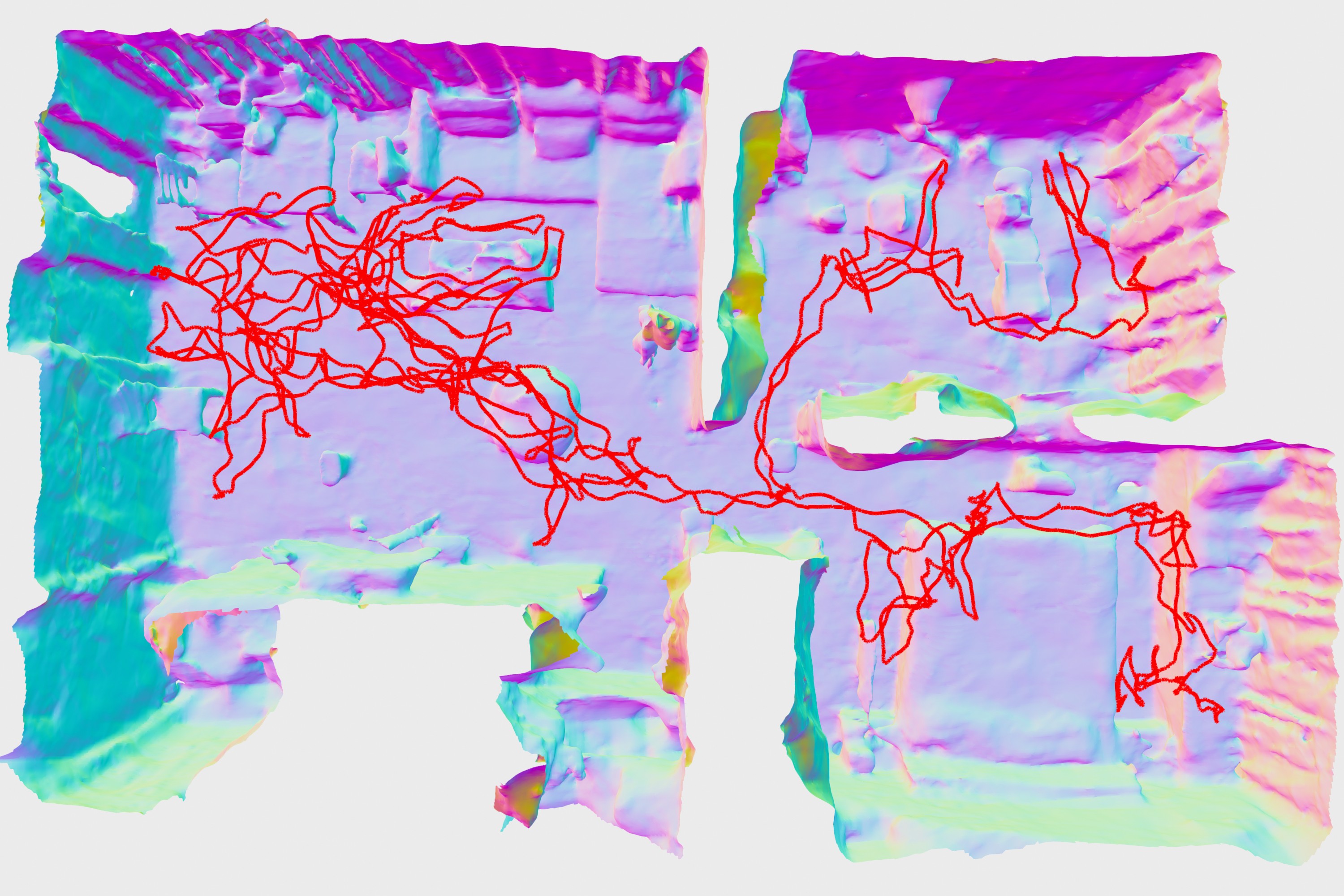} &
    \includegraphics[width=\sz\linewidth]{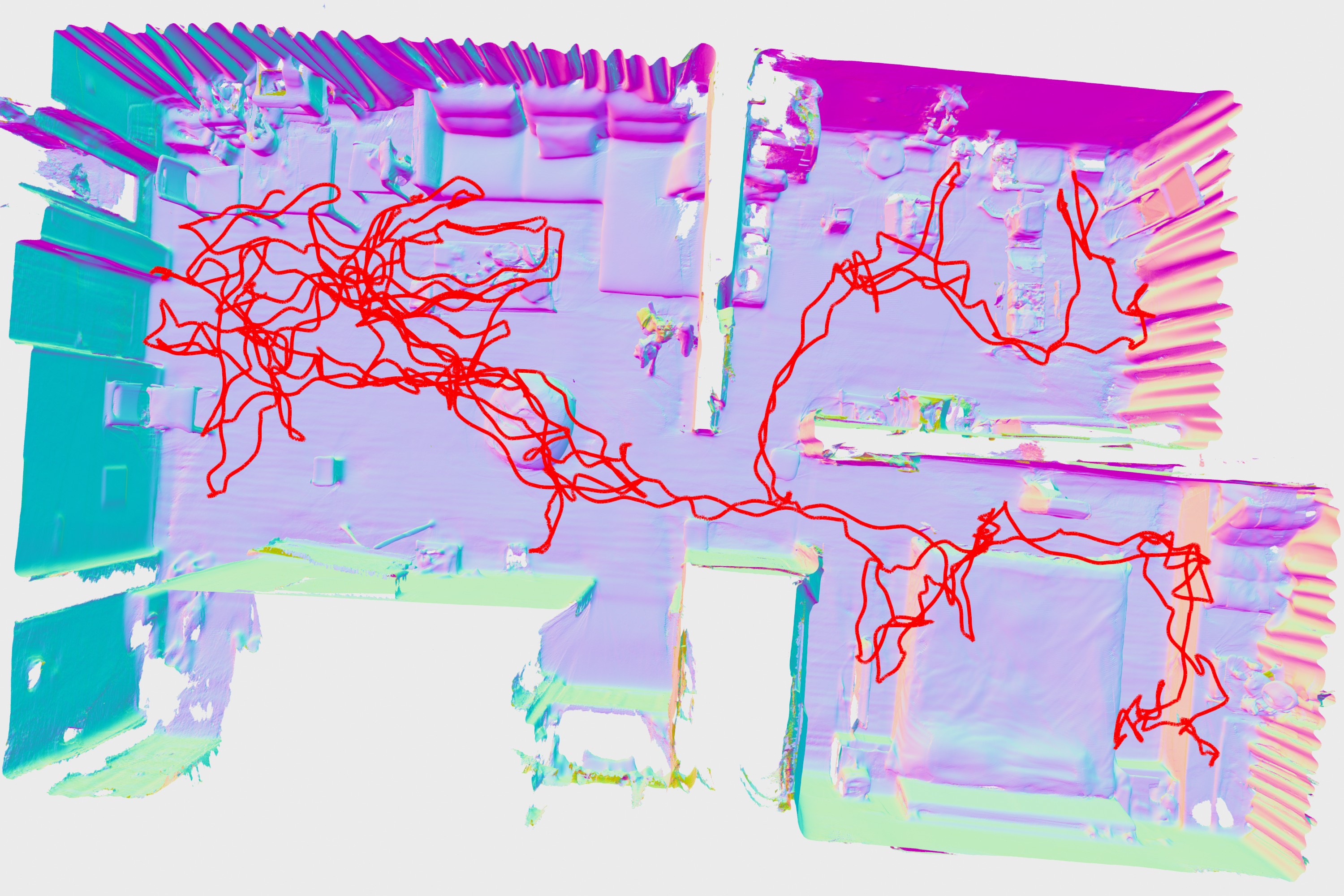} \\
  \end{tabular} 
  \caption{\textbf{3D Reconstruction and Tracking on a Multi-room Apartment.} 
  The camera tracking trajectory is shown in \red{red}.
  iMAP$^*$ and DI-Fusion failed to reconstruct the entire sequence.
  We also show the result of an offline method~\cite{redwood} for reference.}
  \label{fig:self}
\end{figure*}

\boldparagraph{Evaluation on a Larger Scene.}
To evaluate the scalability of our method we captured a sequence in a large apartment with multiple rooms. 
Fig.~\ref{fig:teaser} and Fig.~\ref{fig:self} show the reconstructions obtained using \ours{}, DI-Fusion~\cite{huang2021di} and iMAP$^*$~\cite{imap}. 
For reference we also show the 3D reconstruction using the offline tool Redwood~\cite{redwood} in Open3D~\cite{open3d}. 
We can see that \ours{} has comparable results with the offline method, while iMAP$^*$ and DI-Fusion fails to reconstruct the full sequence.

\begin{table}[t!]
  \centering
  \footnotesize
  \setlength{\tabcolsep}{4pt}
  \resizebox{\linewidth}{!}{
    \begin{tabular}{lccccccc}
      \toprule
         Scene ID & \multicolumn{1}{c}{\makecell{\tt{0000}}} & \multicolumn{1}{c}{\makecell{\tt{0059}}} &  \multicolumn{1}{c}{\makecell{\tt{0106}}} & \multicolumn{1}{c}{\makecell{\tt{0169}}} & \multicolumn{1}{c}{\makecell{\tt{0181}}} & \multicolumn{1}{c}{\makecell{\tt{0207}}} &
         Avg.\\
         \midrule
        {iMAP$^*$~\cite{imap}} & 55.95 & 32.06 & 17.50 &70.51 & 32.10 & 11.91 & 36.67\\
        {DI-Fusion ~\cite{huang2021di}} & 62.99 & 128.00 & 18.50 & 75.80 & 87.88 & 100.19 & 78.89 \\
        {\bf \ours{}} & \textbf{8.64} & \textbf{12.25} & \textbf{8.09} & \textbf{10.28} & \textbf{12.93} & \textbf{5.59} & \textbf{9.63}\\
       \bottomrule
    \end{tabular}}%
    \caption{\textbf{Camera Tracking Results on ScanNet~\cite{dai2017scannet}.} Our approach yields consistently better results on this dataset. ATE RMSE ($\downarrow$) is used as the evaluation metric.}
    \label{tab:scannet}
\end{table}

\subsection{Performance Analysis}
Besides the evaluation on scene reconstruction and camera tracking on various datasets, in the following we also evaluate other characteristics of the proposed pipeline.
\boldparagraph{Computation Complexity.} First, we compare the number of floating point operations (FLOPs) needed for querying color and occupancy/volume density of one 3D point, see~\tabref{tab:time}. 
Our method requires only 1/4 FLOPs of iMAP.
It is worth mentioning that FLOPs in our approach remain the same even for very large scenes.
In contrast, due to the use of a single MLP in iMAP, the capacity limit of the MLP might require more parameters that result in more FLOPs.

\boldparagraph{Runtime.}
We also compare in~\tabref{tab:time} the runtime for tracking and mapping using the same number of pixel samples ($M_t=200$ for tracking and $M=1000$ for mapping).
We can notice that our method is over $2\times$ and $3\times$ faster than iMAP in tracking and mapping. 
This indicates the advantage of using feature grids with shallow MLP decoders over a single heavy MLP.
\begin{table}[t!]
  \centering
  \footnotesize
  \setlength{\tabcolsep}{3pt}
    \begin{tabular}{lcccc}
      \toprule
         &FLOPs $[\times10^3]\!\!\downarrow$ & Tracking [ms]$\downarrow$ &  Mapping [ms]$\downarrow$ \\
         \midrule
        {iMAP~\cite{imap}}    & 443.91 & 101 & 448   \\
        {\bf \ours{}}\hspace{-1em}  & \bf 104.16 & \bf 47  & \bf 130  \\
       \bottomrule
    \end{tabular}%
    \caption{\textbf{Computation \& Runtime.} Our scene representation does not only improve the reconstruction and tracking quality, but is also faster. The runtimes for iMAP are taken from~\cite{imap}.}
    \label{tab:time}
\end{table}

\begin{figure}[!tb]
  \centering
  \footnotesize
  \setlength{\tabcolsep}{1.5pt}
  \newcommand{\sz}{0.32}
  \begin{tabular}{ccc}
    Pixel Samples & Our RGB & Our Depth \\
    \includegraphics[width=\sz\linewidth]{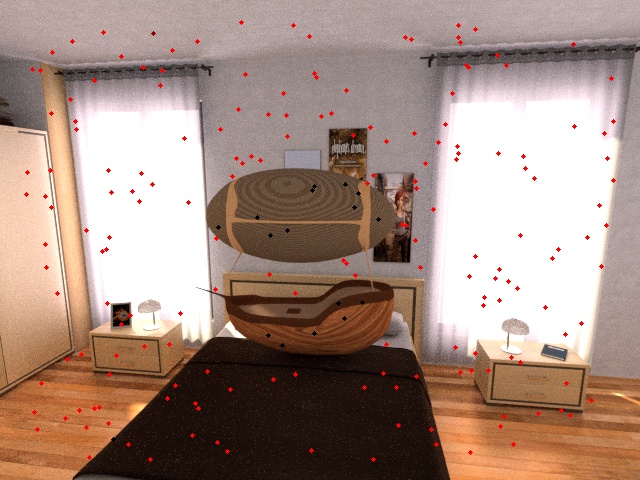} &
    \includegraphics[width=\sz\linewidth]{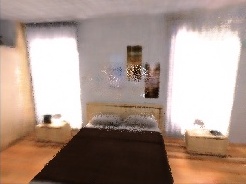} &
    \includegraphics[width=\sz\linewidth]{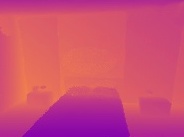} \\
  \end{tabular} 
  \caption{\textbf{Robustness to Dynamic Objects.} We show the sampled pixels overlaid on an image with a dynamic object in the center (left), our rendered RGB (middle) and our rendered depth (right) to illustrate the ability of handling dynamic environments.
  The masked pixel samples during tracking are colored in black, while the used ones are shown in red.}
  \label{fig:dynamic}
\end{figure}

\paragraph{Robustness to Dynamic Objects.}
Here we consider the Co-Fusion dataset~\cite{runz2017co} which contains dynamically moving objects. 
As illustrated in~\figref{fig:dynamic}, our method correctly identifies and ignores pixel samples falling into the dynamic object during optimization, which leads to better scene representation modelling (see the rendered RGB and depths).
Furthermore, we also compare with iMAP$^*$ on the same sequence for camera tracking.
The ATE RMSE scores of ours and iMAP$^*$ is 1.6cm and 7.8cm respectively, 
which clearly demonstrates our robustness to dynamic objects.

\paragraph{Geometry Forecast and Hole Filling.}

\begin{figure}[!tb]
  \centering
  \footnotesize
  \setlength{\tabcolsep}{1.5pt}
  \newcommand{\sz}{0.3}
  \begin{tabular}{ccc}

    iMAP$^*$~\cite{imap} & \ours{} w/o coarse & \ours{} \\
    \includegraphics[width=\sz\linewidth]{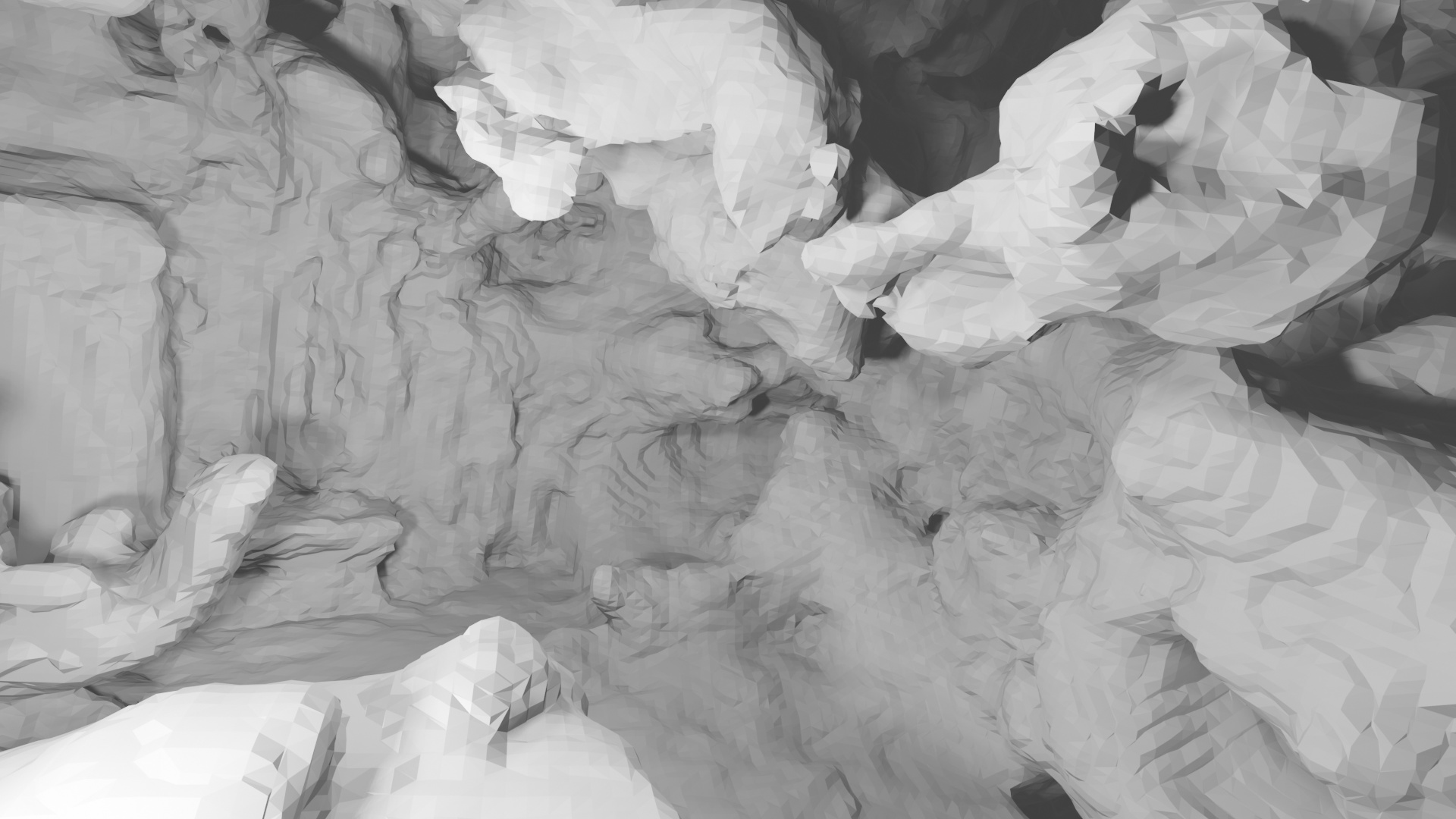} &
    \includegraphics[width=\sz\linewidth]{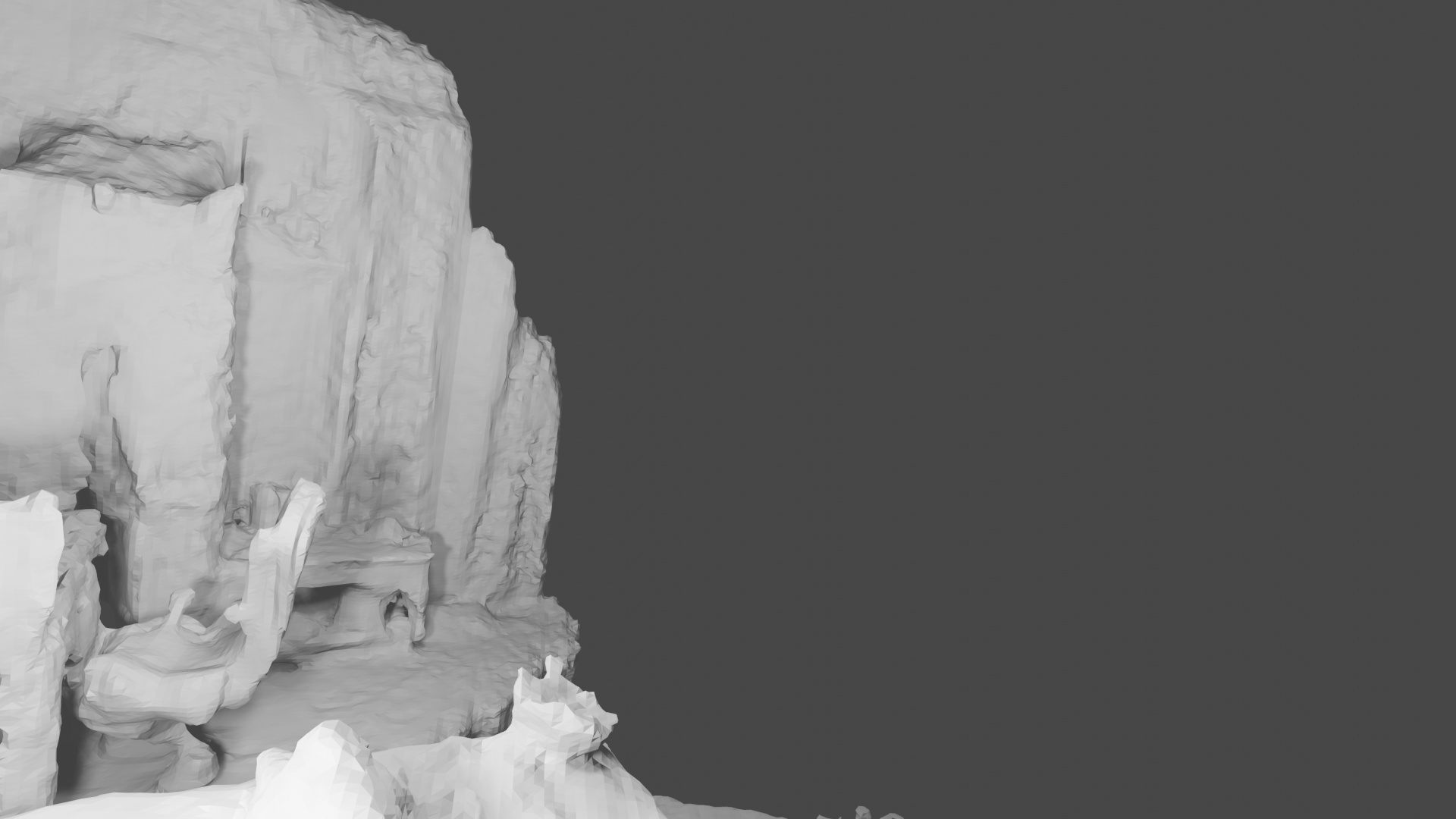}&
    \includegraphics[width=\sz\linewidth]{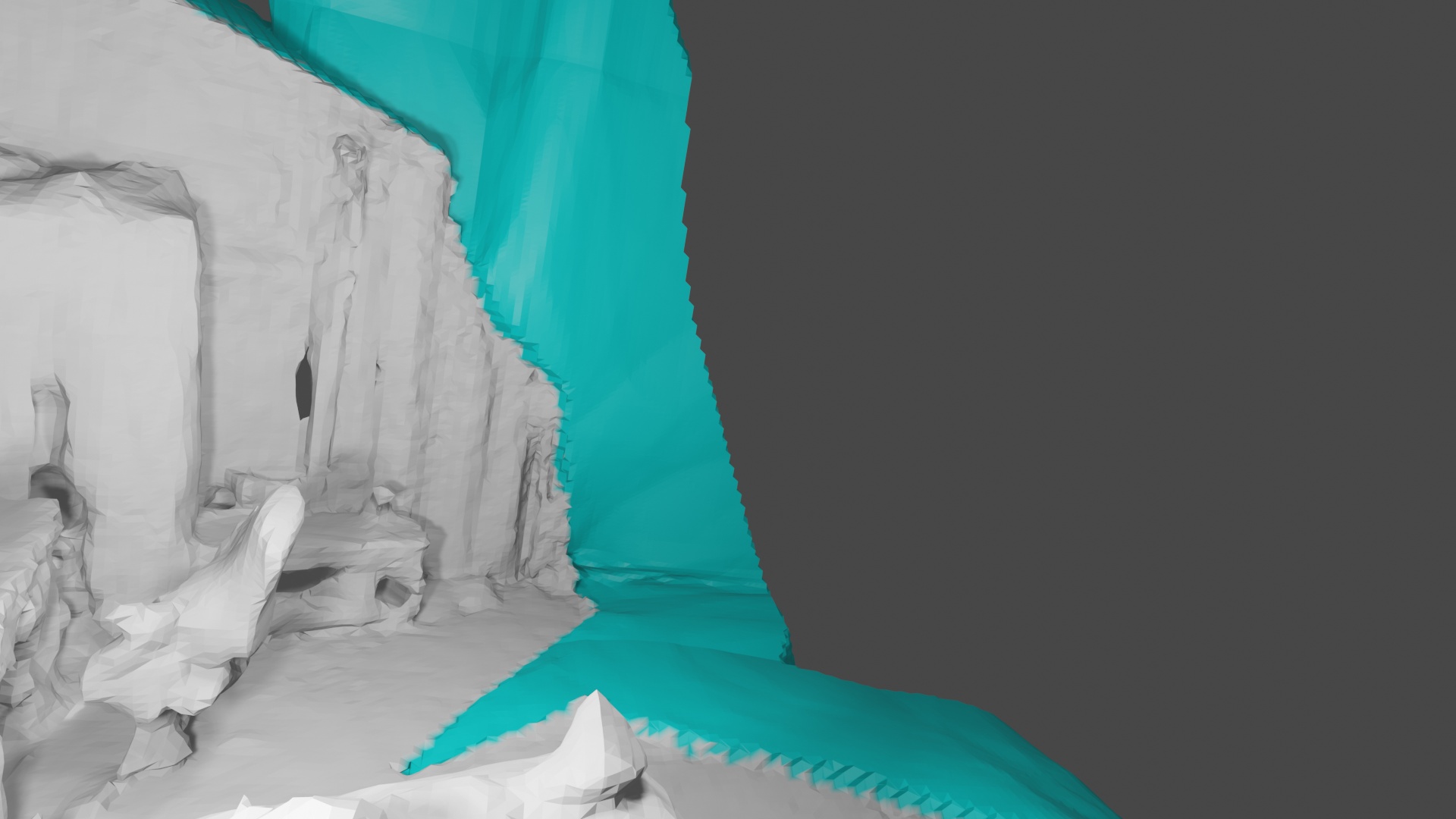} \\
    \includegraphics[width=\sz\linewidth]{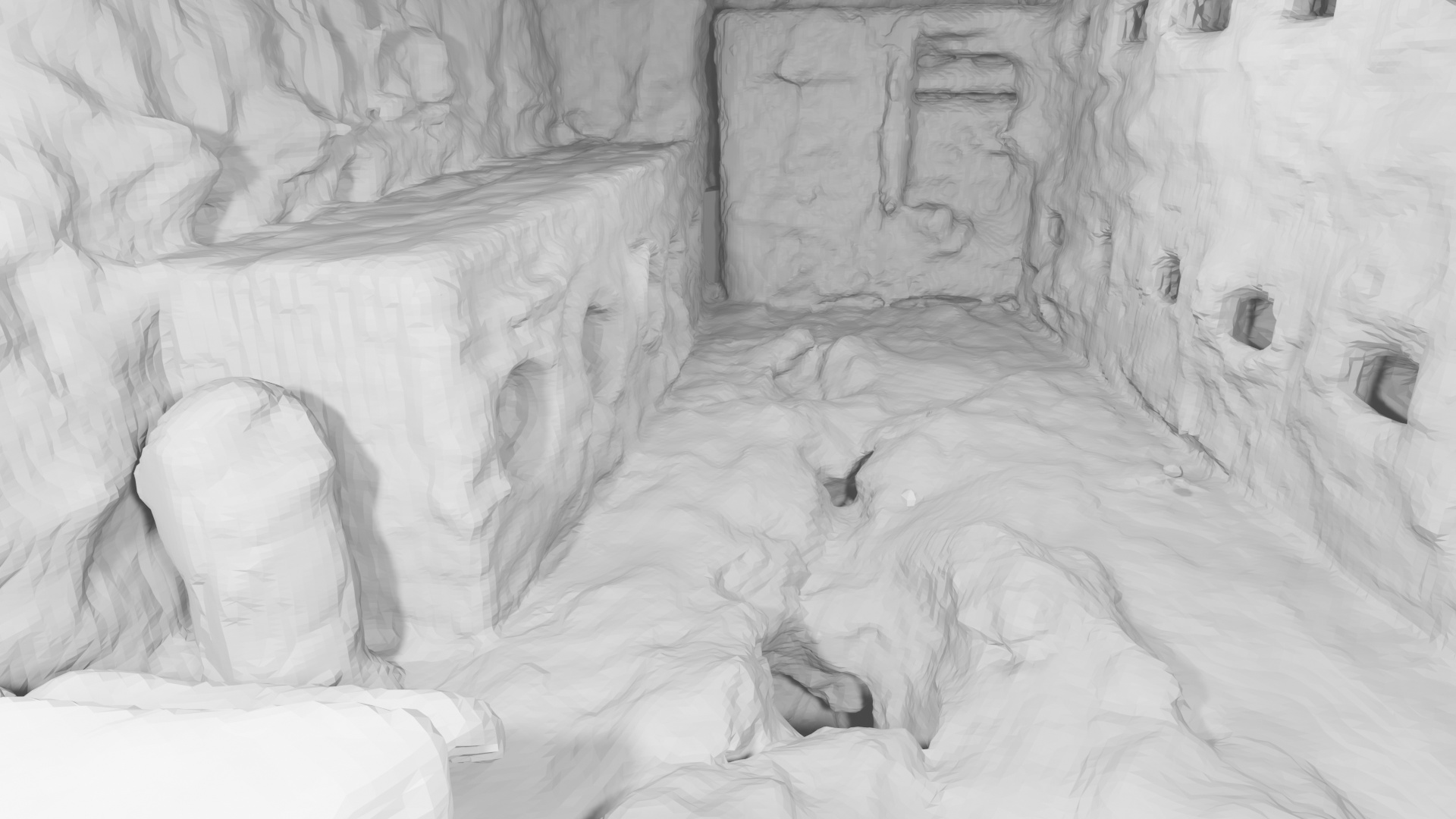} &
    \includegraphics[width=\sz\linewidth]{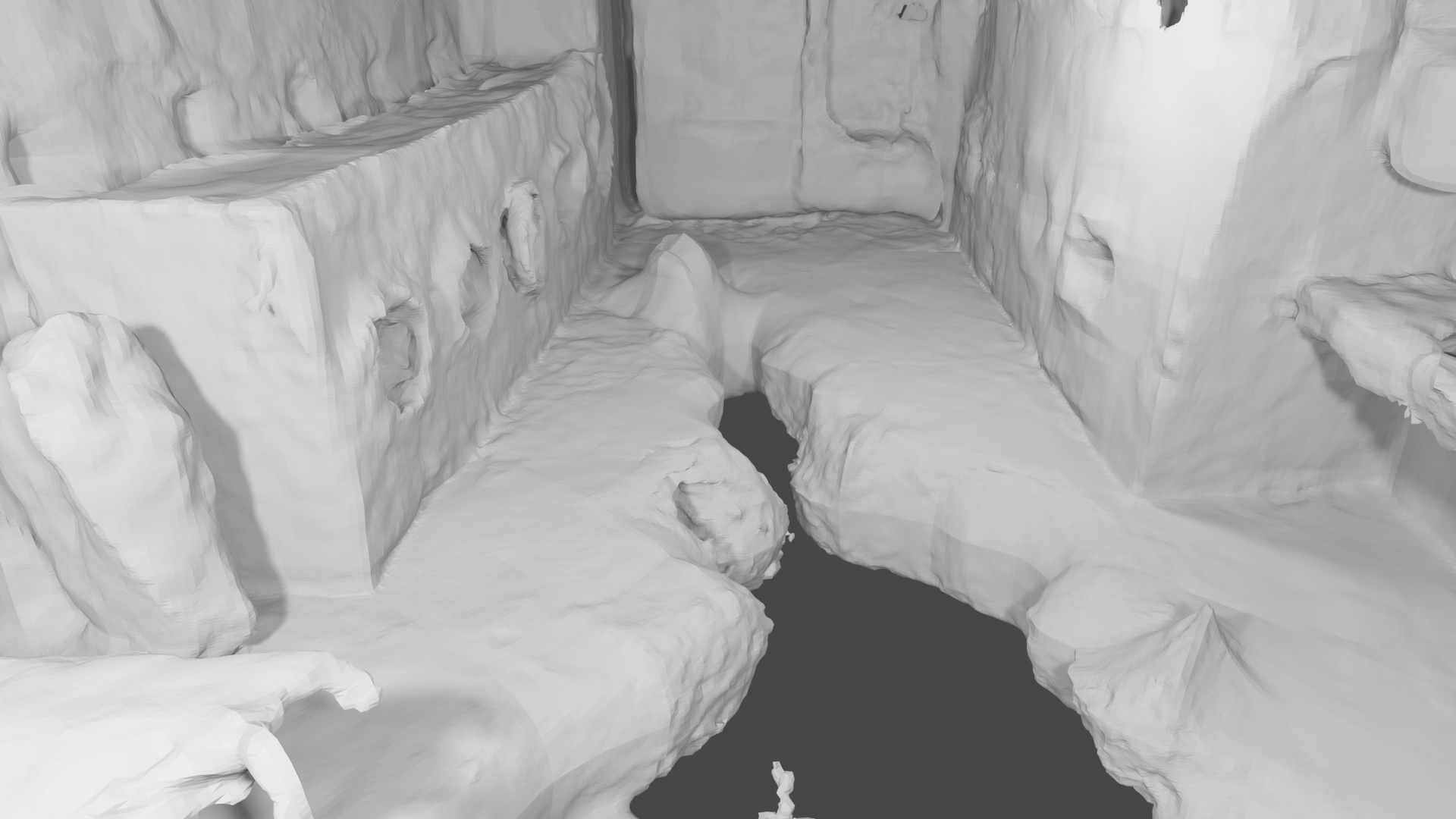} &
    \includegraphics[width=\sz\linewidth]{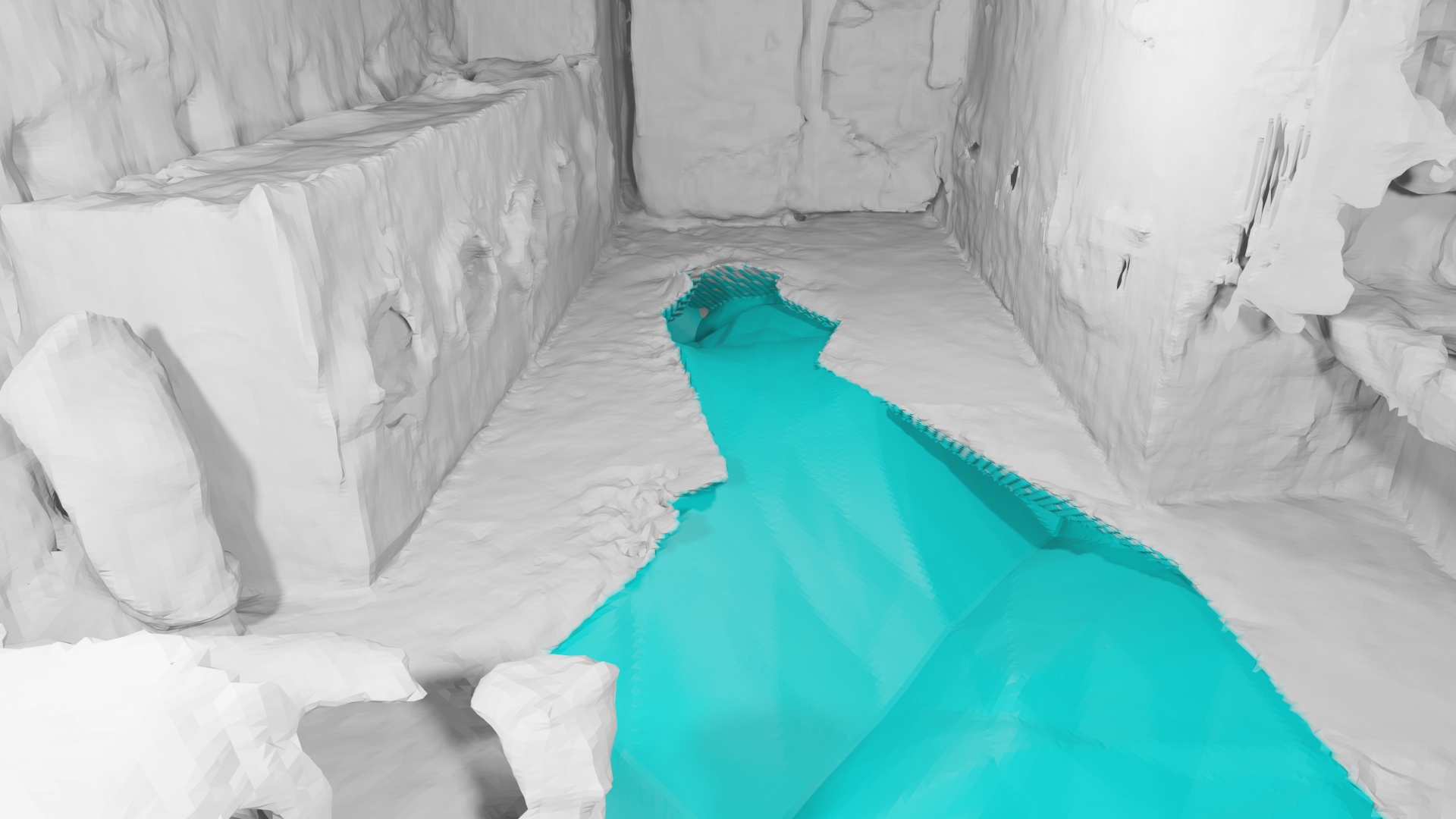} \\
  \end{tabular} 
  \caption{\textbf{Geometry Forecast and Hole Filling.} The white colored area is the region with observations, and cyan indicates the \textcolor{mycyan}{unobserved} but predicted region. 
  Thanks to the use of coarse-level scene prior, our method has better prediction capability compared to iMAP$^*$.
  This in turn also improves our tracking performance.
  }
  \label{fig:coarse}
\end{figure}

As illustrated in Fig.~\ref{fig:coarse}, we are able to complete unobserved scene regions thanks to the use of coarse-level scene prior. %
In contrast, the unseen regions reconstructed by iMAP$^*$ are very noisy since no scene prior knowledge is encoded in iMAP$^*$.

\subsection{Ablation Study}
\label{sec:ablation_study}
In this section, we investigate the choice of our hierarchical architecture and  the importance of color representation.

\begin{figure}[t!]
  \centering
  \includegraphics[width=0.87\linewidth]{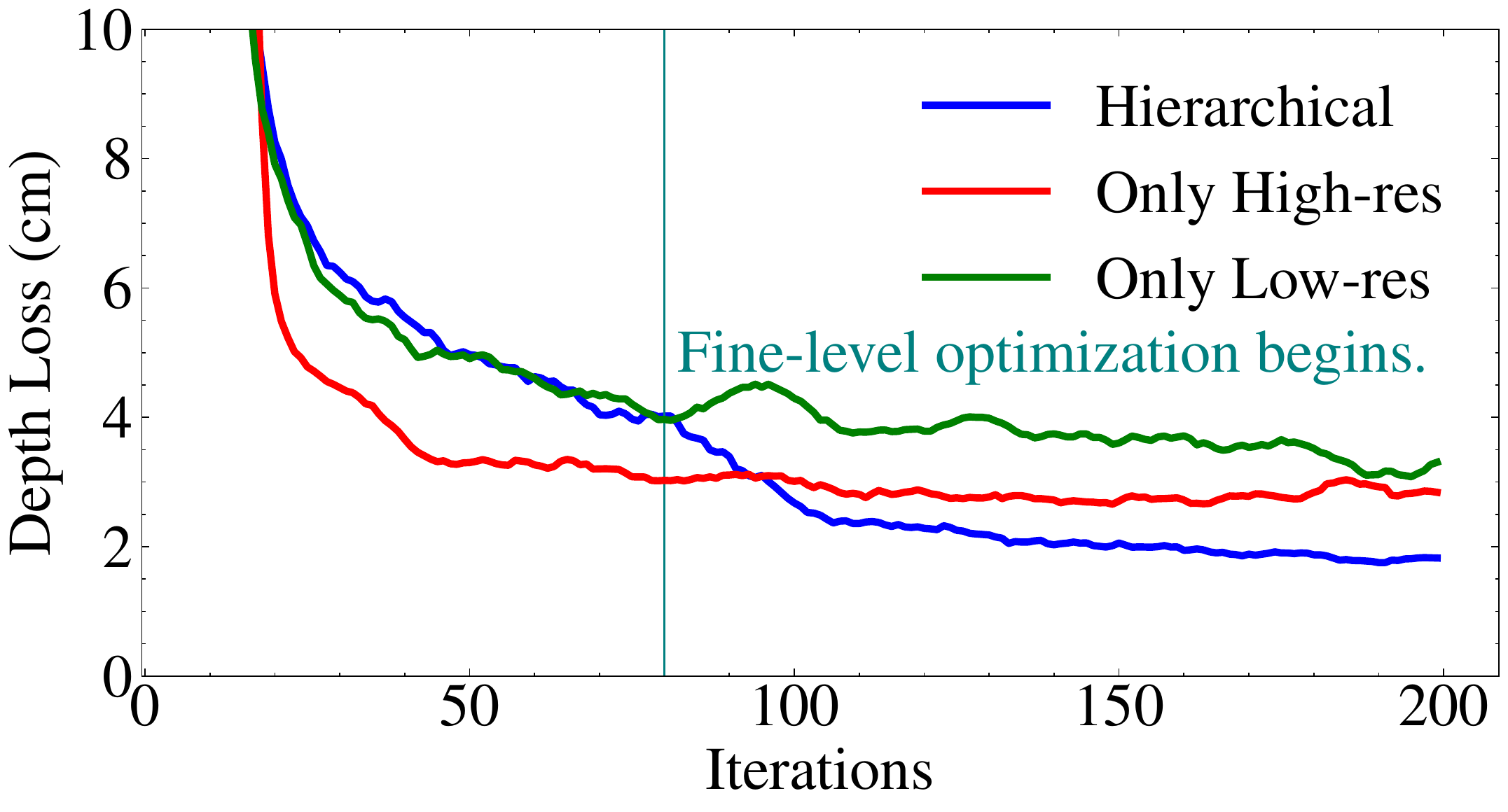}
  \caption{\textbf{Hierarchical Architecture Ablation.} 
  Geometry optimization on a single depth image on Replica~\cite{replica19arxiv} with different architectures.
  The curves are smoothed for better visualization.
  }
  \label{fig:hierarchical_plot}
\end{figure}

\boldparagraph{Hierarchical Architecture.}
\figref{fig:hierarchical_plot} compares our hierarchical architecture against: a) one feature grid with the same resolution as our fine-level representation (Only High-res); b) one feature grid with mid-level resolution (Only Low-res). 
Our hierarchical architecture can quickly add geometric details when the fine-level representation participates in the optimization, which also leads to better convergence.

\boldparagraph{Local BA.} We verify the effectiveness of local bundle adjustment on ScanNet~\cite{dai2017scannet}.
If we do not jointly optimize camera poses for $K$ keyframes together with the scene representation (w/o Local BA in~\tabref{tab:ab_color_BA}), 
the camera tracking is not only significantly less accurate, but also less robust.

\boldparagraph{Color Representation.}
In~\tabref{tab:ab_color_BA} we compare our method without the photometric loss $\cL_p$ in~\eqnref{eq:photo_loss}.
It shows that, although our estimated colors are not perfect due to the limited optimization budget and the lack of sampling points, learning such a color representation still plays an important role for accurate camera tracking.

\boldparagraph{Keyframe Selection.}
We test our method using iMAP’s keyframe selection strategy (w/ iMAP keyframes in~\tabref{tab:ab_color_BA}) where they select keyframes from the entire scene. 
This is necessary for iMAP to prevent their simple MLP from forgetting the previous geometry. 
Nevertheless, it also leads to slow convergence and inaccurate tracking.

\section{Conclusion}  \label{sec:conclusion}
We presented \ours{}, a dense visual SLAM approach that combines the advantages of neural implicit representations with the scalability of an hierarchical grid-based scene representation.
Compared to a scene representation with a single big MLP, our experiments demonstrate that our representation (tiny MLPs + multi-res feature grids) not only guarantees fine-detailed mapping and high tracking accuracy, but also faster speed and much less computation due to the benefit of local scene updates.
Besides, our network is able to fill small holes and extrapolate scene geometry into unobserved regions which in turn stabilizes the camera tracking.

\boldparagraph{Limitations.}
The predictive ability of our method is restricted to the scale of the coarse representation. 
In addition, our method does not perform loop closures, which is an interesting future direction.
Finally, although traditional methods lack some of the features, there is still a performance gap to the learning-based approaches that needs to be closed.

\boldparagraph{Acknowledgements.}
The authors thank the Max Planck ETH Center for Learning Systems (CLS) for supporting Songyou Peng. We also thank Edgar Sucar for providing additional implementation details about iMAP. Special thanks to Chi Wang for offering the data collection site. This work was partially supported by the NSFC (No.~62102356), Zhejiang Lab (2021PE0AC01). Weiwei Xu is partially supported by NSFC (No.~61732016).

\begin{table}[t!]
  \centering
  \footnotesize
  \setlength{\tabcolsep}{3pt}
    \begin{tabular}{l||ccc|c}
      \toprule
          ATE RMSE ($\downarrow$) & {w/o Local BA} & {w/o $\cL_p$} & {w/ iMAP keyframes} & Full \\
         \midrule
         Mean & 37.74 & 32.02 & 12.10 &9.63\\
         Std. & 30.97 & 21.98 & 3.38 &0.62\\
      \bottomrule
    \end{tabular}%
    \caption{\textbf{Ablation Study.} We investigate the usefulness of local BA, color representation, as well as our keyframe selection strategy. We run each scene 5 times and calculate their mean and standard deviation of ATE RMSE ($\downarrow$). We report the average values over 6 scenes in ScanNet~\cite{dai2017scannet}.
    }
    \label{tab:ab_color_BA}
\end{table}

{\small
\bibliographystyle{ieee_fullname}
\bibliography{egbib}
}

\clearpage

\setcounter{section}{0}
\setcounter{figure}{0}
\setcounter{table}{0}
\renewcommand\thesection{\Alph{section}}
\renewcommand\thetable{\Alph{table}}
\renewcommand\thefigure{\Alph{figure}}

\title{\vspace{-0.5em}-- Supplementary Material --\\NICE-SLAM: Neural Implicit Scalable Encoding for SLAM}

\author{
Zihan Zhu$^{1,2 *}$ \qquad Songyou Peng$^{2,4*}$\qquad Viktor Larsson$^{3}$ \qquad Weiwei Xu$^{1}$ \qquad Hujun Bao$^{1}$\\
Zhaopeng Cui$^{1}$$\ssymbol{2}$  \qquad Martin R. Oswald$^{2,5}$ \qquad Marc Pollefeys$^{2,6}$\vspace{0.5em}\\
$^{1}$State Key Lab of CAD\&CG, Zhejiang University\qquad $^{2}$ETH Zurich\qquad
$^{3}$Lund University\\$^{4}$MPI for Intelligent Systems, T\"ubingen \qquad
$^{5}$University of Amsterdam\qquad $^{6}$Microsoft
}

\maketitle

In the supplementary material we present the following:

\begin{itemize}
    \item Implementation details and parameters~(Section~\ref{sec:imp})
    \item Additional experiments and ablations (Section~\ref{sec:exp})
\end{itemize}

\section{Implementation Details}  \label{sec:imp}

\subsection{Frustum Feature Selection}

\label{sec:feature_selection}
The grid-based representation allows us to only optimize the geometry within the current viewing frustum while keeping the rest of the scene geometry fixed. However, naive optimization for all voxels will affect features even just slightly outside the viewing frustum because of trilinear interpolation. 
This is illustrated in Fig.~\ref{fig:interpolation_problem}. The rays \emph{A} and \emph{B} are viewing rays from the current frame and an active keyframe, respectively. Including these rays in the optimization will update the feature at \emph{X} (marked in the figure) due to trilinear interpolation. 
However, updating this feature will also affect the ray \emph{C} coming from an inactive keyframe.

To solve the problem, we propose to only update features fully inside the current viewing frustum during the optimization, see~\figref{fig:feature_selection_method}. %
In this way, it will not only preserve the previously reconstructed geometry, but also significantly reduce the number of parameters during optimization.

\subsection{Hierarchical Feature Grid Initialization}

\boldparagraph{Coarse-level Feature Grid.}%
The coarse-level feature grid is randomly initialized in all experiments.

\boldparagraph{Mid-level Feature Grid.}%
The mid-level feature grid is also randomly initialized in all experiments, except for the result shown in~\figref{fig:coarse} in the main paper, where it is initialized to free space to better visualize the predictions from the coarse-level grid. 
Empirically we find that the random initialization gives slightly better convergence compared to initializing from a fixed feature vector corresponding to the free space.

\boldparagraph{Fine-level Feature Grid.}%
The fine-level feature grid is initialized to ensure the output of the fine-level decoder $f^2$ as zero, as it is added in a residual manner onto the occupancy predicted from the mid-level features. 
This guarantees a smooth energy transition in the coarse-to-fine optimization. 
During the training of the fine-level decoder from ConvONet~\cite{Peng2020ECCV}, we add additional regularization loss to enforce that, if the fine-level feature is zero, no matter what the concatenated mid-level feature is, the output residual should always be zero. 
This regularization allows us to zero-initialize the fine-level grid at runtime.

\subsection{Justification for Design Choices.}

\boldparagraph{Why 3-level Feature Grids?} 
We show in~\figref{fig:hierarchical_plot} in the main paper that using hierarchical grids leads to better convergence compared to a single level, and we find that the current design guarantees a good balance between the quality and real-time capability / memory consumption (only 12 MB for Replica scenes).
We also conduct an ablation study on the number of levels of feature grids in~\tabref{tab:levels_rec}. It shows that the 3-level feature grid is a good balance between the reconstruction quality and computational efficiency.

\begin{figure}[!t]
  \begin{subfigure}{0.23\textwidth}
    \centering\includegraphics[width=\linewidth]{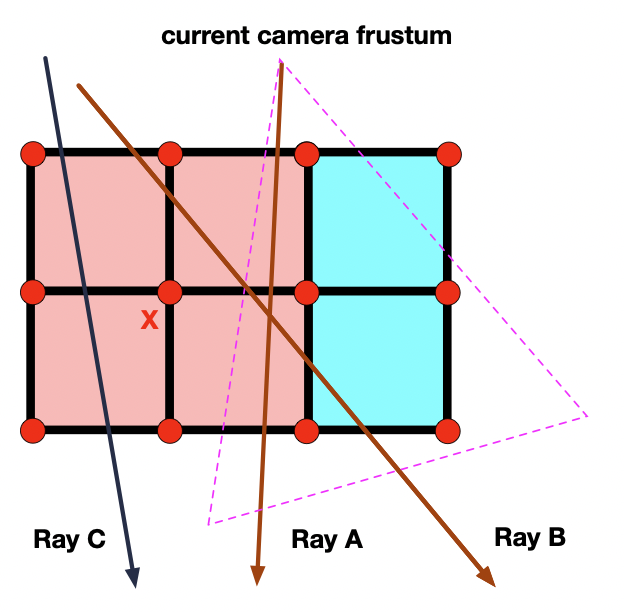}
    \caption{Interpolation problem.} \label{fig:interpolation_problem}
  \end{subfigure}%
  \hspace*{\fill}   %
  \begin{subfigure}{0.24\textwidth}
    \centering\includegraphics[width=\linewidth]{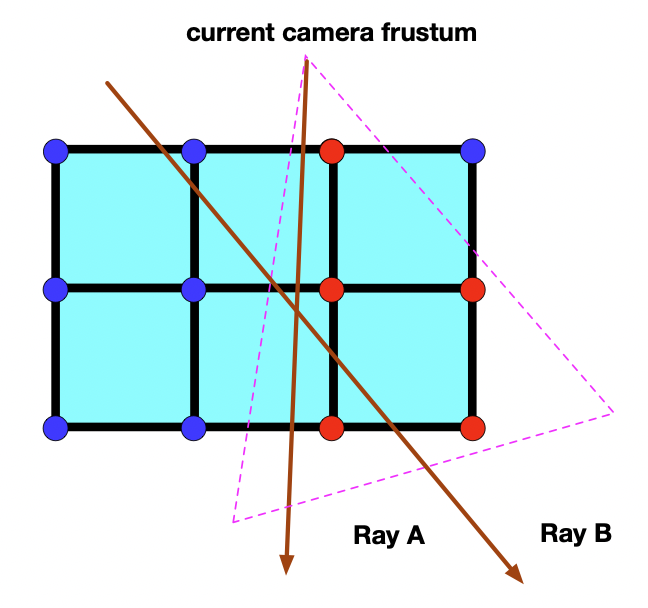}
    \caption{Feature selection.} \label{fig:feature_selection_method}
  \end{subfigure}%
\caption{2D illustration of the feature grid. The lattice points correspond to features. %
The optimized and fixed features are shown in red and blue respectively.
} 
\end{figure}

\begin{table}[!t]
  \centering
  \footnotesize
  \setlength{\tabcolsep}{0.36em}
    \begin{tabular}{lccccccccc}
      \toprule
       
        \textbf{Levels}& \textbf{2} & \textbf{3} & \textbf{4}  \\
     \midrule
          {\bf FLOPs} [$\times10^3$] $\downarrow$  & 58.45 & 104.16 & 155.95\\
     \midrule
          {\bf Depth L1} [cm] $\downarrow$ 
          & \textbf{1.86} & 1.87 & 1.96 \\
          {\bf Acc. } [cm] $\downarrow$ 
          &  2.87 & \textbf{2.78} & 3.15 \\
          {\bf Comp. } [cm] $\downarrow$ 
          &  2.76 & 2.76  & \textbf{2.40}\\
          {\bf Comp. Ratio} [$<$ 5cm \%] $\uparrow$ 
          &    91.24 & 91.37 & \textbf{93.60} \\
      \bottomrule
    \end{tabular}%
    \caption{\textbf{Ablation on the Levels of Feature Grids.} 
    Reconstruction results on Replica \texttt{room-0} with ground truth camera pose.}
    \label{tab:levels_rec}
\end{table}

\boldparagraph{Why is the Mid-level Output not a Residual to the Coarse-level Output?}
The coarse grid has a significantly larger voxel size (side of $>1$ meter) than the mid and fine levels, so updating the coarse-level feature would affect a large area. 
To ensure small local updates for efficiency, we disconnect coarse level from mid and fine levels, and only use coarse level for prediction.

\subsection{Mesh Visualization}
The reconstructed scene is represented implicitly using hierarchical feature grids. We use the marching cubes algorithm~\cite{marchingcubes} to create a mesh for the visualization purpose. 
For every observed point we predict its occupancy value using the fine-level decoder and color from the color decoder. 
For those unseen points in the predicted regions (\ie ~voxels with partial observations in the coarse grid), we predict occupancy from the coarse-decoder and set the color to \textcolor{mycyan}{cyan} for visualization as shown in~\figref{fig:hierarchical_plot} in our main paper and the supplementary video. Other points are assigned zero occupancy.  %
The same resolution is used in marching cubes for both iMAP$^*$ and \ours{}. 

\subsection{Decoder Pretraining}
We use the Synthetic Indoor Scene Dataset provided in ConvONet~\cite{Peng2020ECCV} to pre-train the encoder-decoder. Furthermore, we use the Point Cloud Encoder instead of the Voxel Encoder.
All levels are trained with room\_grid64 setting in ConvONet~\cite{Peng2020ECCV}. The feature dimension for all the feature grids is 32. As for hyperparameters used for the pretraining process, we follow the same setting  as ConvONet~\cite{Peng2020ECCV}. 

\subsection{Hyperparameters}
Here we report detailed hyperparameters of online tracking and mapping used for both \ours{} and iMAP$^*$. We perform tracking for every frame and optimize the geometry every fifth frame, except for TUM RGB-D where we optimize the geometry every frame. All parameters are tuned to keep a good balance between the accuracy and the efficiency.

\boldparagraph{\ours{}.}%
For scene geometry optimization, we use a maximum of 60 iterations for all datasets. In terms of tracking, we use 10 iterations for small-scale synthetic datasets (Replica and Co-Fusion). For the large-scale real datasets including ScanNet and our self-captured scene, we use 50 iterations for tracking. For TUM RGB-D dataset we use 200 iterations.

The learning rate for tracking on Replica\cite{replica19arxiv},  TUM RGB-D~\cite{sturm2012benchmark}, ScanNet~\cite{dai2017scannet}, Self-captured, and Co-Fusion~\cite{runz2017co} are $1e{-3}$, $1e{-2}$, $5e{-4}$, $3e{-3}$, $1e{-3}$ respectively. The learning rate for optimizing the coarse-level is $1e{-3}$, for mid-level is $1e{-1}$, for fine- and color-level is $5e{-3}$. The learning rate for selected keyframes' camera parameters during the mapping is $1e{-3}$, except for Co-Fusion where we set the learning rate to 0.

\begin{table}[!t]
  \centering
  \footnotesize
  \setlength{\tabcolsep}{0.6em}
    \begin{tabular}{lccccccc}
      \toprule
       
        \textbf{Mapping Iterations}& \textbf{15} & \textbf{30} & \textbf{60} & \textbf{120} & \textbf{240} \\
     \midrule
          {\bf Depth L1} [cm] $\downarrow$ 
          & 2.31 & 2.03 & 1.87 & 1.74  & 1.59 \\
          {\bf Acc. } [cm] $\downarrow$ 
          &  2.90 & 2.84 & 2.78 & 2.80  & 2.78 \\
          {\bf Comp. } [cm] $\downarrow$ 
          &  3.14  & 2.91 & 2.76 & 2.65  & 2.50 \\
          {\bf Comp. Ratio} [$<$ 5cm \%] $\uparrow$ 
          &  89.15 & 90.55 & 91.37 & 91.94 & 92.76 \\
      \bottomrule
    \end{tabular}%
    \caption{\textbf{Ablation on Mapping Iterations.} Reconstruction results on Replica \texttt{room-0} with ground truth camera poses.}
    \label{tab:map_iter_rec}
\end{table}

\boldparagraph{iMAP$^*$.}
For all datasets except TUM RGB-D~\cite{sturm2012benchmark}, we use 50 iterations for tracking and 300 iterations for joint optimization. For TUM RGB-D~\cite{sturm2012benchmark}, we use 200 and 300 iterations respectively. %
The learning rate for tracking on Replica\cite{replica19arxiv}, TUM RGB-D~\cite{sturm2012benchmark}, ScanNet~\cite{dai2017scannet}, Self-captured, and Co-Fusion~\cite{runz2017co} are $5e{-4}$, $5e{-3}$, $2e{-3}$, $1e{-3}$, $5e{-4}$ respectively. The learning rate for joint optimization is $2e{-4}$.

\section{Additional Experiments} \label{sec:exp}

\begin{figure}[!t]
  \centering
  \small
  \setlength{\tabcolsep}{0.3em}
  \begin{tabular}
  {
  >{\centering}m{0.01\textwidth}
    >{\centering}m{0.2\textwidth}
  >{\centering\arraybackslash}m{0.2\textwidth}}
    
    & \makecell{w/o Frame Loss} & \makecell{w/ Frame Loss} \\
    \rotatebox{90}{iMAP$^*$} & \includegraphics[width=\linewidth]{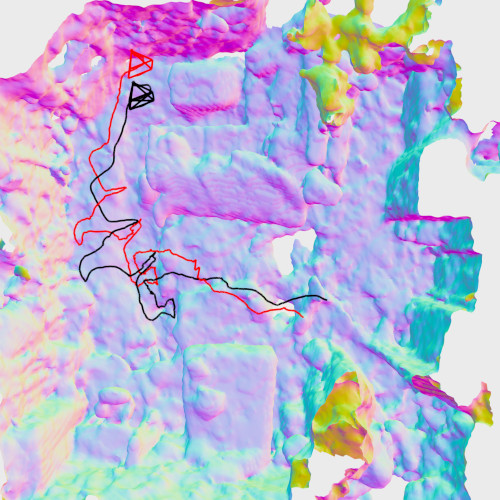} &
    \includegraphics[width=\linewidth]{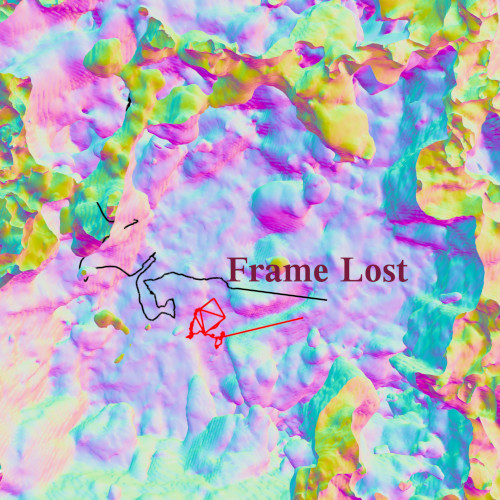} \\
    \rotatebox{90}{\ours{} w/o coarse-level} & \includegraphics[width=\linewidth]{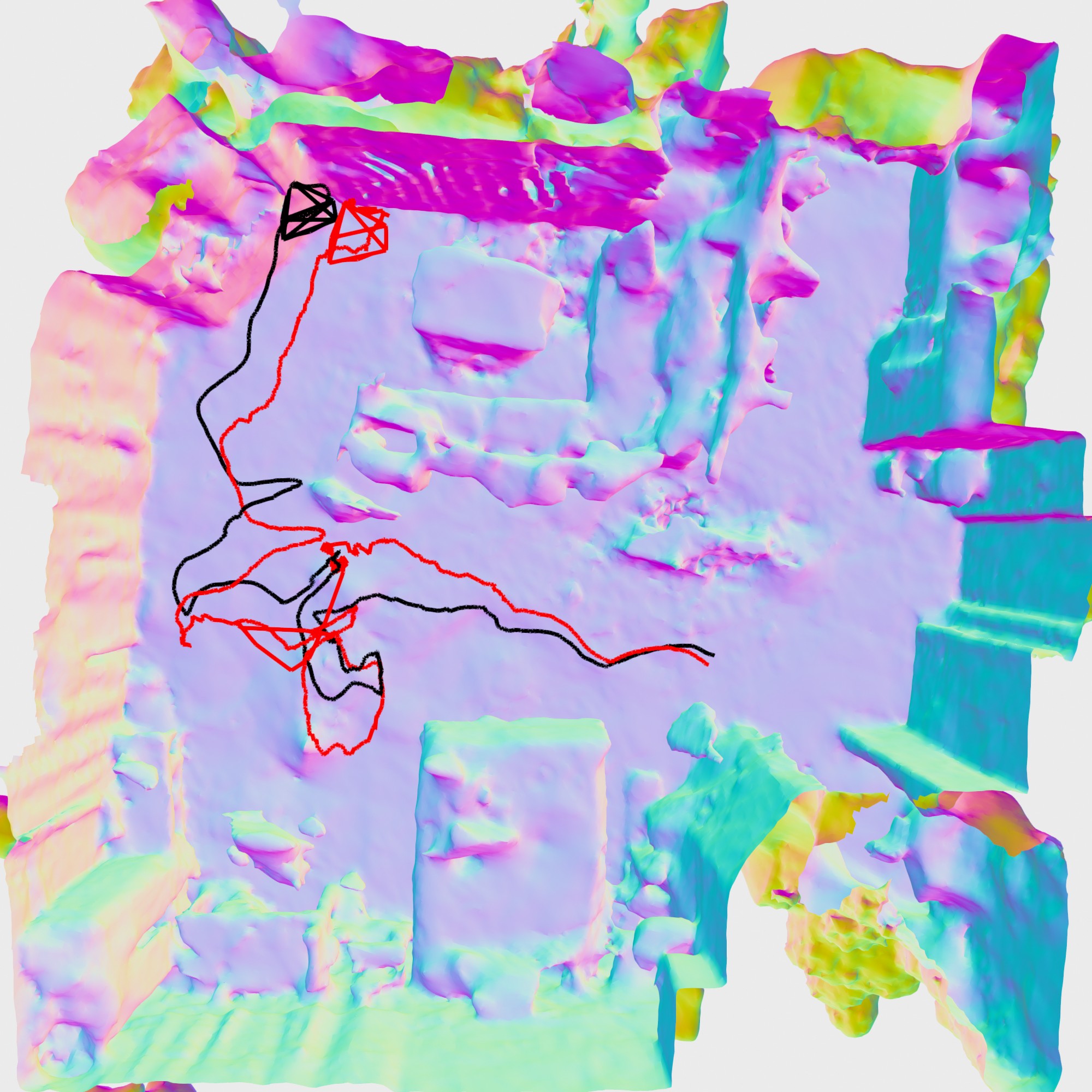} &
    \includegraphics[width=\linewidth]{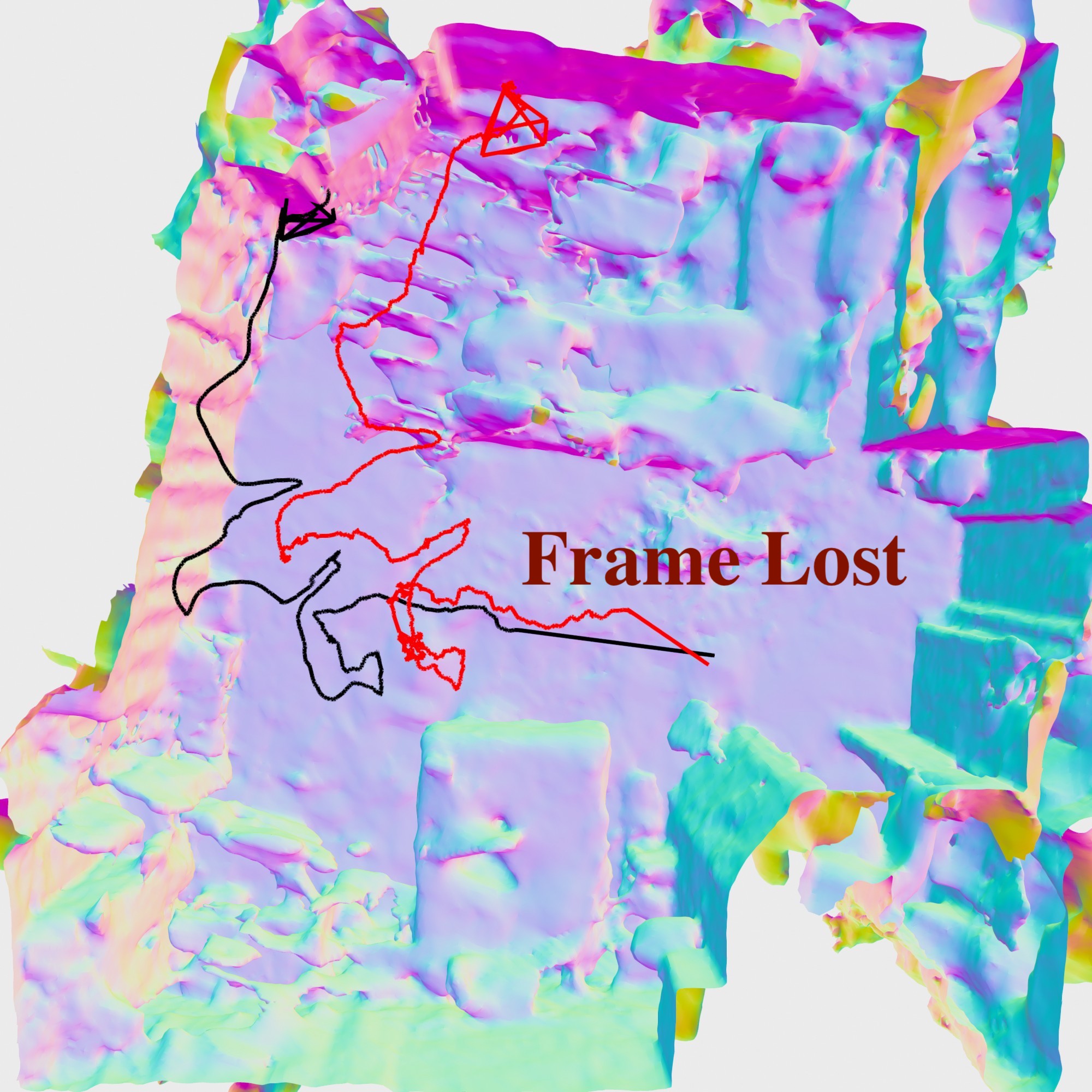} \\
    \rotatebox{90}{\ours{}} & \includegraphics[width=\linewidth]{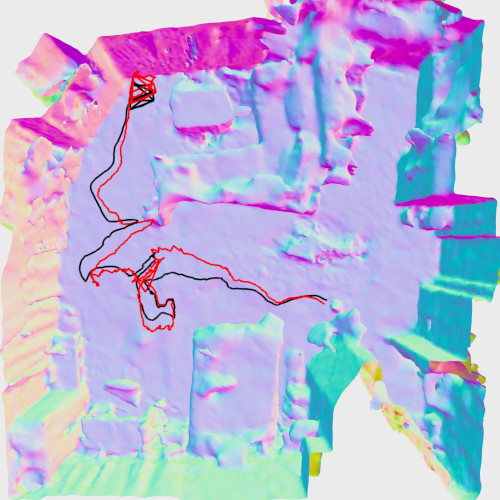} &
    \includegraphics[width=\linewidth]{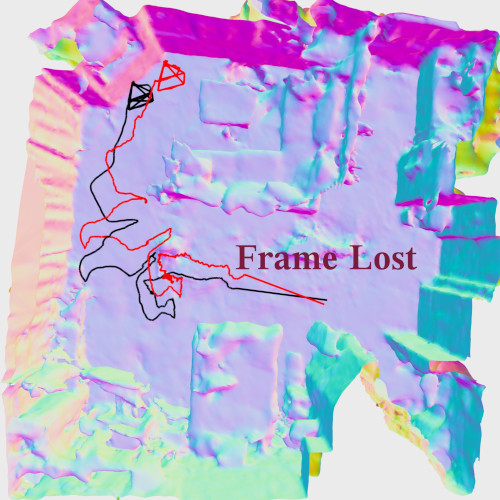} \\
   
  \end{tabular} 
  \caption{\textbf{Robustness to Frame Loss.} We show the results at frame 2100 after frame loss at frame 2000. The black trajectory is the ground truth from ScanNet~\cite{dai2017scannet}, and the \red{red} trajectory indicates tracking results. The missing frames corresponds to the straight line in the middle.} 
  \label{fig:frame_lost}
\end{figure}

\subsection{Frame Loss Robustness}%
We simulate extreme frame loss on ScanNet \texttt{scene0000\_00} by skipping 100 frames from frame ID 2001 to 2100. 
As visualized in~\figref{fig:frame_lost}, iMAP$^*$ struggles to recover camera poses and scene geometry, even given 1500 iterations. 
In contrast, our \ours{} is able to recover the camera pose using only 300 iterations. 
This is due to the use of coarse-level geometric representation which improves the prediction capability.

\subsection{Number of Mapping and Tracking Iterations}
\begin{figure}[!t]
  \centering
  \begin{tabular}{c}
    \includegraphics[width=0.9\linewidth]{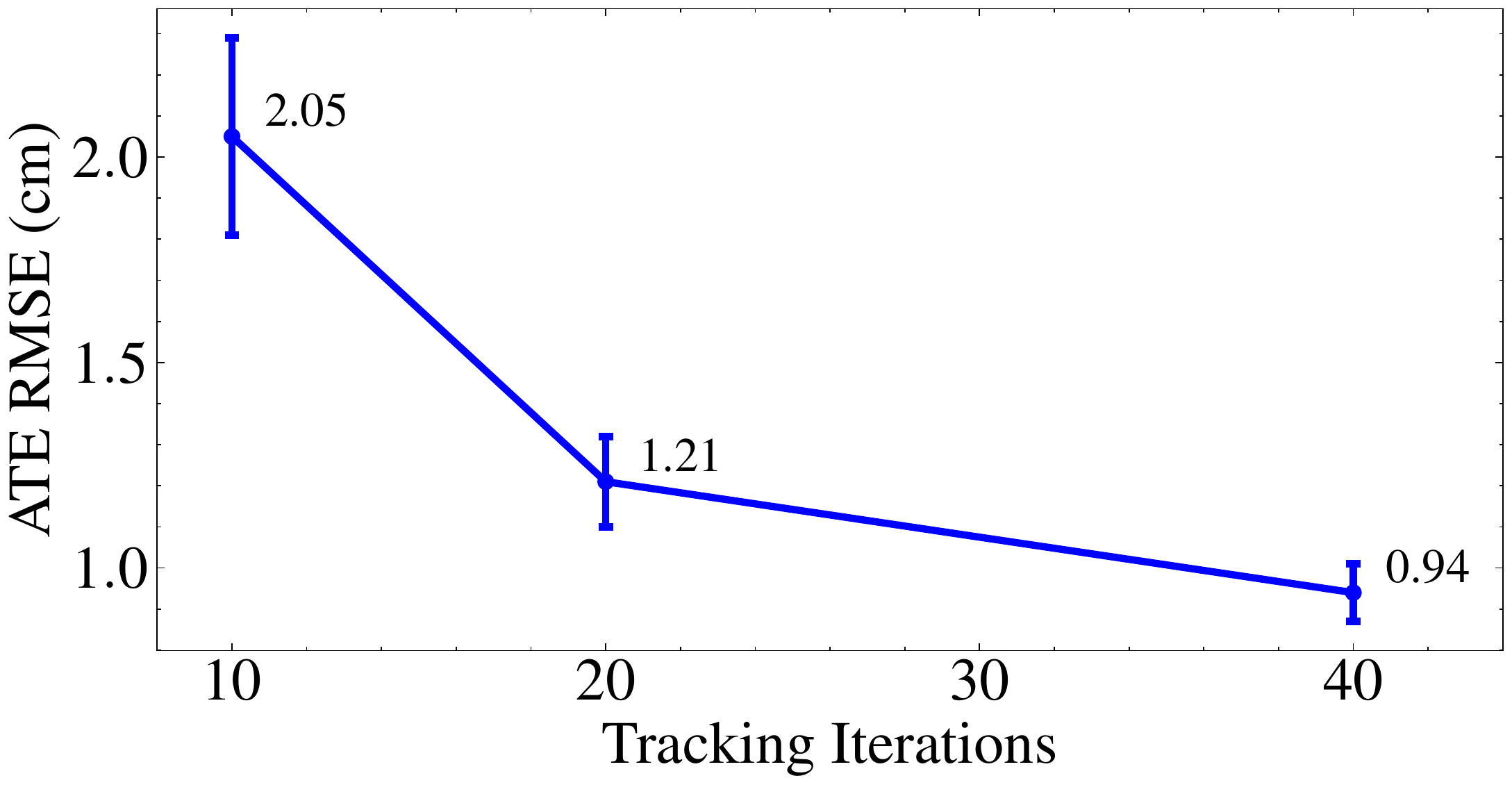}\\
    \includegraphics[width=0.9\linewidth]{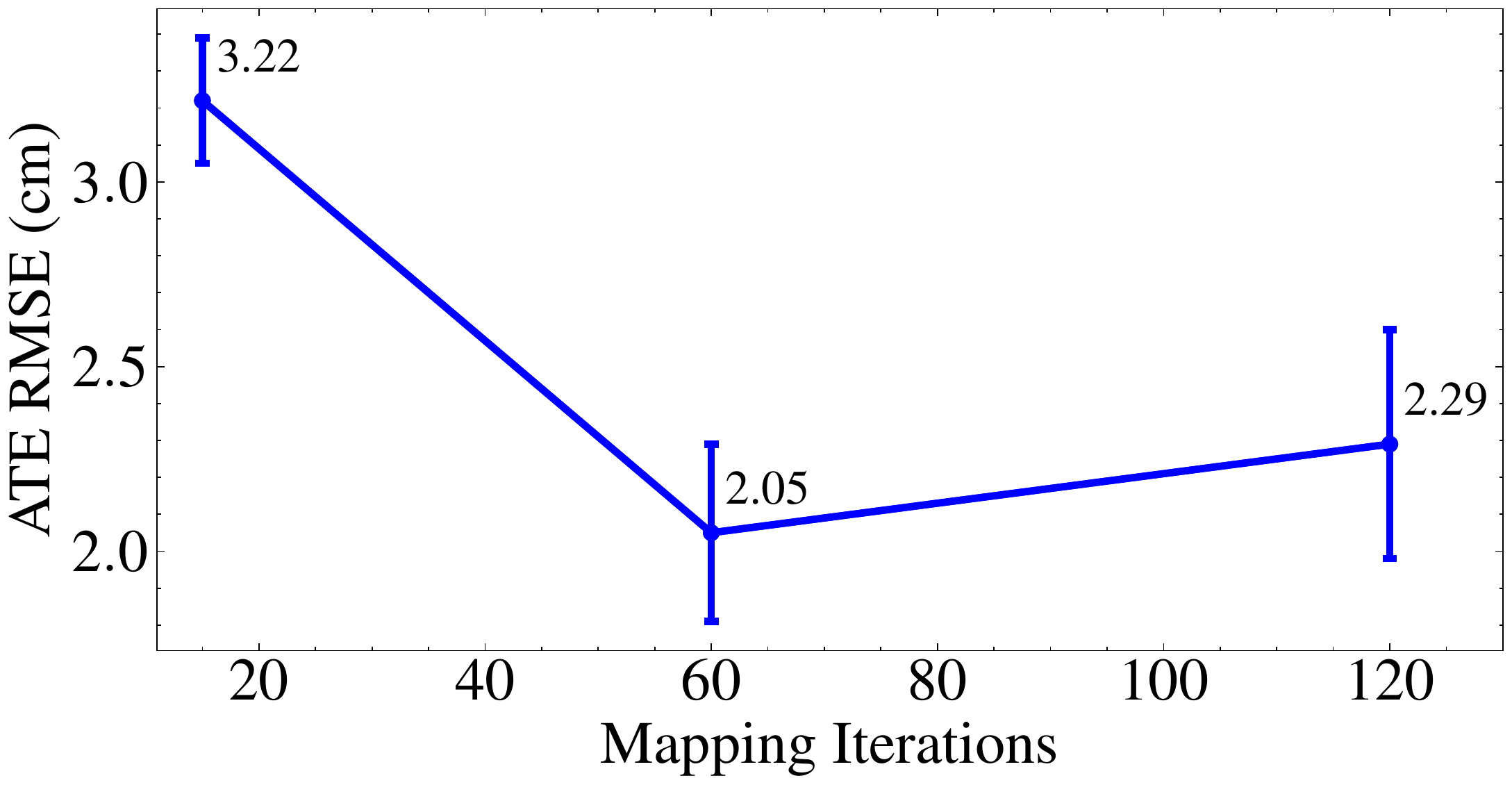}\\
  \end{tabular}
  \caption{
    \textbf{Ablation on the tracking performance.} ATE RMSE (cm) is used as the metric.
  }
  \label{fig:iter_ate}
\end{figure}

We show in \figref{fig:iter_ate} how the number of tracking and mapping iterations affects the tracking performance. We also give ground truth camera pose and evaluate reconstruction with different mapping iterations in \tabref{tab:map_iter_rec}.

\subsection{Frustum Feature Selection}
\begin{figure*}[ht]
  \centering
  \small
  \setlength{\tabcolsep}{0.2em}
  \begin{tabular}
  {
  >{\centering}m{0.07\textwidth}
  >{\centering}m{0.17\textwidth}
  >{\centering}m{0.17\textwidth}
  >{\centering}m{0.17\textwidth}
  >{\centering}m{0.17\textwidth}
  >{\centering\arraybackslash}m{0.17\textwidth}}
    Frame & {\tt 1500} & {\tt 1600} & {\tt 1700} & {\tt 1800} & {\tt 1900}\\
    
    \makecell{Ours w/} &  \includegraphics[width=\linewidth]{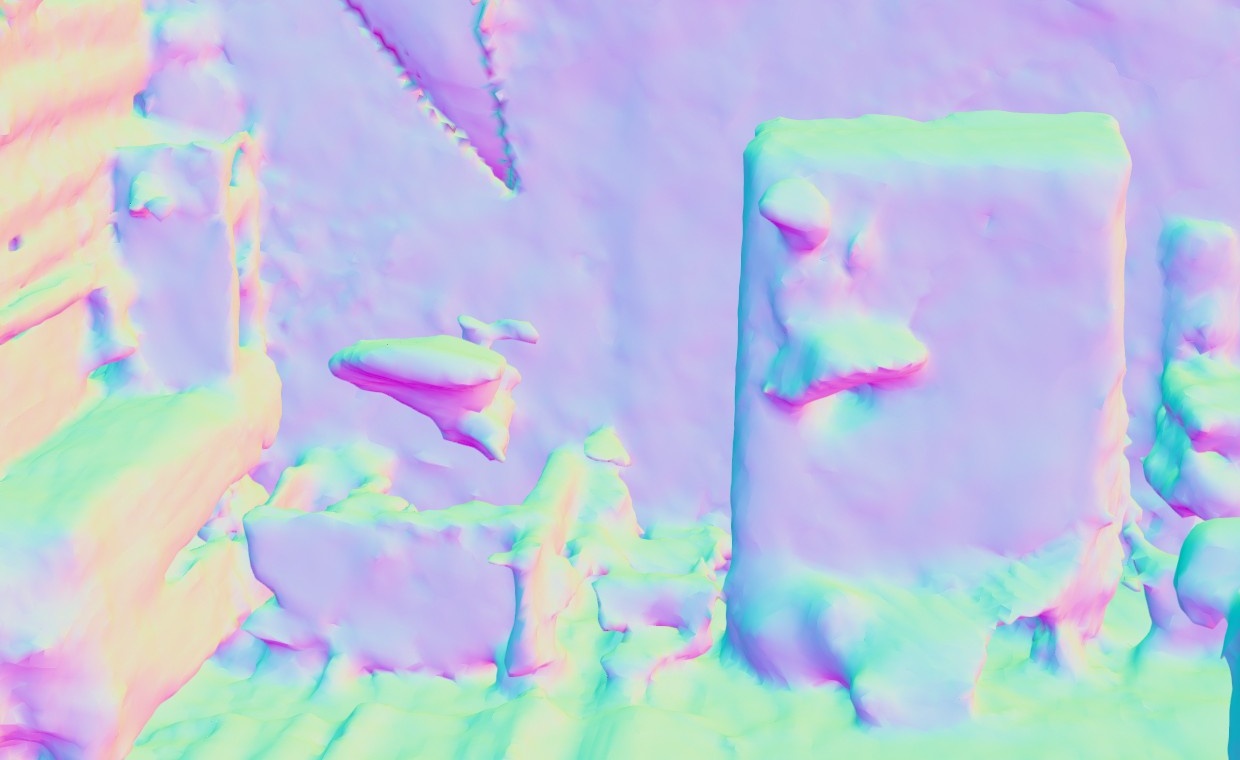} &
    \includegraphics[width=\linewidth]{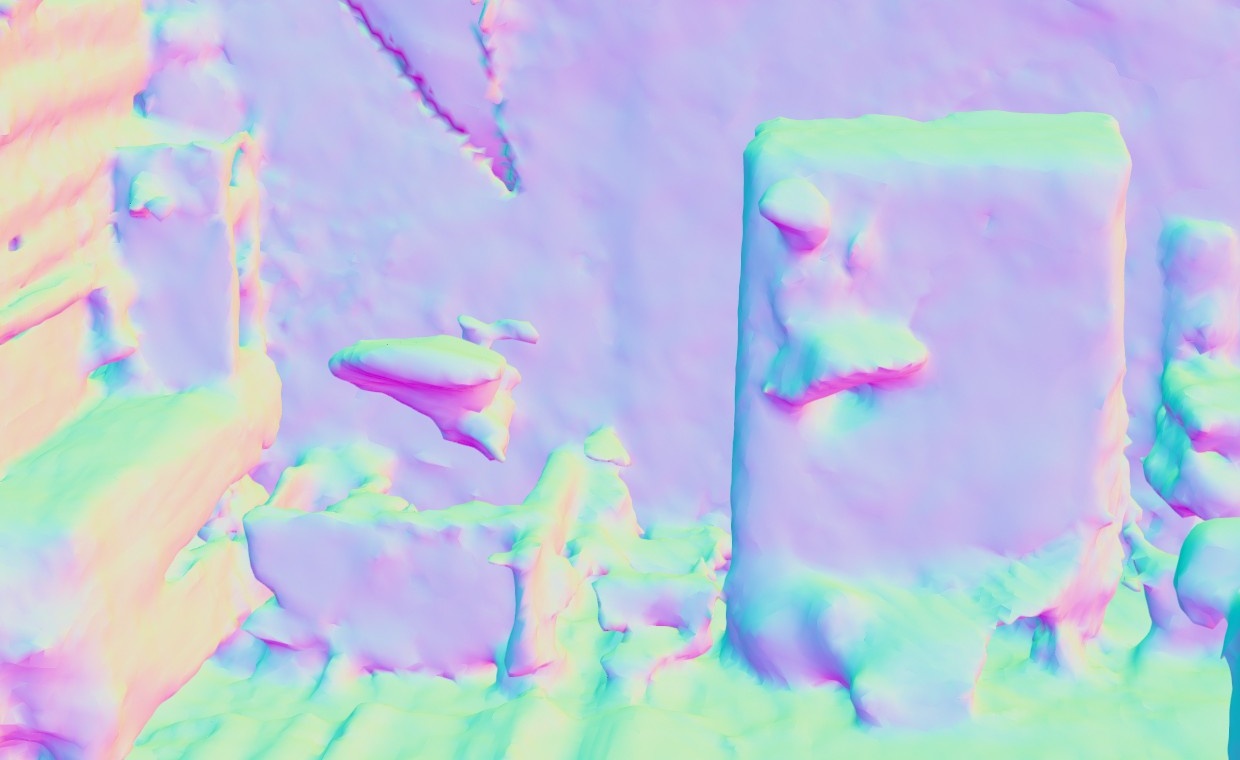} &
    \includegraphics[width=\linewidth]{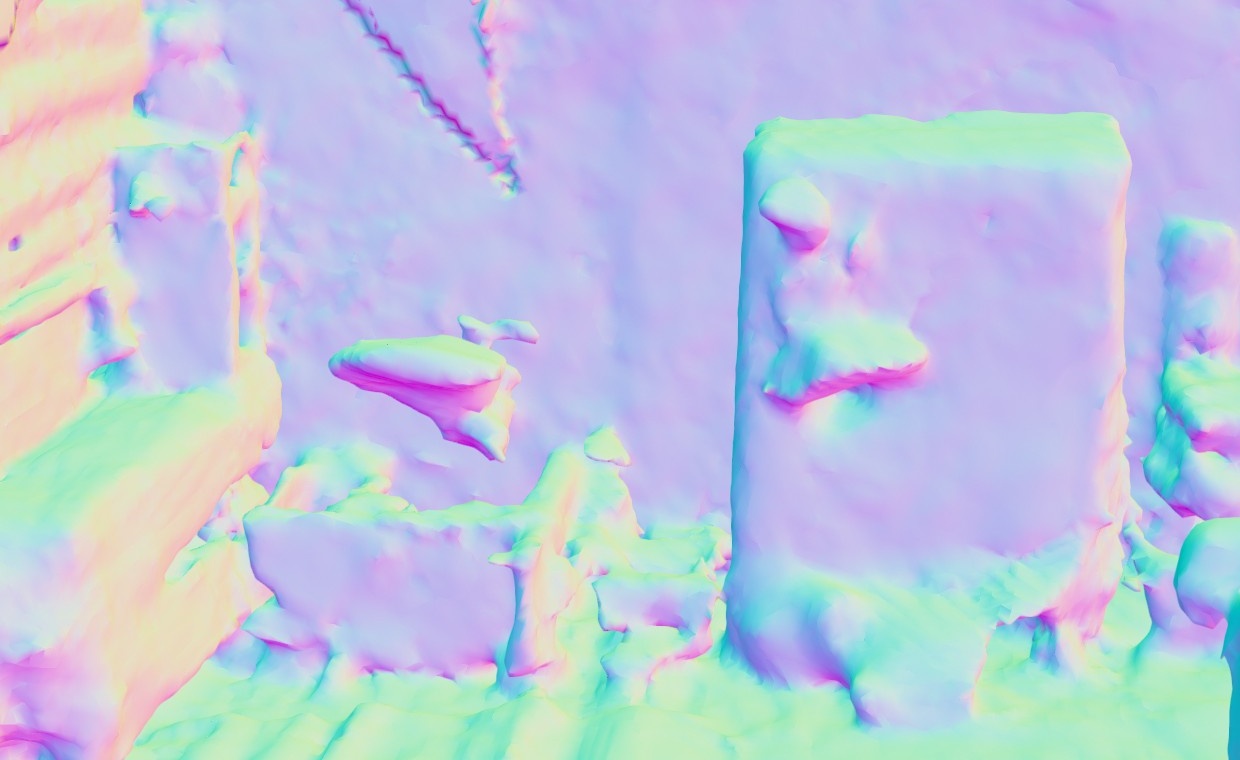} &
    \includegraphics[width=\linewidth]{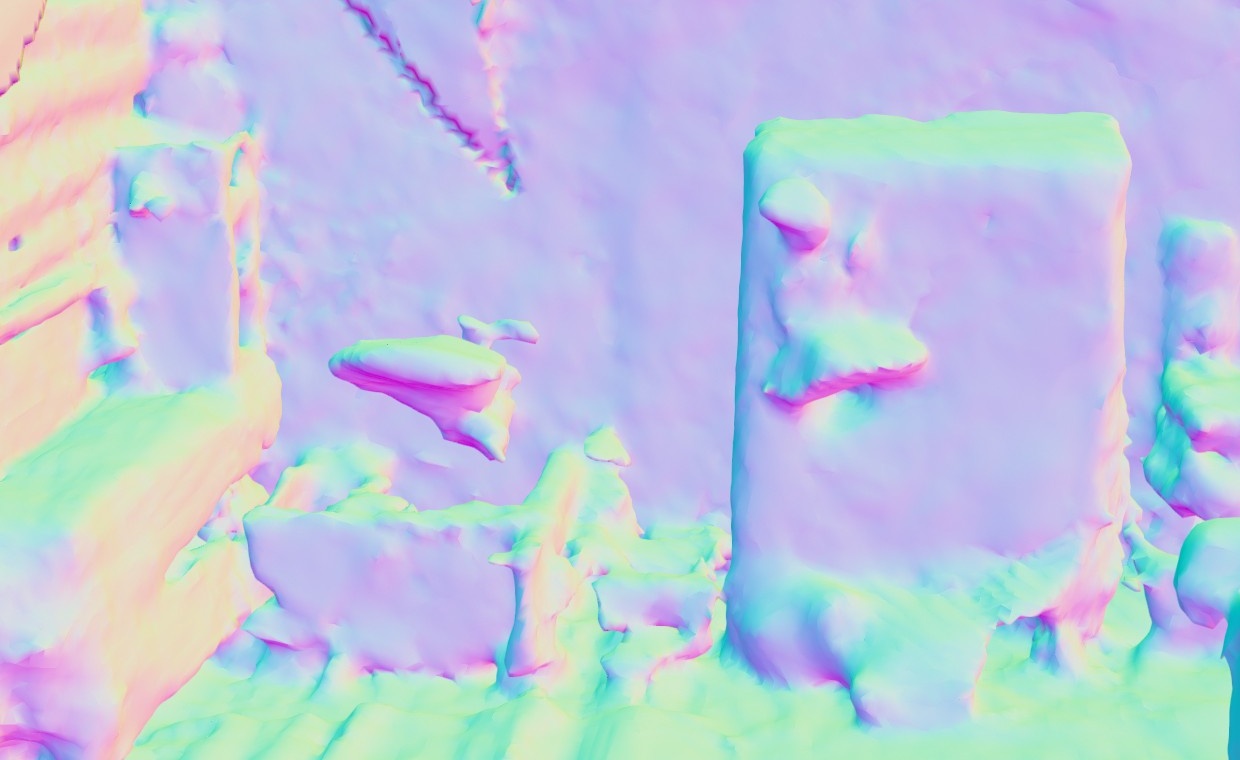} &
    \includegraphics[width=\linewidth]{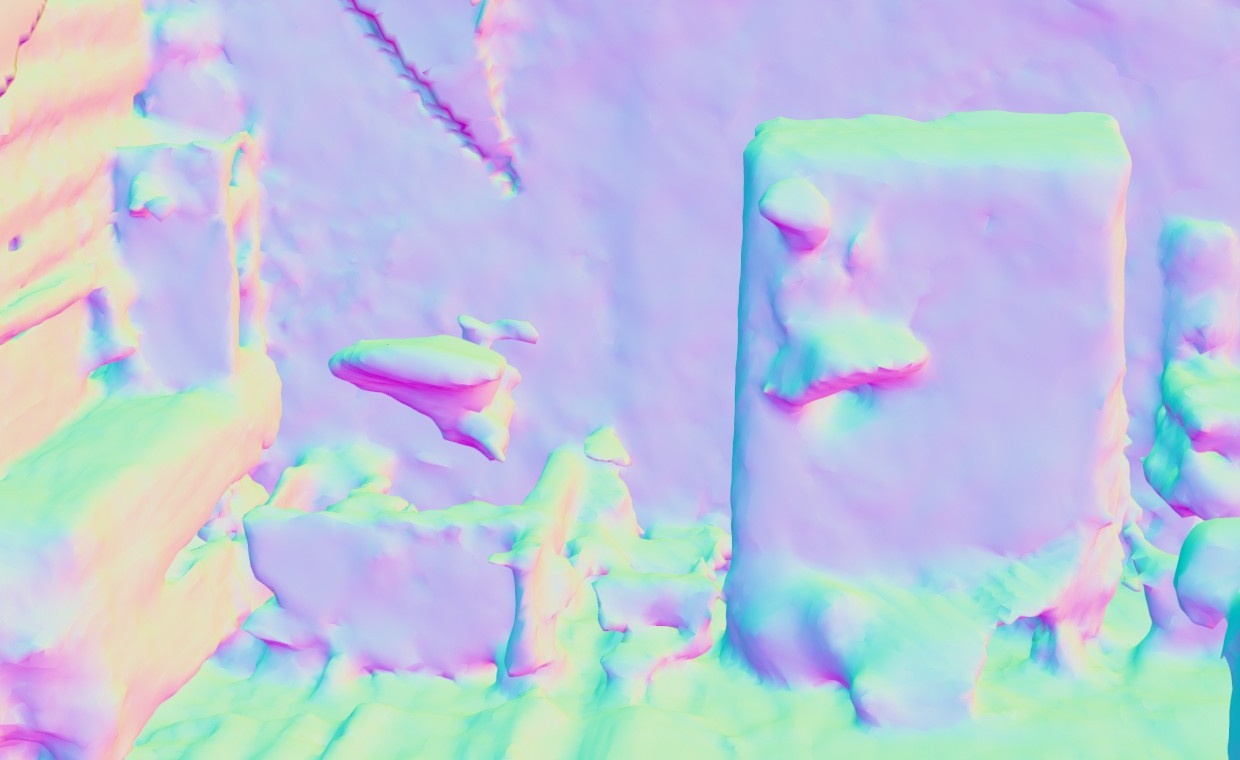} \\
    \makecell{Ours w/o} &
    \includegraphics[width=\linewidth]{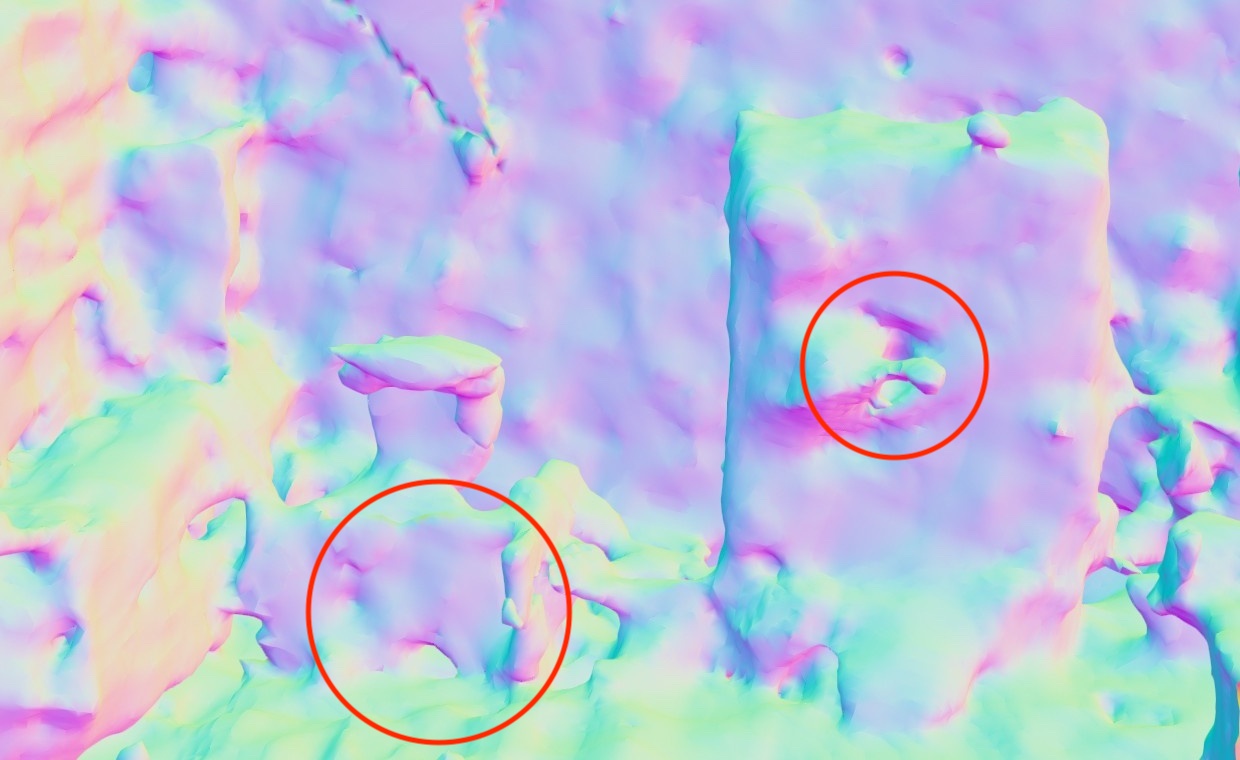} &
    \includegraphics[width=\linewidth]{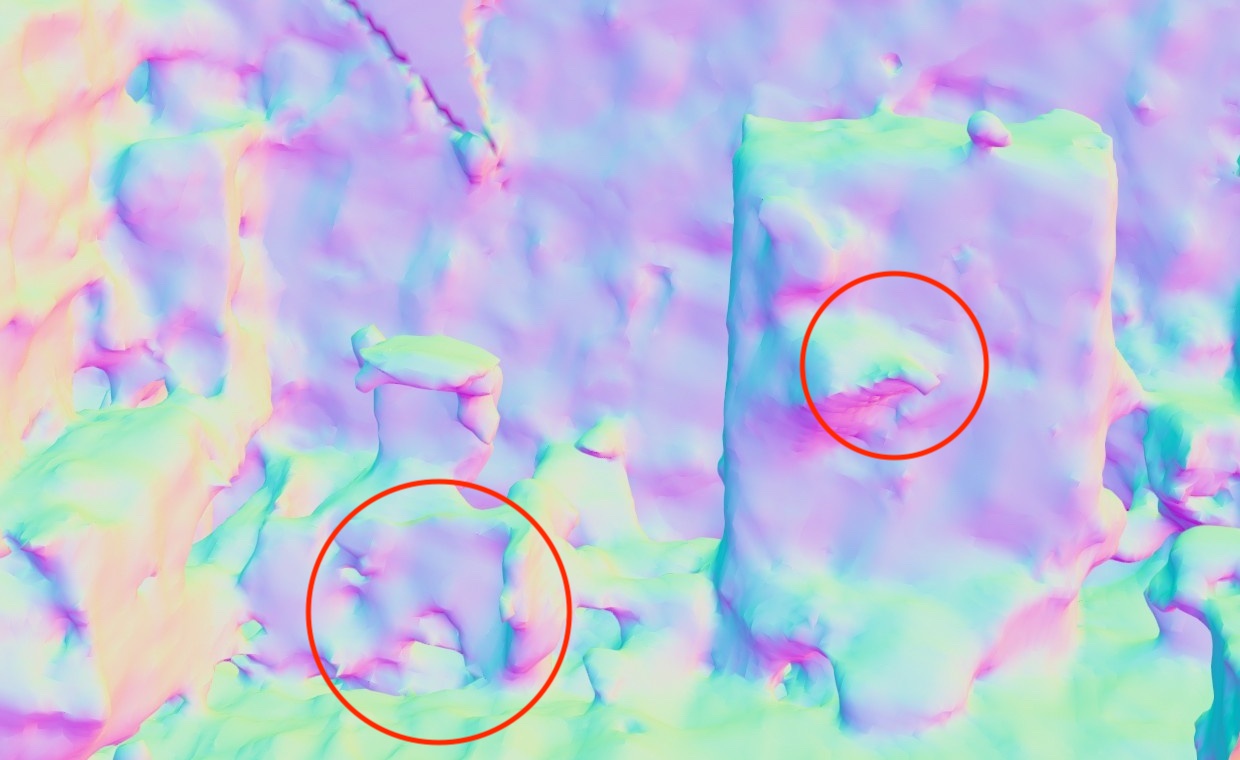} &
    \includegraphics[width=\linewidth]{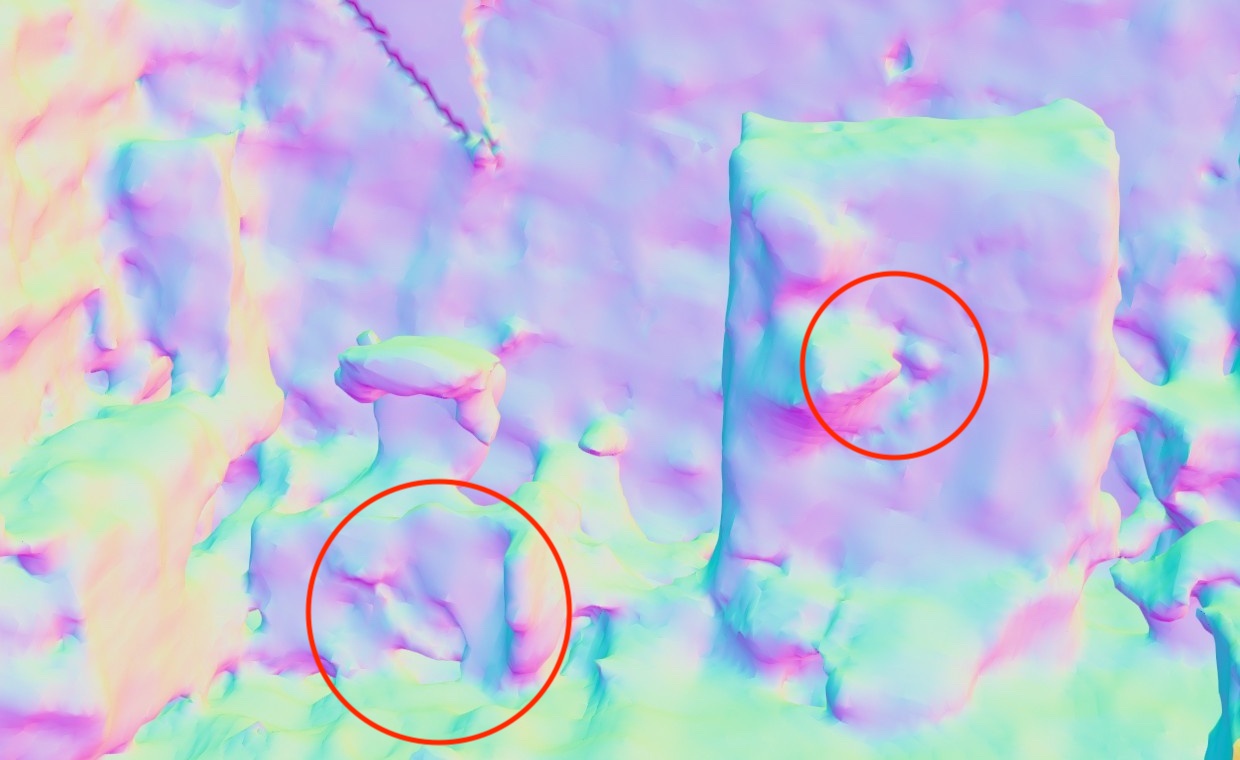} &
    \includegraphics[width=\linewidth]{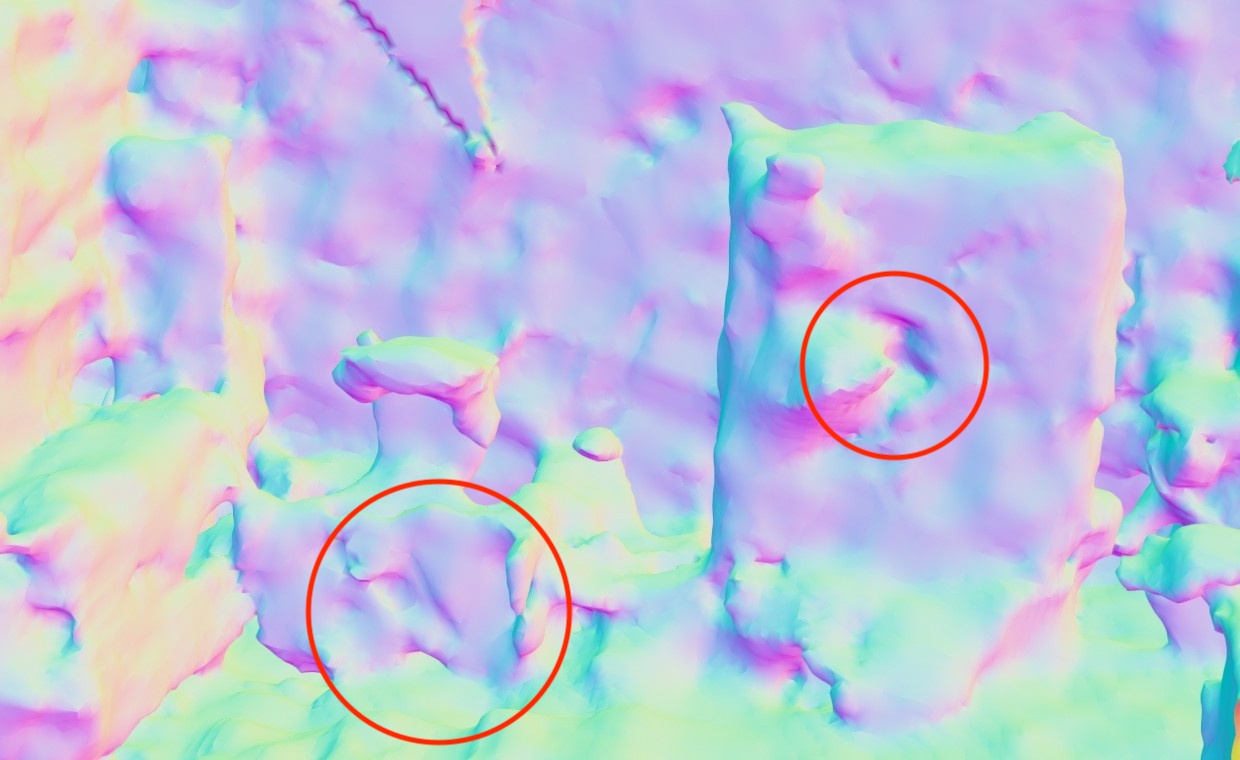} &
    \includegraphics[width=\linewidth]{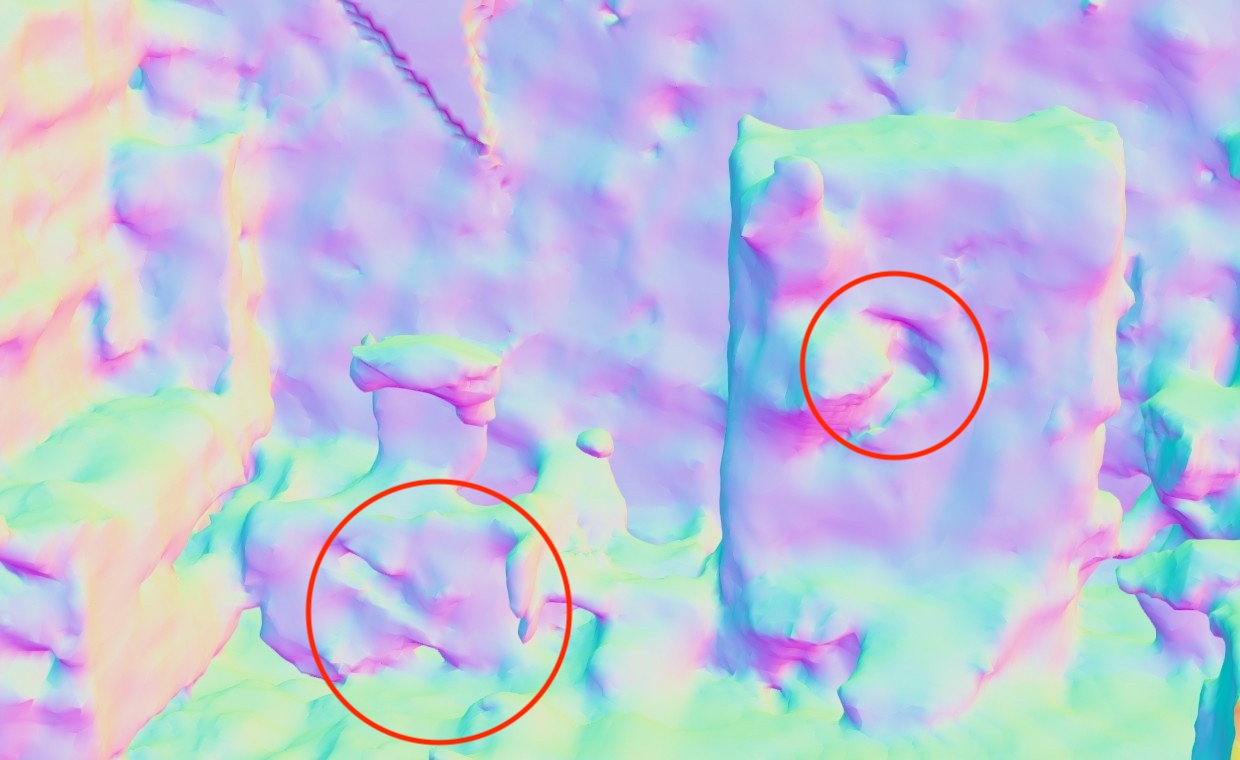} \\
  \end{tabular} 
  \caption{\textbf{Ablation on Frustum Feature Selection.} We show our method with and without the frustum feature selection run on sequence scene0000\_00 in the ScanNet datasets. During these frames the camera is scanning other parts of the scene. The cutout shown in the figure is part of the previously reconstructed geometry and should remain constant. The mesh is visualized with the vertex normal.}
  \label{fig:feature_selection}
   \vspace{10pt}
\end{figure*}

To highlight the importance of current frustum feature selection (see Section~\ref{sec:feature_selection}), we run our system with and without the selection process. The results are shown in Fig.~\ref{fig:feature_selection}. Without fixing the border features, significant artifacts appear in the reconstruction (\figref{fig:interpolation_problem}).

\subsection{More Results on Replica Dataset~\cite{replica19arxiv}}
Here we provide the detailed results for all Replica scenes.
\tabref{tab:replica_per_scene} shows the quantitative comparison when considering the average metric values for 5 consecutive runs, and only evaluate without unseen regions that are outside all camera's viewing frustums.
What is more, as done in~\cite{imap} we also report the best metrics in 5 consecutive runs
under all regions in \tabref{tab:replica_prev_per_scene}.
As can be noticed, our iMAP re-implementation iMAP$^*$ has similar performance over the original iMAP.

\begin{figure*}[htbp]
  \centering
  \footnotesize
  \setlength{\tabcolsep}{1.5pt}
  \newcommand{\sz}{0.18}  %
  \newcommand{\swa}{0.23}
  \newcommand{\swb}{0.23}
  \newcommand{\swc}{0.23}
  \newcommand{\swd}{0.23}
  \begin{tabular}{lcccc}

    & {\tt Color Ambient} & {\tt Color Shadow} & {\tt Grey} & {\tt Normal} \\

    \makecell{\rotatebox{90}{iMAP$^*$~\cite{imap}}} &
    \makecell{\includegraphics[width=\swa\linewidth,height=\sz\linewidth]{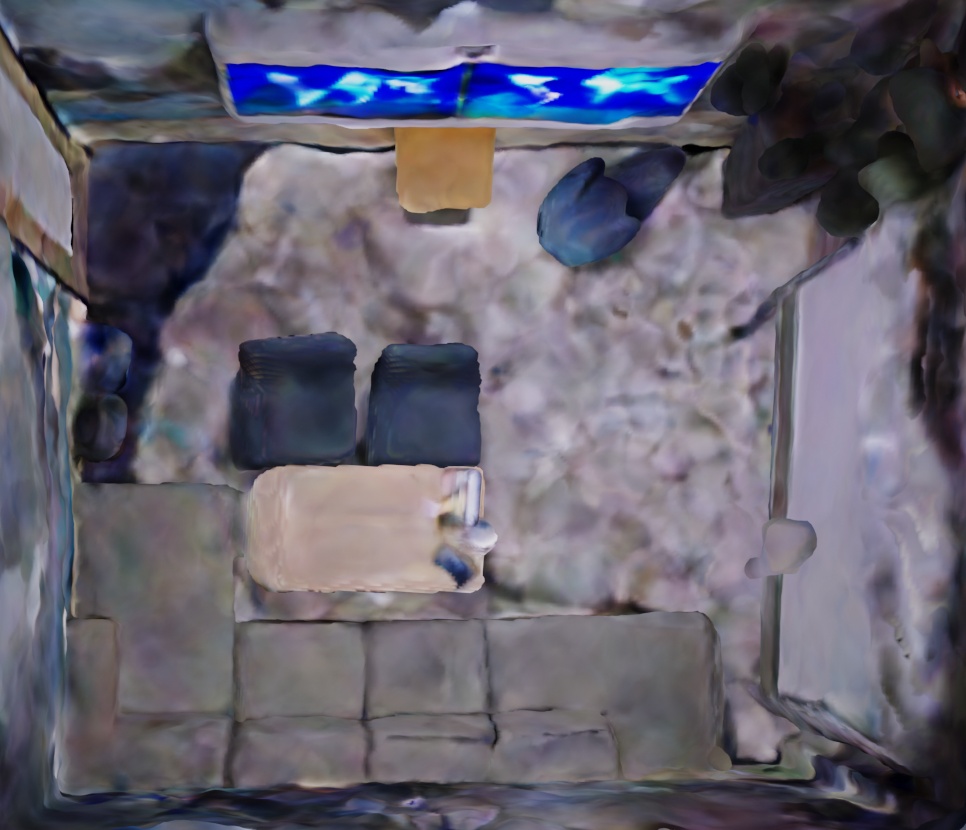}} & 
    \makecell{\includegraphics[width=\swb\linewidth,height=\sz\linewidth]{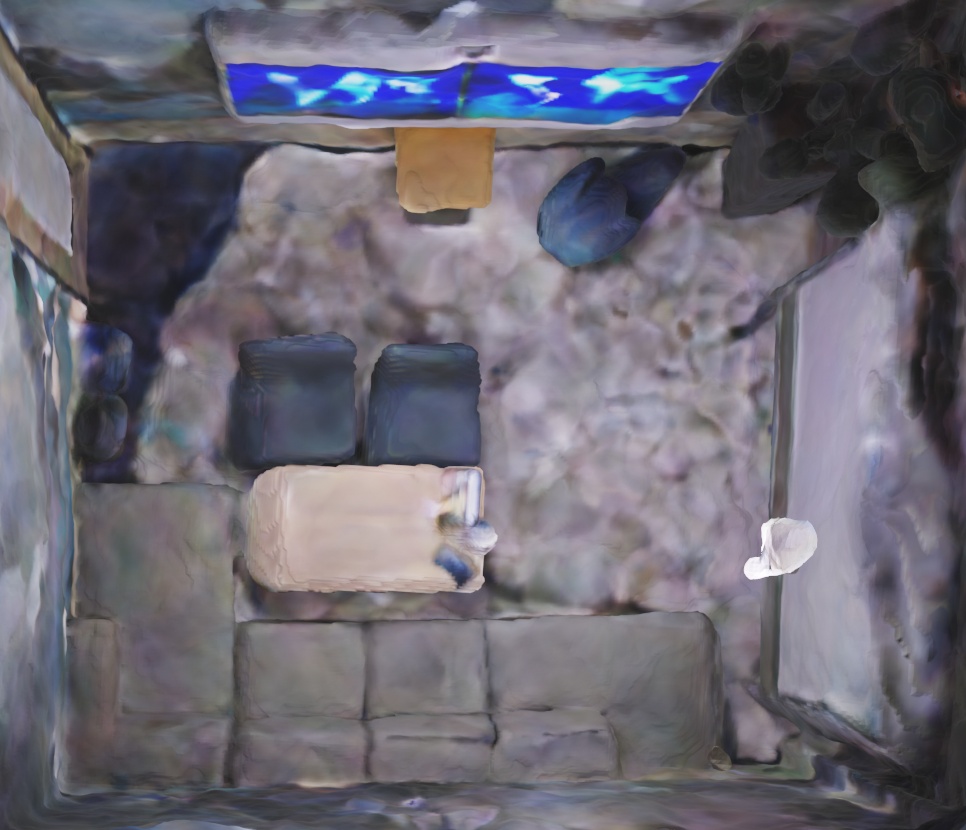}} &
    \makecell{\includegraphics[width=\swc\linewidth,height=\sz\linewidth]{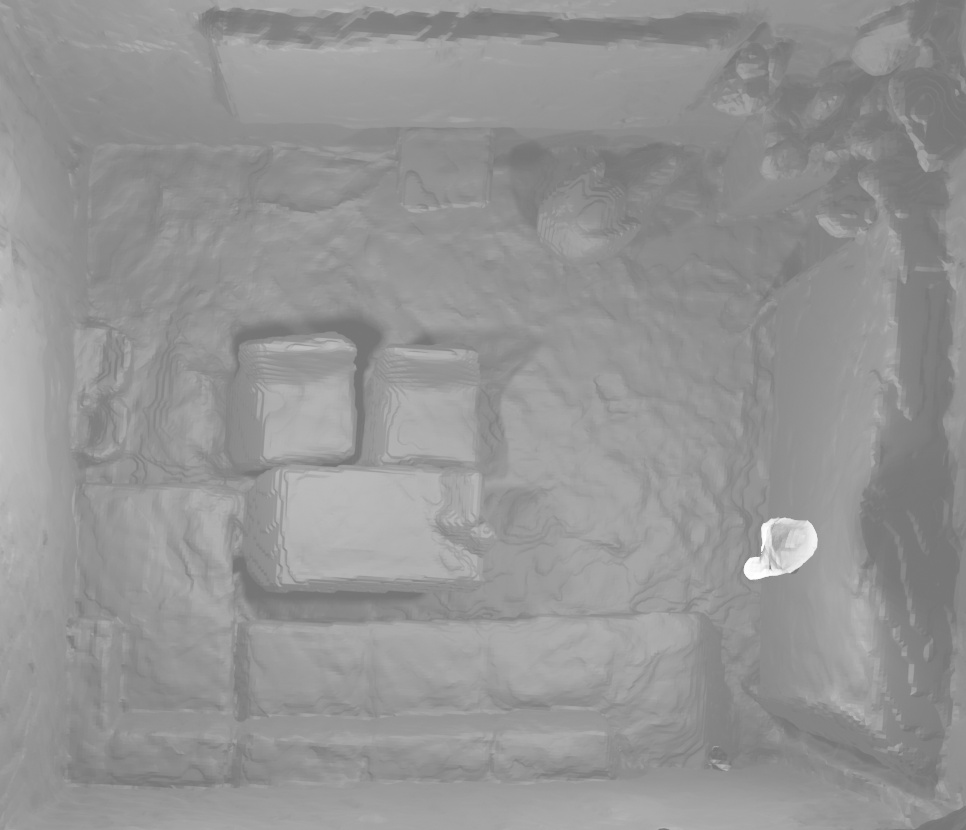}} &
    \makecell{\includegraphics[width=\swd\linewidth,height=\sz\linewidth]{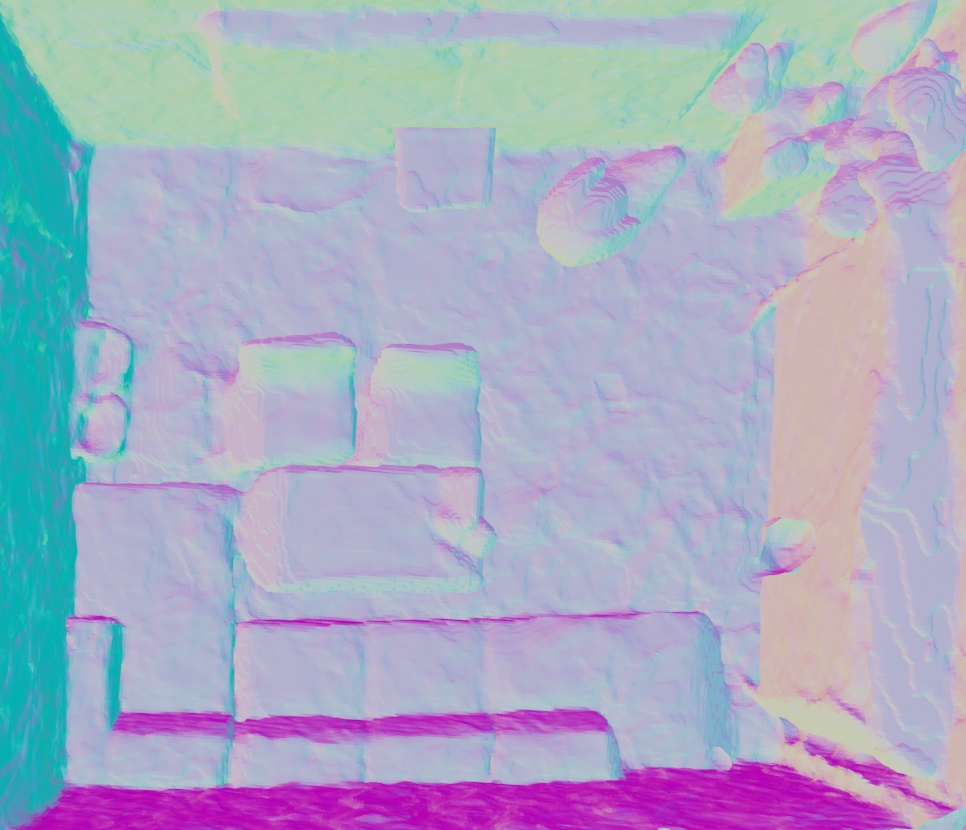}} \\
    
    \makecell{\rotatebox{90}{\ours{} (Ours)}} &
    \makecell{\includegraphics[width=\swa\linewidth,height=\sz\linewidth]{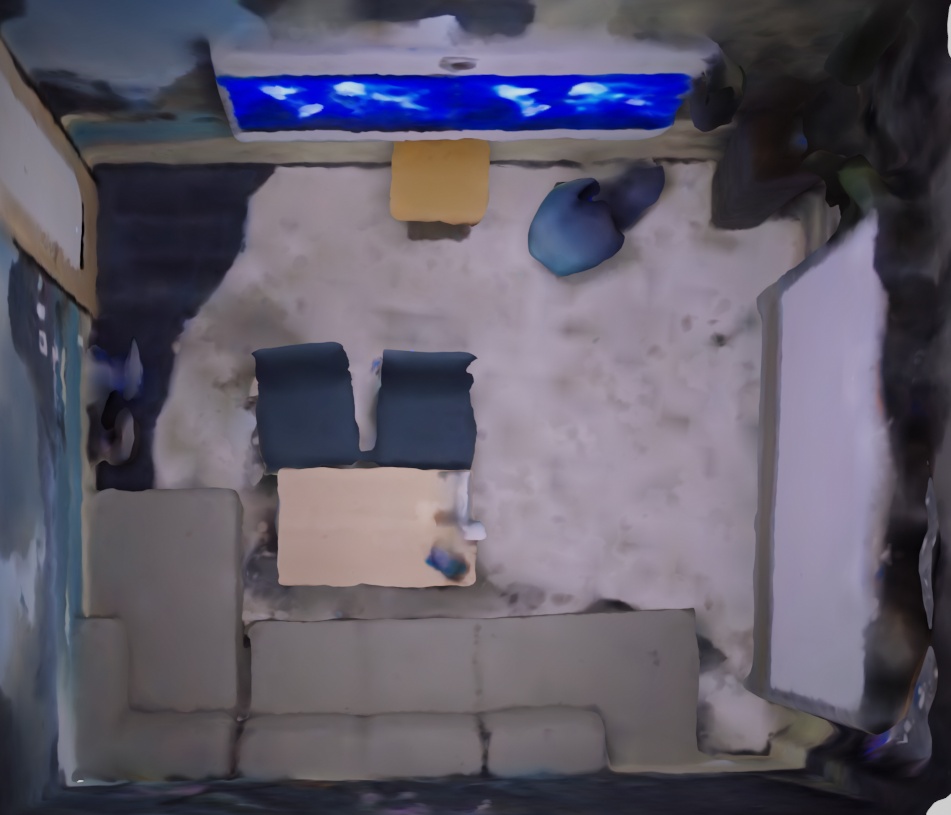}} & 
    \makecell{\includegraphics[width=\swb\linewidth,height=\sz\linewidth]{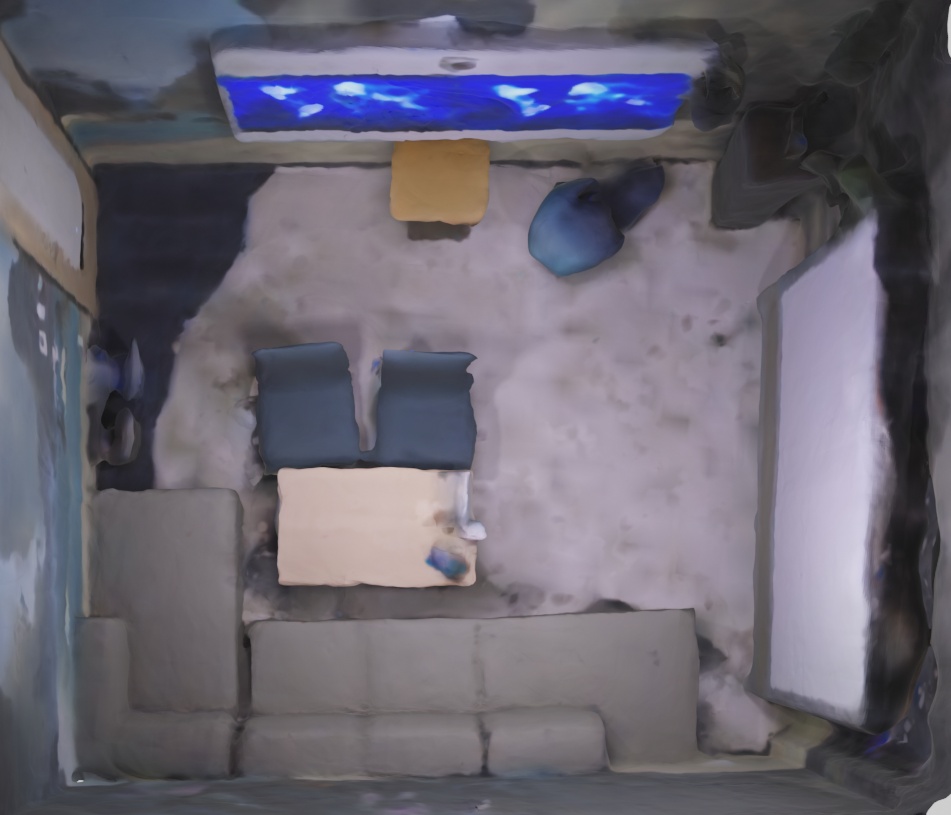}} &
    \makecell{\includegraphics[width=\swc\linewidth,height=\sz\linewidth]{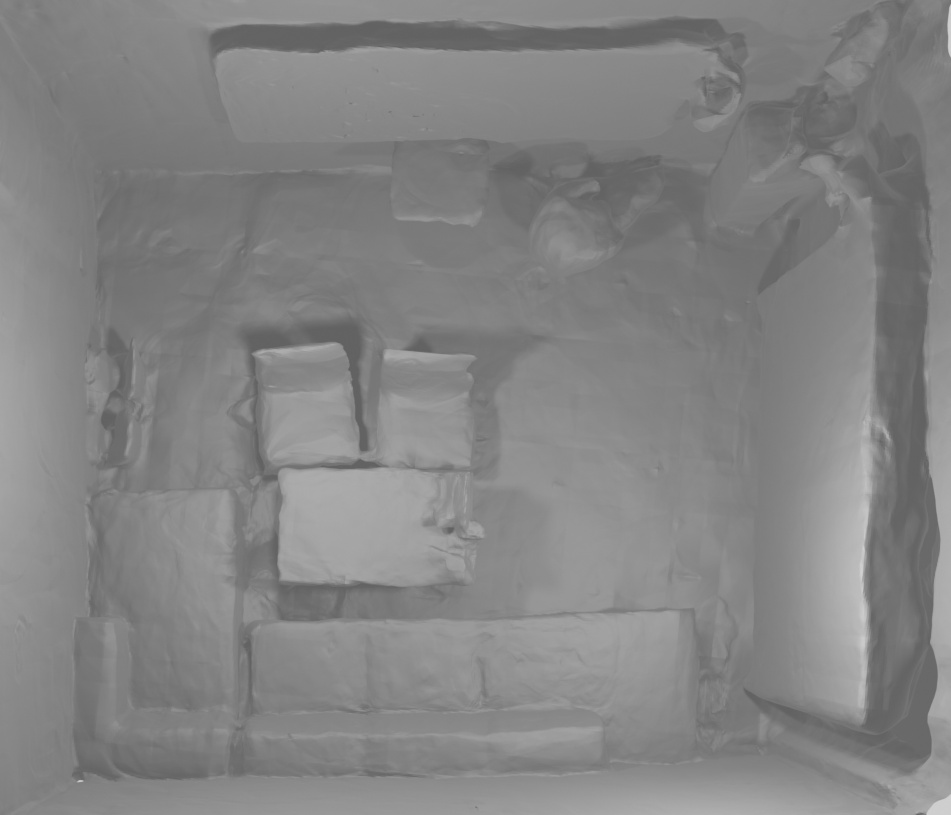}} &
    \makecell{\includegraphics[width=\swd\linewidth,height=\sz\linewidth]{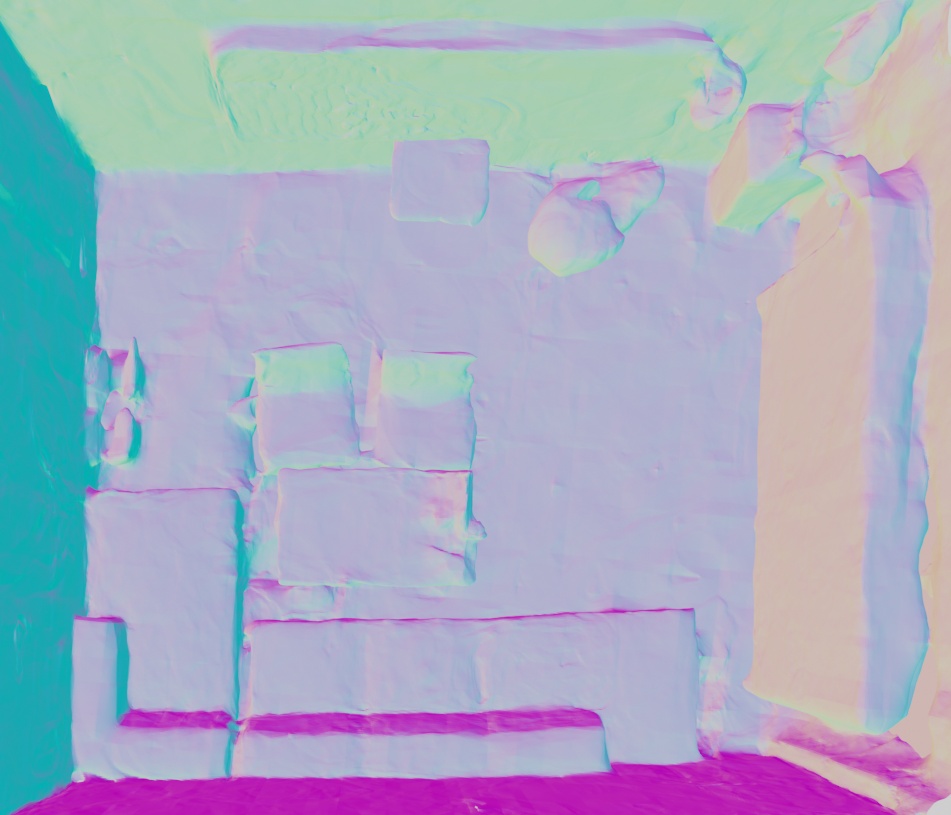}} \\
    
    \makecell{\rotatebox{90}{iMAP$^*$~\cite{imap}}} &
    \makecell{\includegraphics[width=\swa\linewidth,height=\sz\linewidth]{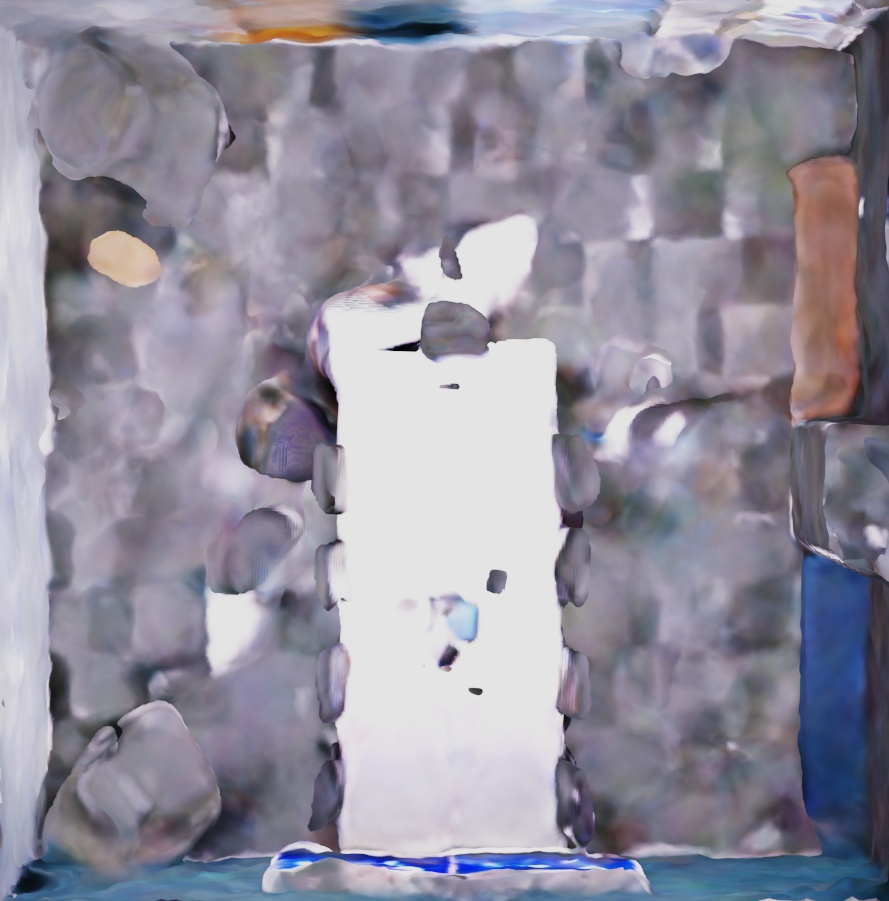}} & 
    \makecell{\includegraphics[width=\swb\linewidth,height=\sz\linewidth]{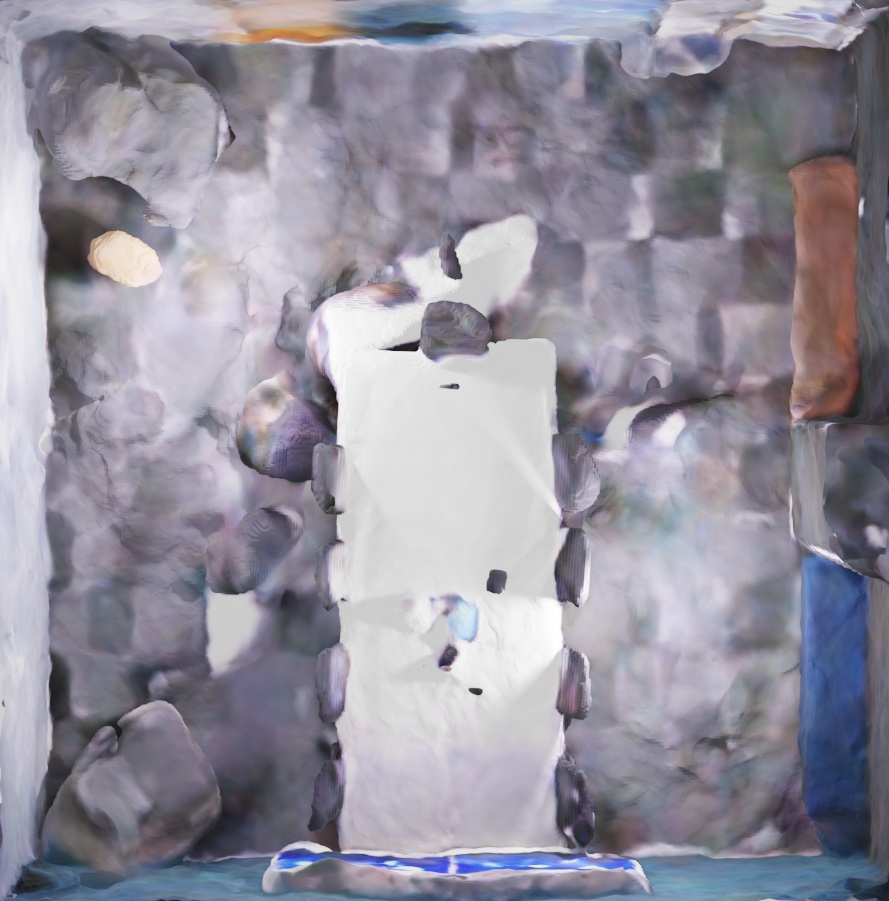}} &
    \makecell{\includegraphics[width=\swc\linewidth,height=\sz\linewidth]{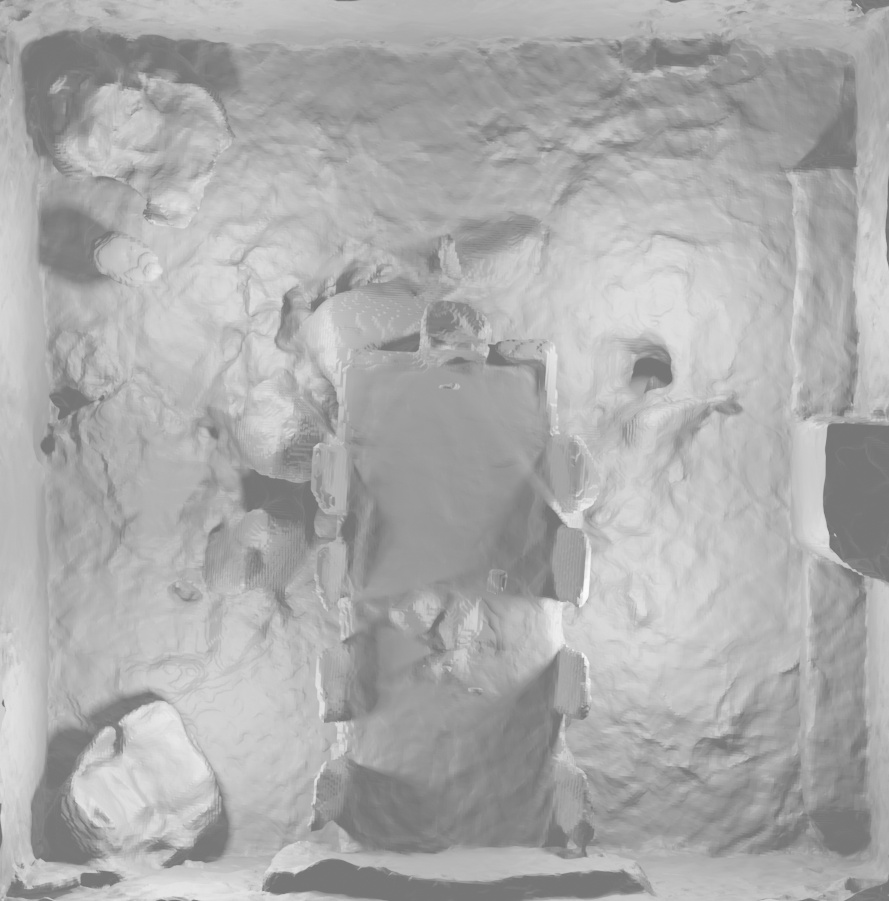}} &
    \makecell{\includegraphics[width=\swd\linewidth,height=\sz\linewidth]{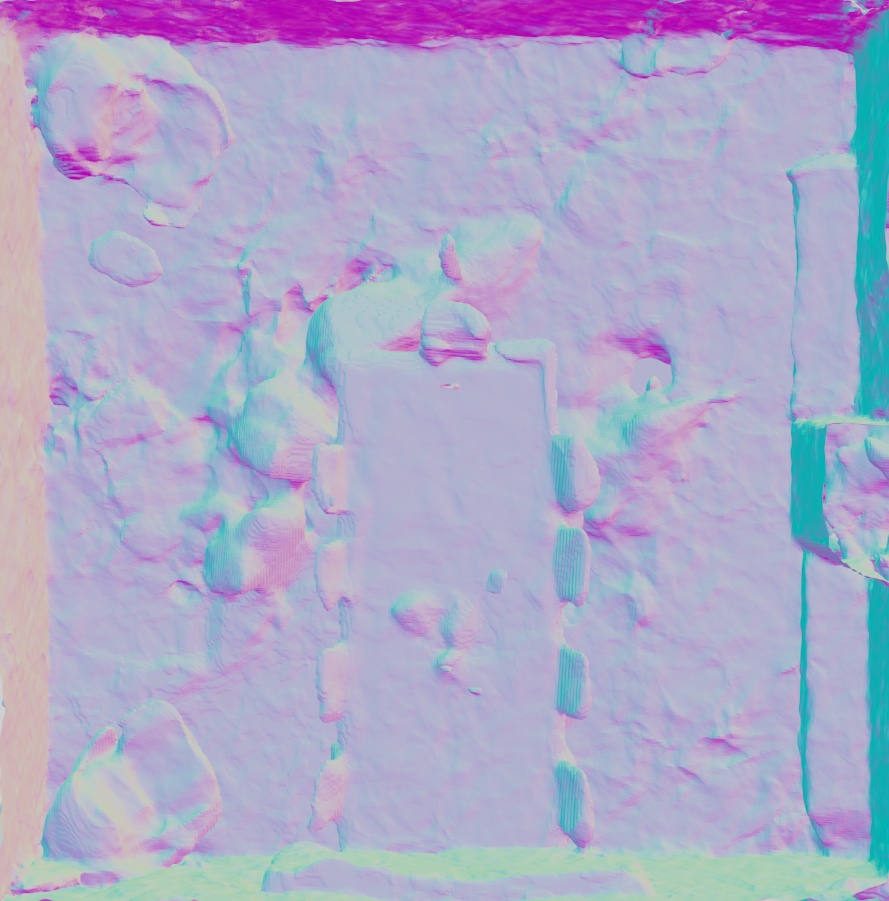}} \\
    
    \makecell{\rotatebox{90}{\ours{} (Ours)}} &
    \makecell{\includegraphics[width=\swa\linewidth,height=\sz\linewidth]{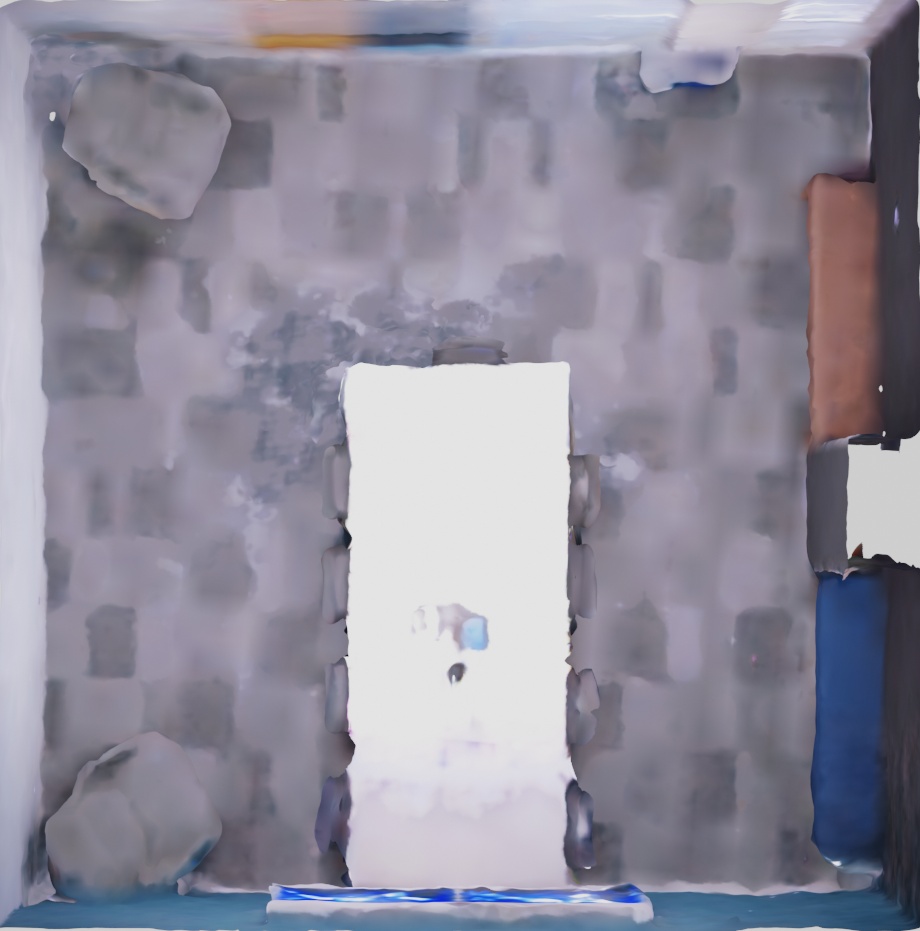}} & 
    \makecell{\includegraphics[width=\swb\linewidth,height=\sz\linewidth]{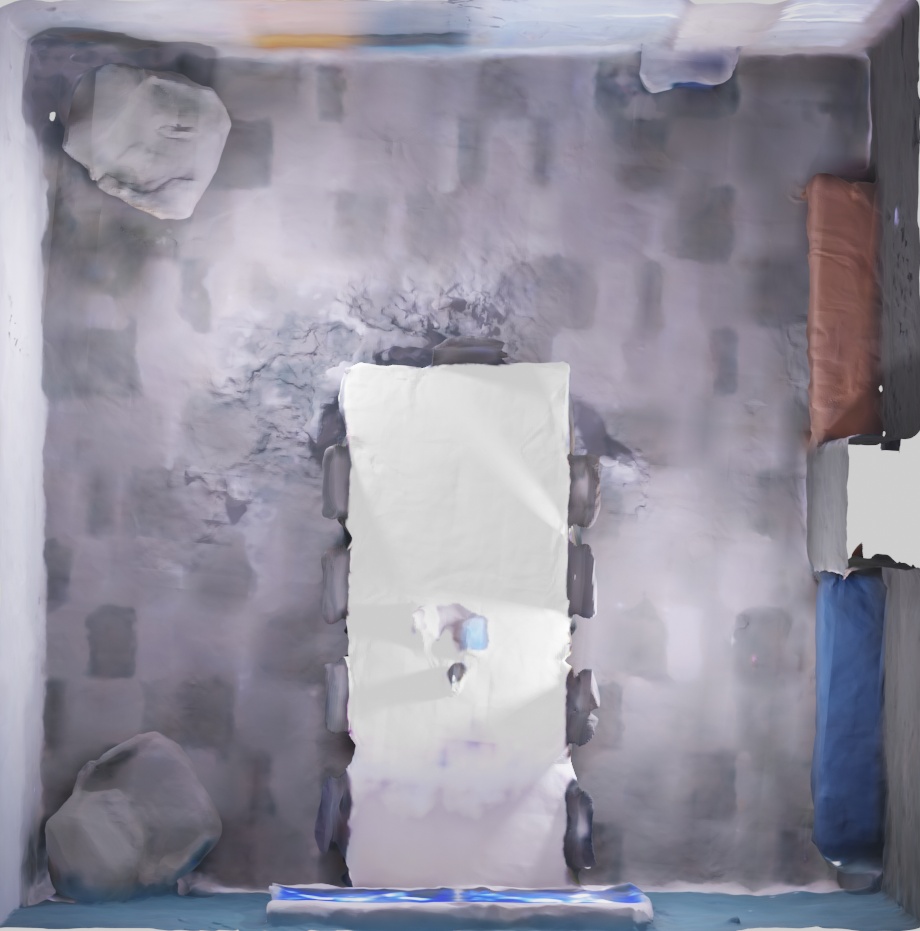}} &
    \makecell{\includegraphics[width=\swc\linewidth,height=\sz\linewidth]{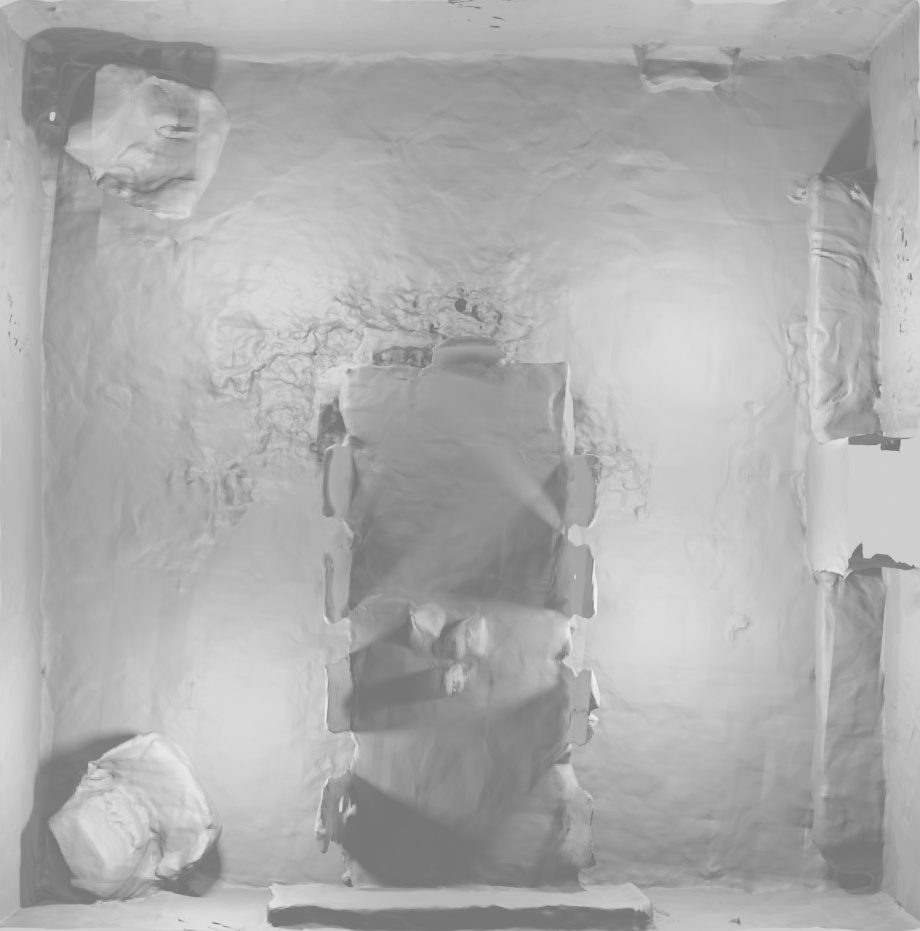}} &
    \makecell{\includegraphics[width=\swd\linewidth,height=\sz\linewidth]{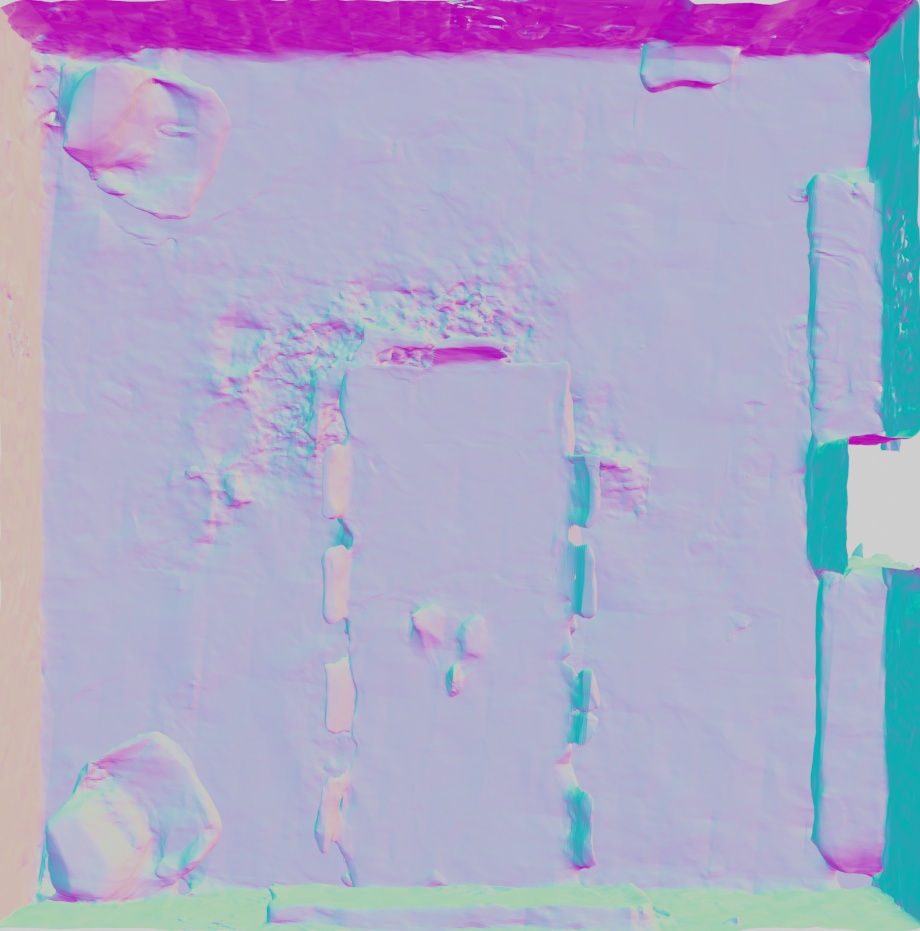}} \\
    
  \end{tabular} 
  \caption{\textbf{More Results on Replica Dataset~\cite{replica19arxiv}}. %
  We visualize the final reconstruction on two scenes including office-0 (top two rows) and office-4 (bottom two rows).
  To better show the differences, we use different rendering settings. As can be visualized, our NICE-SLAM produces high-quality geometry and colors.
  }
  \label{fig:more_replica}
\end{figure*}

In addition, to better highlight the performance differences, we provide additional visualizations using different rendering settings in Fig.~\ref{fig:more_replica}.

\begin{table*}[!ht]
  \centering
  \footnotesize
  \setlength{\tabcolsep}{0.36em}
    \begin{tabular}{clcccccccccccccccccc}
      \toprule
         & & \multicolumn{1}{c}{\makecell{\tt{room-0}}} & \multicolumn{1}{c}{\makecell{\tt{room-1}}} &  \multicolumn{1}{c}{\makecell{\tt{room-2}}} & \multicolumn{1}{c}{\makecell{\tt{office-0}}} & \multicolumn{1}{c}{\makecell{\tt{office-1}}} & \multicolumn{1}{c}{\makecell{\tt{office-2}}}& \multicolumn{1}{c}{\makecell{\tt{office-3}}} & \multicolumn{1}{c}{\makecell{\tt{office-4}}} & Avg. \\
        \midrule

          \multirow{4}{*}{\makecell{\textbf{TSDF-Fusion} \\ Res. = 512\\(536.87MB)}}
      
        & {\bf Depth L1} [cm] $\downarrow$ 
          & 6.38 & 5.33 & 6.84 & 4.74 & 4.62 & 11.32 & 9.89 & 6.49 & 6.95\\
          & {\bf Acc. } [cm] $\downarrow$ 
          & 1.87 & 2.48 & 1.69 & 1.14 & 0.96 & 1.63 & 2.08 & 1.74 & 1.70\\
          & {\bf Comp. } [cm] $\downarrow$ 
          & 3.60 & 3.20 & 2.85 & 1.72 & 2.31 & 3.66 & 3.69 & 3.91 & 3.12\\
          & {\bf Comp. Ratio} [$<$ 5cm \%] $\uparrow$ 
          & 88.33 & 89.82 & 90.38 & 93.55 & 90.35 & 86.74 & 85.35 & 86.31 & 88.85\\
          
          \midrule
          
          \multirow{4}{*}{\makecell{\textbf{TSDF-Fusion} \\ Res. = 256\\(67.10MB)}}
      
        & {\bf Depth L1} [cm] $\downarrow$ 
          & 6.69 & 5.47 & 7.47 & 4.97 & 5.28 & 12.30 & 11.17 & 7.20 & 7.57 \\
          & {\bf Acc. } [cm] $\downarrow$ 
          & 1.76 & 2.11 & 1.59 & 1.15 & 0.97 & 1.56 & 1.98 & 1.66 & \textbf{1.60} \\
          & {\bf Comp. } [cm] $\downarrow$ 
          & 3.85 & 3.36 & 3.33 & 1.93 & 2.68 & 4.17 & 4.22 & 4.37 & 3.49 \\
          & {\bf Comp. Ratio} [$<$ 5cm \%] $\uparrow$ 
          & 86.29 & 88.44 & 86.63 & 91.73 & 87.88 & 82.95 & 81.31 & 83.38 & 86.08\\
          
        \midrule
        \multirow{4}{*}{\makecell{\textbf{iMAP$^*$~\cite{imap}}\\(1.04MB)}}   
         & {\bf Depth L1} [cm] $\downarrow$ 
          & 5.70 & 4.93 & 6.94 & 6.43 & 7.41 & 14.23 & 8.68 & 6.80 & 7.64\\
          & {\bf Acc.} [cm] $\downarrow$
          & 5.66 & 5.31 & 5.64 & 7.39 & 11.89 & 8.12 & 5.62 & 5.98 & 6.95\\
          & {\bf Comp.} [cm] $\downarrow$
          & 5.20 & 5.16 & 5.04 & 4.35 & 5.00 & 6.33 & 5.47 & 6.10 & 5.33\\
          & {\bf Comp. Ratio} [$<$ 5cm \%] $\uparrow$
          & 67.67 & 66.41 & 69.27 & 71.97 & 71.58 & 58.31 & 65.95 & 61.64 & 66.60 \\
      \midrule
      \multirow{4}{*}{\makecell{\textbf{DI-Fusion~\cite{huang2021di}}\\(3.78MB)}}  
        & {\bf Depth L1} [cm] $\downarrow$ 
          & 6.66 & 96.82 & 36.09 & 7.36 & 5.05 & 13.73 & 11.41 & 9.55 & 23.33\\
          & {\bf Acc.} [cm] $\downarrow$
          &  1.79 & 49.00 & 26.17 & 70.56 & 1.42 & 2.11 & 2.11 & 2.02  & 19.40\\
          & {\bf Comp.} [cm] $\downarrow$
          & 3.57 & 39.40 & 17.35 & 3.58 & 2.20 & 4.83 & 4.71 & 5.84  & 10.19\\
          & {\bf Comp. Ratio} [$<$ 5cm \%] $\uparrow$
          &  87.77 & 32.01 & 45.61 & 87.17 & 91.85 & 80.13 & 78.94 & 80.21 & 72.96 \\
     \midrule
     \multirow{4}{*}{{\makecell{\textbf{\ours{}}\\(12.02MB)}}} 
        & {\bf Depth L1} [cm] $\downarrow$ 
          & 2.11 & 1.68 & 2.90 & 1.83 & 2.46 & 8.92 & 5.93 & 2.38 &  \textbf{3.53} \\
          & {\bf Acc. } [cm] $\downarrow$ 
          & 2.73 & 2.58 & 2.65 & 2.26 & 2.50 & 3.82 & 3.50 & 2.77 & 2.85 \\
          & {\bf Comp. } [cm] $\downarrow$ 
          & 2.87 & 2.47 & 3.00 & 2.02 & 2.36 & 3.57 & 3.83 & 3.84 & \textbf{3.00} \\
          & {\bf Comp. Ratio} [$<$ 5cm \%] $\uparrow$ 
          & 90.93 & 92.80 & 89.07 & 94.93 & 92.61 & 85.20 & 82.98 & 86.14 & \textbf{89.33}\\

      \bottomrule
    \end{tabular}%
    \caption{\textbf{Reconstruction Results for the Replica Dataset (Average of 5 runs).}}
    \label{tab:replica_per_scene}
\end{table*}

\begin{table*}[!tb]
  \centering
  \footnotesize
  \setlength{\tabcolsep}{0.36em}
    \begin{tabular}{clccccccccc}
      \toprule
         & & \tt{room-0} & \tt{room-1} & \tt{room-2}  & \tt{office-0} &  \tt{office-1} & \tt{office-2} & \tt{office-3} & \tt{office-4} & Avg. \\
        \midrule

      \multirow{3}{*}{\makecell{\textbf{TSDF-Fusion} \\ Res. = 512\\(536.87MB)}}  & {\bf Acc.} [cm]  
      & 5.20 & 2.83 & 1.60 & 1.66 & 1.06 & 2.29 & 2.50 & 2.18 & 2.42\\
      & {\bf Comp.} [cm] 
      & 5.05 & 4.60 & 4.50 & 1.06 & 9.57 & 5.84 & 4.16 & 4.30 & 4.89 \\
      & {\bf Comp. Ratio} [$<$ 5cm \%] 
      & 75.07 & 79.03 & 86.01 & 80.19 & 77.80 & 80.69 & 82.29 & 83.00 & 80.51 \\
      \midrule
      \multirow{3}{*}{\makecell{\textbf{TSDF-Fusion} \\ Res. = 256\\(67.10MB)}}  & {\bf Acc.} [cm]  
      & 4.17 & 2.69 & 1.49 & 1.65 & 1.09 & 2.24 & 2.37 & 2.16 & \textbf{2.23} \\
      & {\bf Comp.} [cm] 
      & 5.65 & 4.85 & 5.04 & 10.88 & 9.85 & 6.94 & 4.93 & 4.95 & 6.64 \\
      & {\bf Comp. Ratio} [$<$ 5cm \%] 
      & 72.99 & 77.19 & 83.10 & 78.52 & 76.43 & 75.66 & 76.74 & 79.01 & 77.46 \\
      \midrule
      \multirow{3}{*}{\makecell{\textbf{iMAP~\cite{imap}}\\(1.04MB)}}    
          & {\bf Acc.} [cm] $\downarrow$
          & 3.58 & 3.69 & 4.68 & 5.87 & 3.71 & 4.81 & 4.27 & 4.83 & 4.43 \\
          & {\bf Comp.} [cm] $\downarrow$
          & 5.06 & 4.87 & 5.51 & 6.11 & 5.26  & 5.65 & 5.45 & 6.59 &  5.56\\
          & {\bf Comp. Ratio} [$<$ 5cm \%] $\uparrow$
          & 83.91 & 83.45 & 75.53 & 77.71& 79.64 & 77.22& 77.34 & 77.63 & 79.06\\
      \midrule
      \multirow{3}{*}{\makecell{\textbf{iMAP$^*$~\cite{imap}}\\(1.04MB)}}  
          & {\bf Acc.} [cm] $\downarrow$
          & 4.07 & 3.86 &  5.17 & 5.40 & 4.04 & 5.23 & 4.30 & 4.98 & 4.63 \\
          & {\bf Comp.} [cm] $\downarrow$
          & 4.73 & 4.32 &  5.53 & 4.95 & 5.27  & 5.40 & 4.94  & 5.08 & 5.03\\
          & {\bf Comp. Ratio} [$<$ 5cm \%] $\uparrow$
          & 79.12 & 76.21 & 69.19 & 77.47 & 76.70 & 70.53 & 73.51 & 71.81 & 74.32 \\
     \midrule
      \multirow{3}{*}{\makecell{\textbf{DI-Fusion~\cite{huang2021di}}\\(3.78MB)}}  & {\bf Acc.} [cm] 
      & 2.02 & 277.51 & 24.94 & 61.73 & 1.75 & 2.63 & 2.97 & 2.11 & 46.96\\
      & {\bf Comp.} [cm] 
      & 3.90 & 82.87 & 20.16 & 12.08 & 8.76 & 6.89 & 5.70 & 5.96 & 18.29 \\
      & {\bf Comp. Ratio} [$<$ 5cm \%] 
      & 86.58 & 24.77 & 41.50 & 74.20 & 79.22 & 73.36 & 70.24 & 78.26 & 66.02 \\

      \midrule
      \multirow{3}{*}{\makecell{\textbf{\ours{}}\\(12.02MB)}} & {\bf Acc.} [cm]
 
          & 2.97 & 3.23 & 3.46 & 5.47 & 3.33 & 4.40 & 3.55 & 2.87 & 3.66 \\
          & {\bf Comp. } [cm] $\downarrow$ 
          & 3.30 & 3.07 & 3.75 & 4.54 & 3.83  & 3.90 & 4.49 & 3.91 & \textbf{3.85}\\
          & {\bf Comp. Ratio} [$<$ 5cm \%] $\uparrow$ 
          & 89.51 & 86.01 & 81.14 & 85.27 & 88.01 & 82.61 & 79.49 & 85.33 & \textbf{84.67}\\
      
      \bottomrule
    \end{tabular}%
    \caption{\textbf{Reconstruction Results for the Replica Dataset (Best in 5 runs).}
    The numbers for iMAP are directly taken from~\cite{imap}.
    }
    \label{tab:replica_prev_per_scene}
\end{table*}

\subsection{More Results on ScanNet~\cite{dai2017scannet}}
\begin{figure*}[h]
  \centering
  \footnotesize
  \setlength{\tabcolsep}{1.5pt}
  \newcommand{\sz}{0.17}
  \begin{tabular}{lcccccc}
    & {\tt Frame 1000} & {\tt Frame 2000} & {\tt Frame 3000} & {\tt Frame 4000} & {\tt Frame 5000}\\
    \rotatebox{90}{\hspace{16pt}iMAP$^*$~\cite{imap}} &  
    \includegraphics[width=\sz\linewidth]{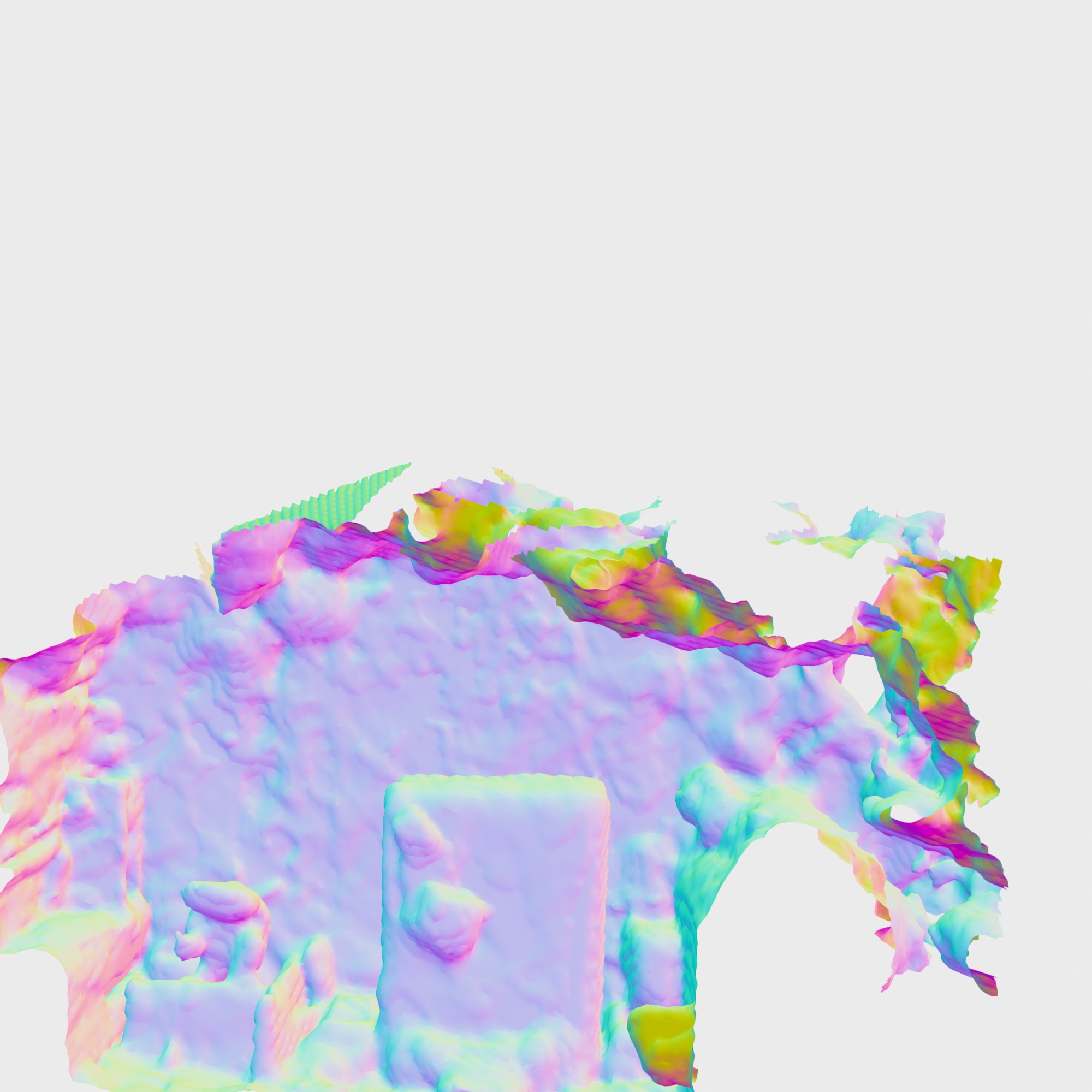} &
    \includegraphics[width=\sz\linewidth]{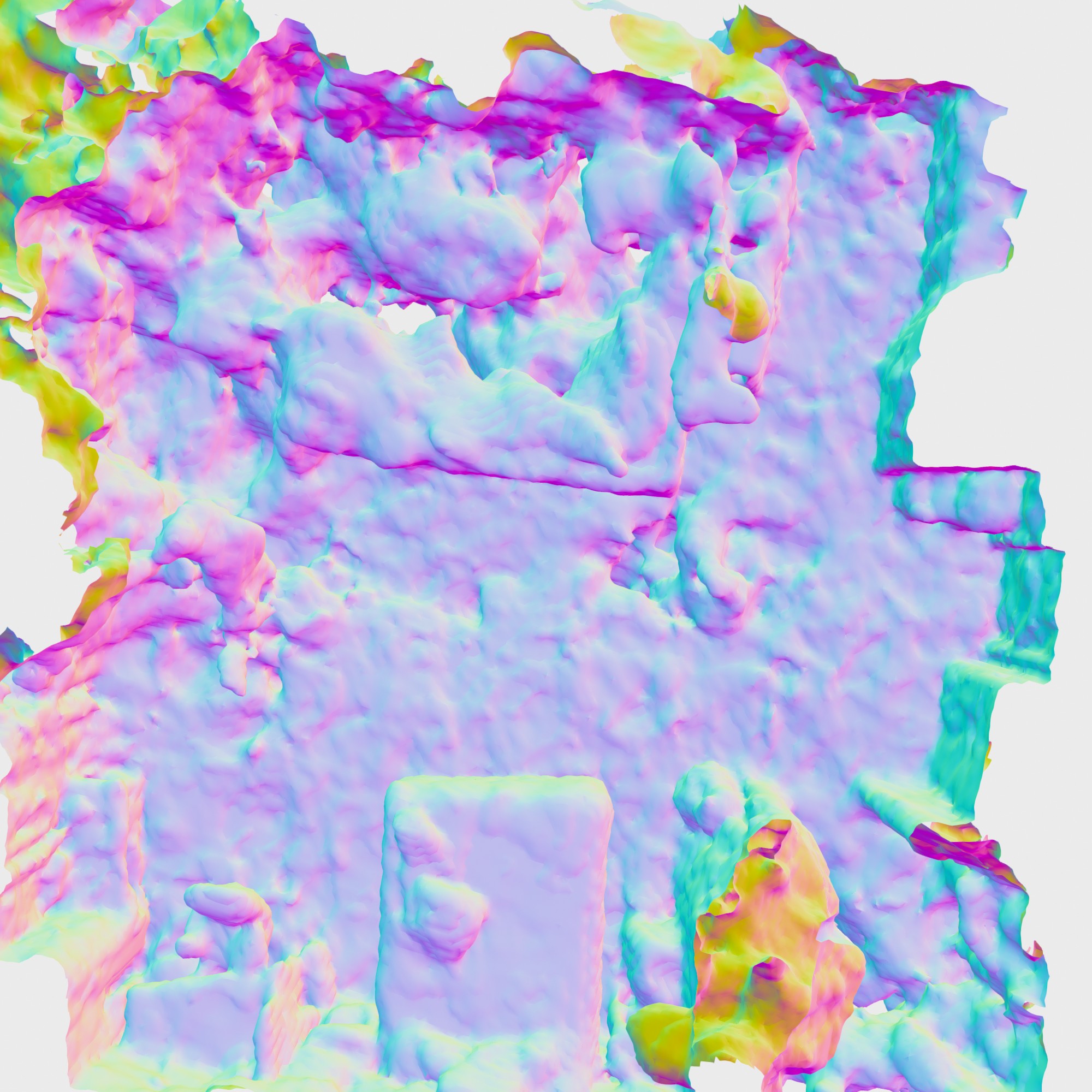} &
    \includegraphics[width=\sz\linewidth]{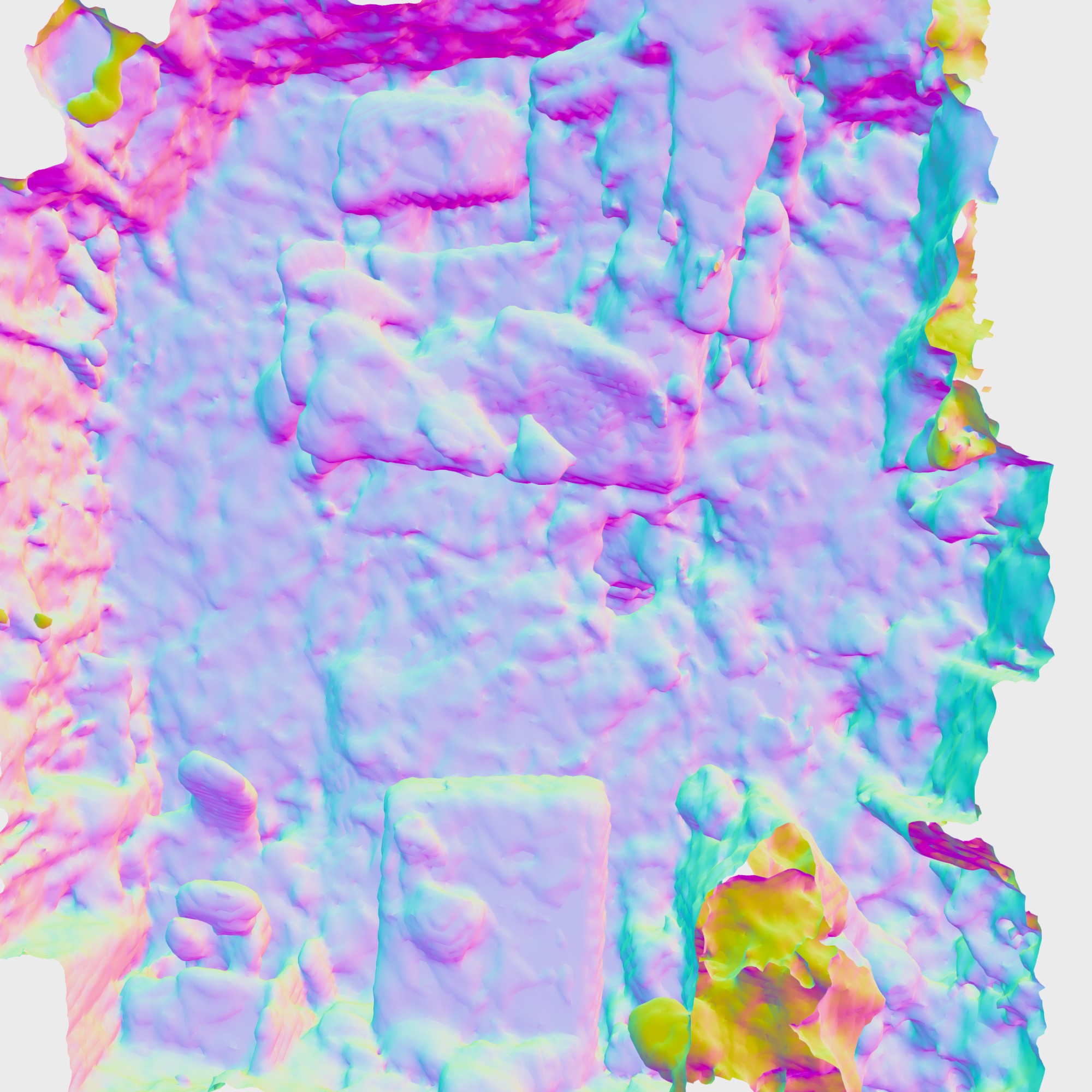} &
    \includegraphics[width=\sz\linewidth]{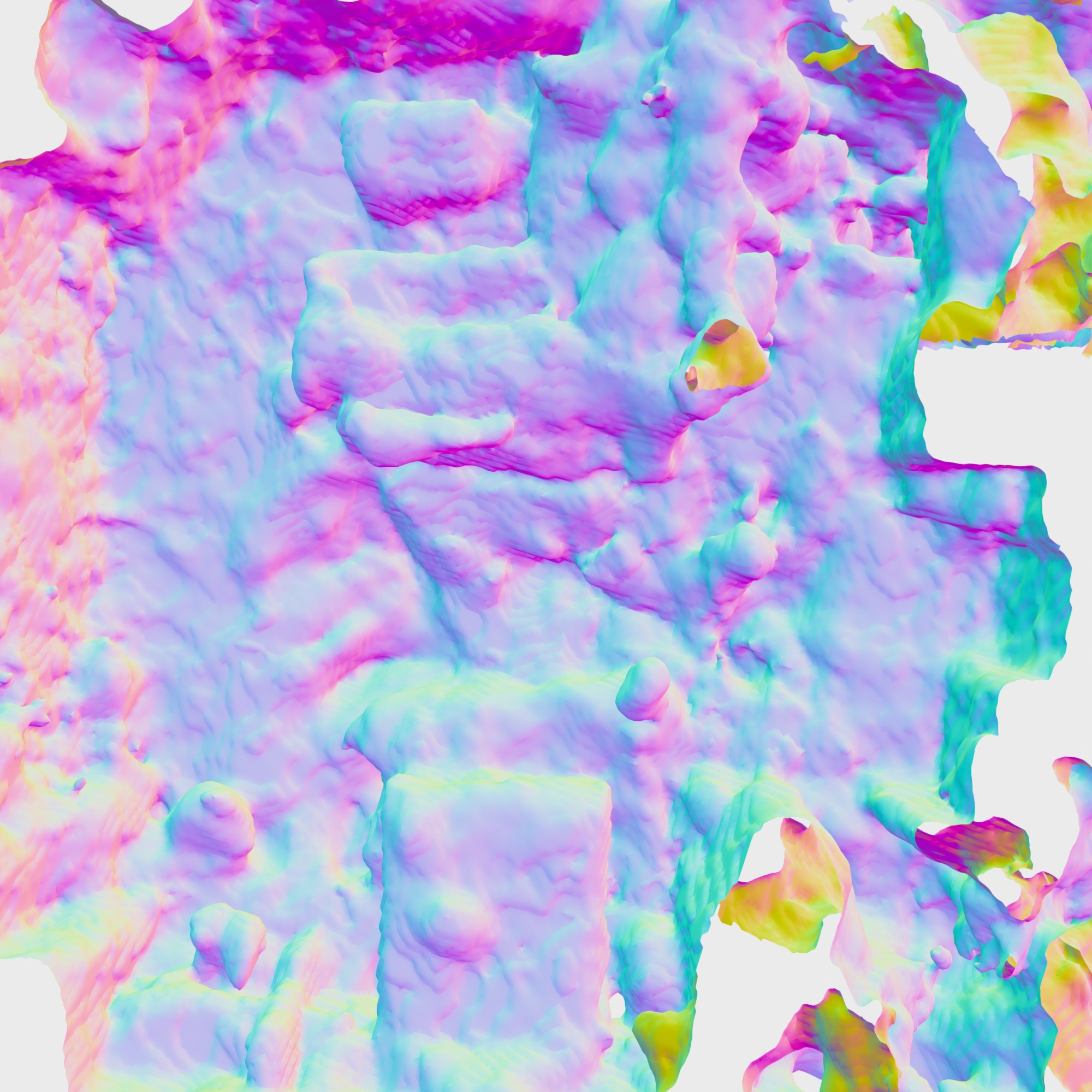} &
    \includegraphics[width=\sz\linewidth]{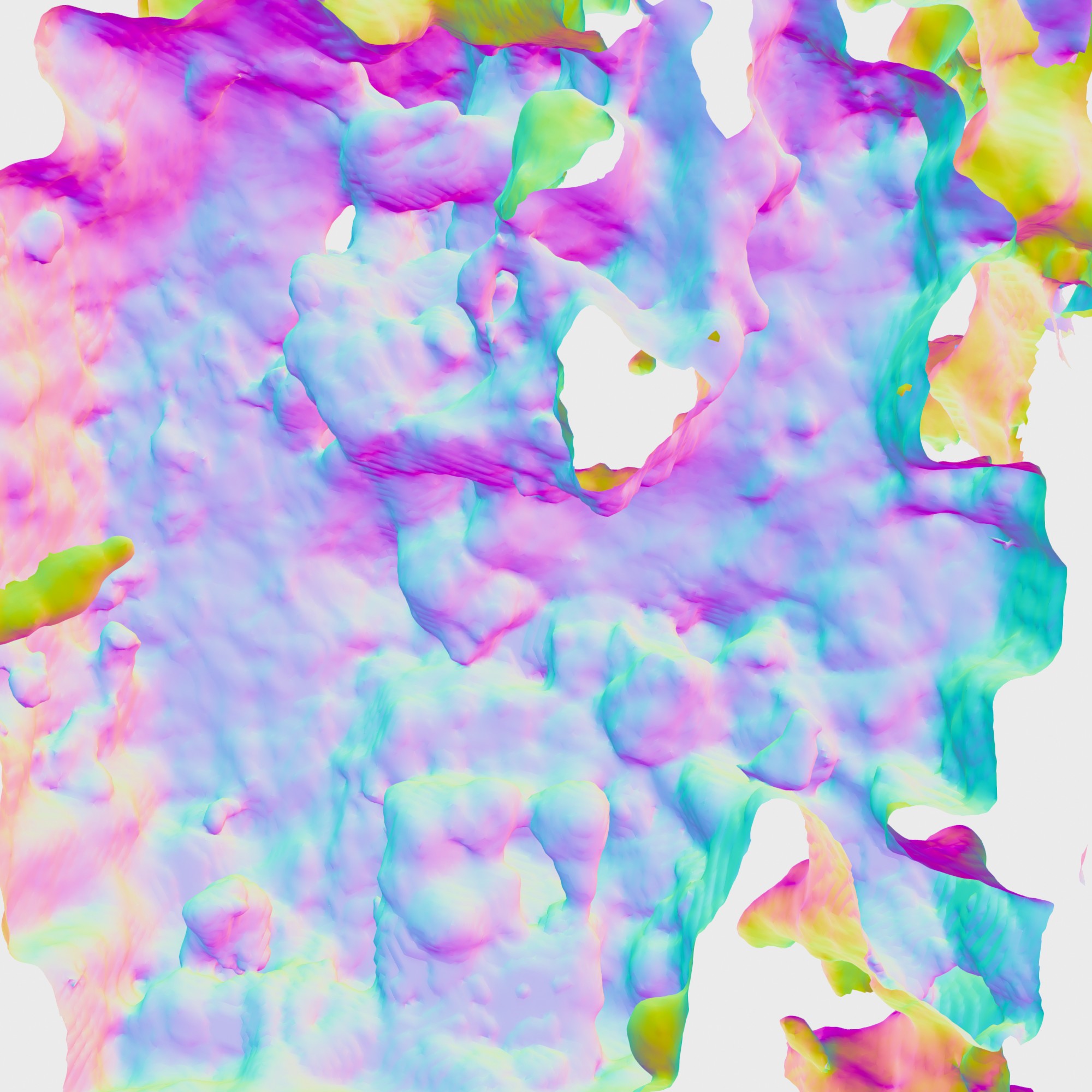} \\
    \rotatebox{90}{\hspace{0pt}\ours{} (Ours)} &
    \includegraphics[width=\sz\linewidth]{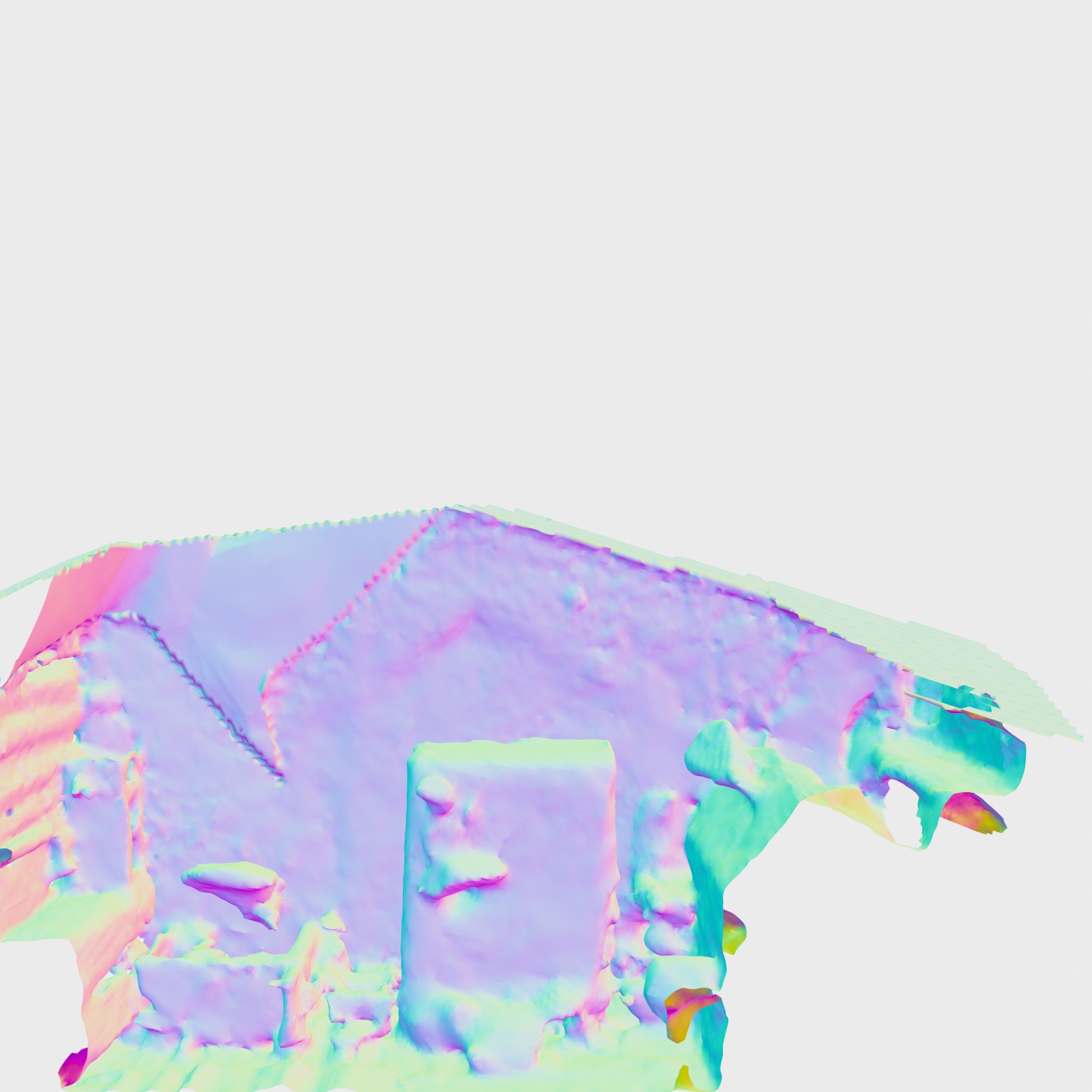} &
    \includegraphics[width=\sz\linewidth]{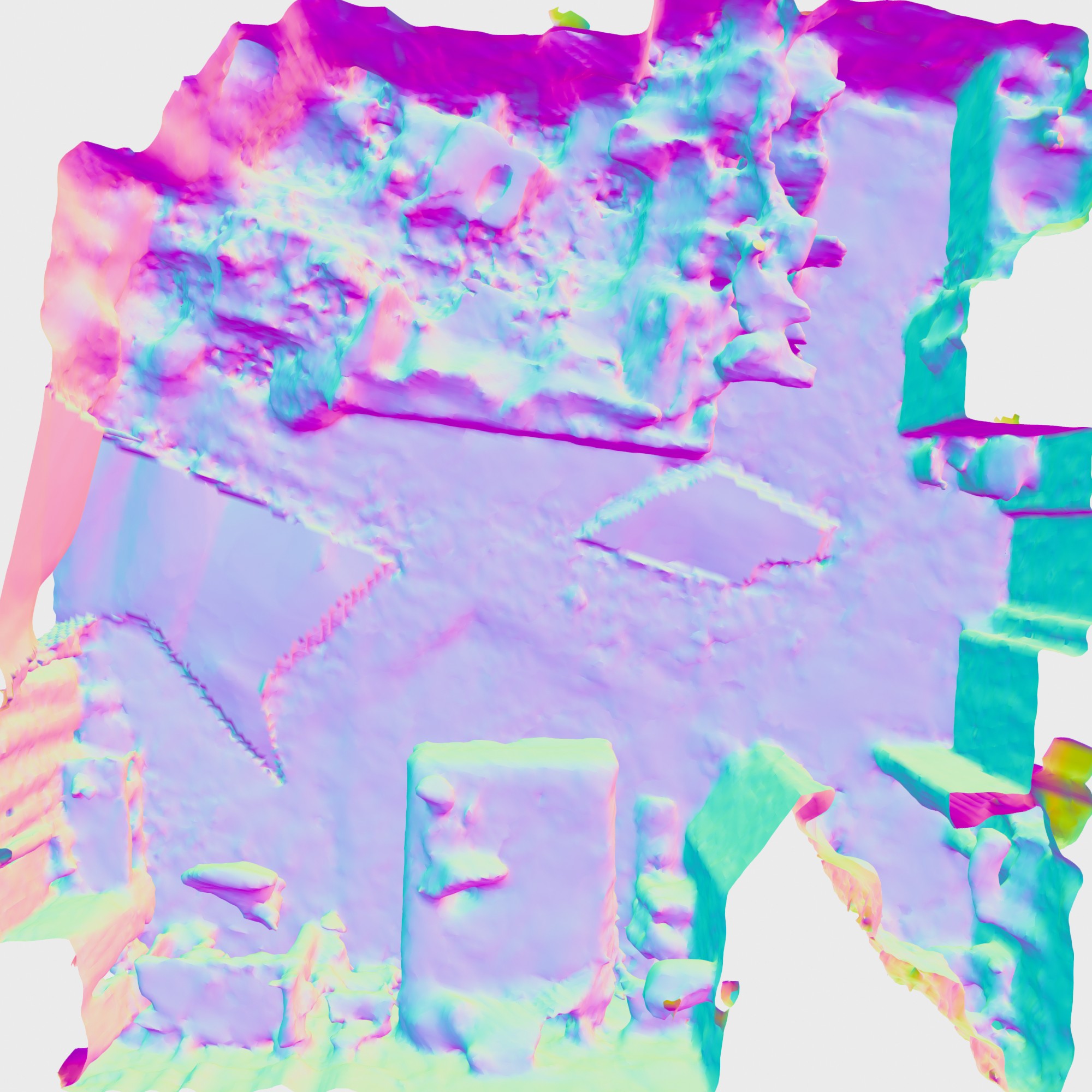} &
    \includegraphics[width=\sz\linewidth]{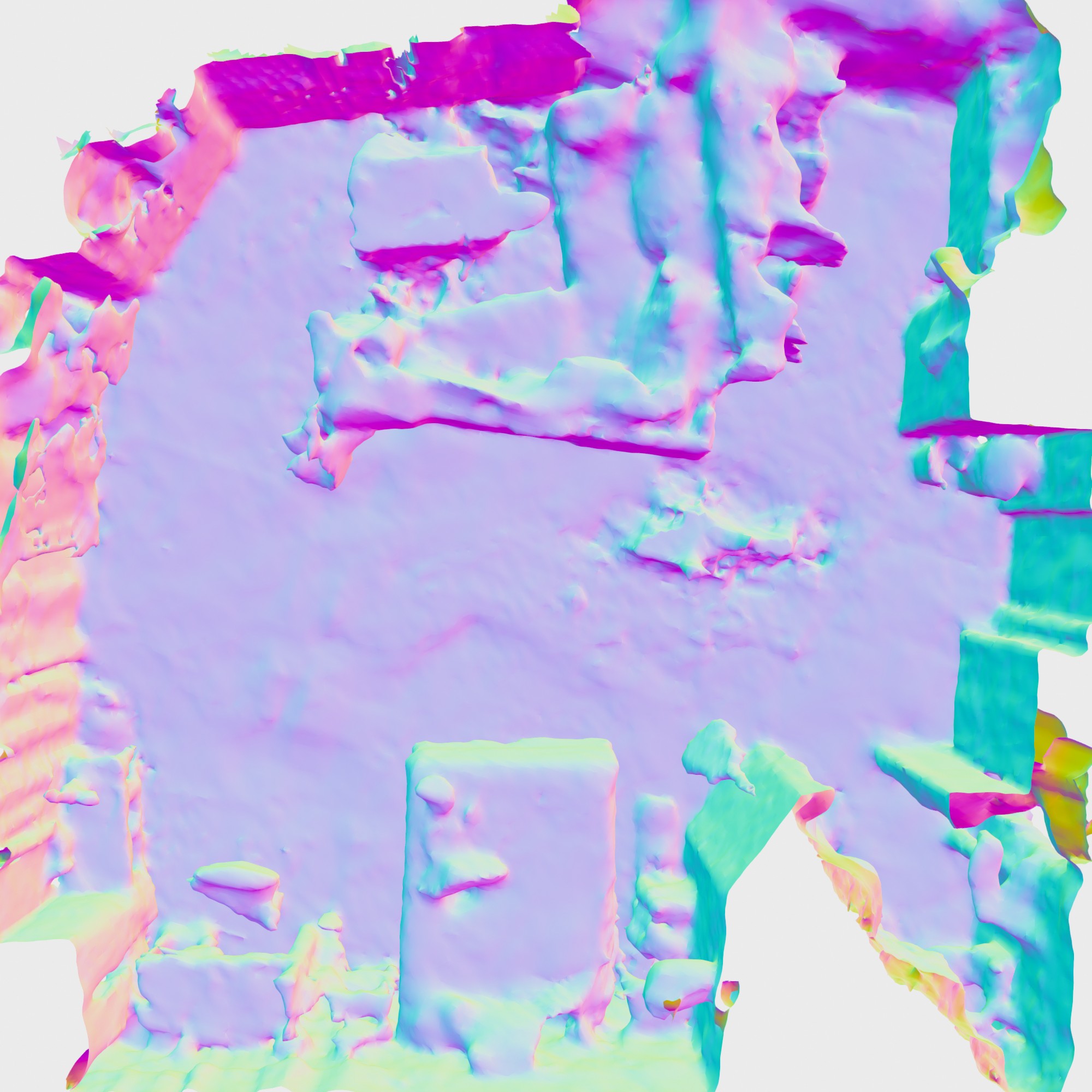} &
    \includegraphics[width=\sz\linewidth]{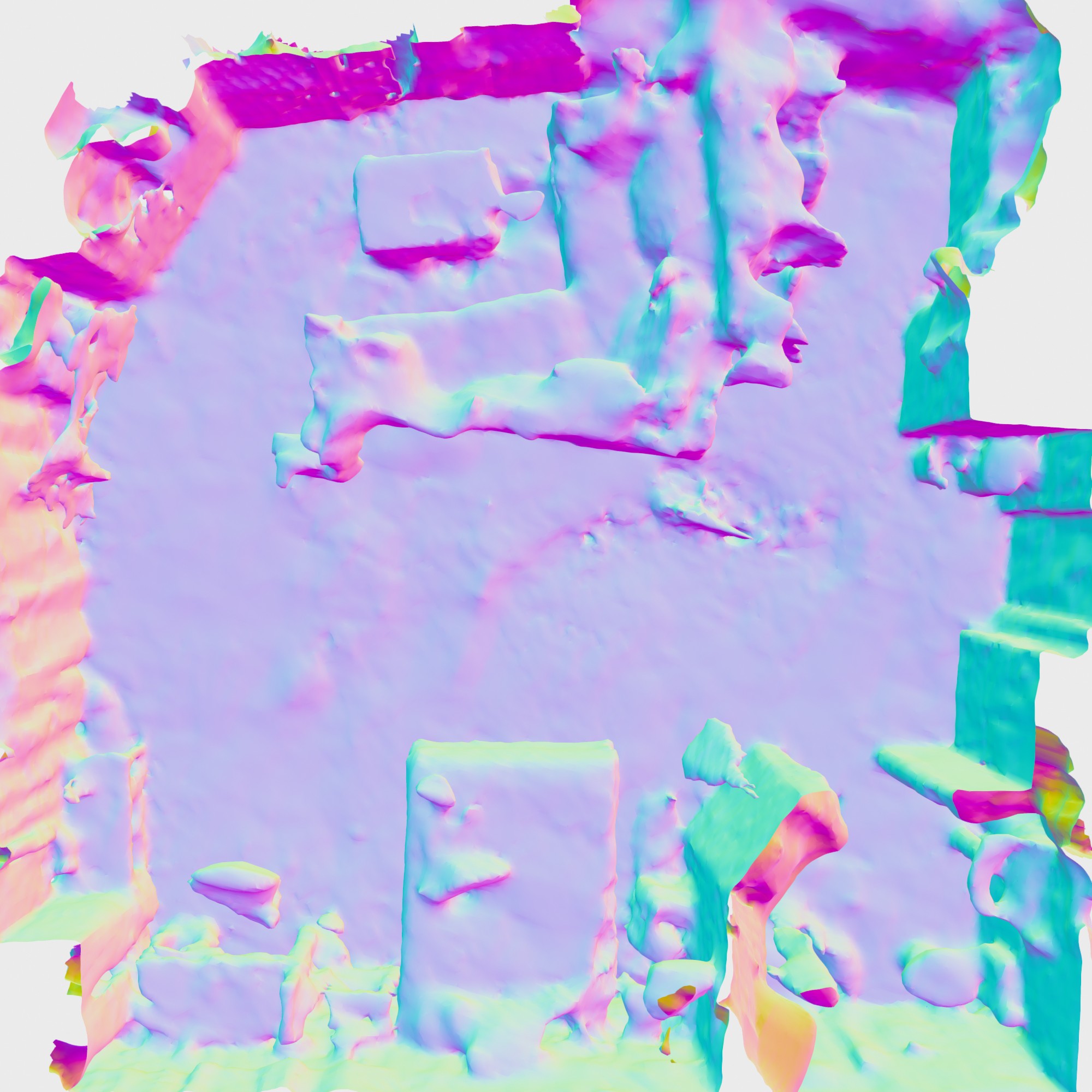} &
    \includegraphics[width=\sz\linewidth]{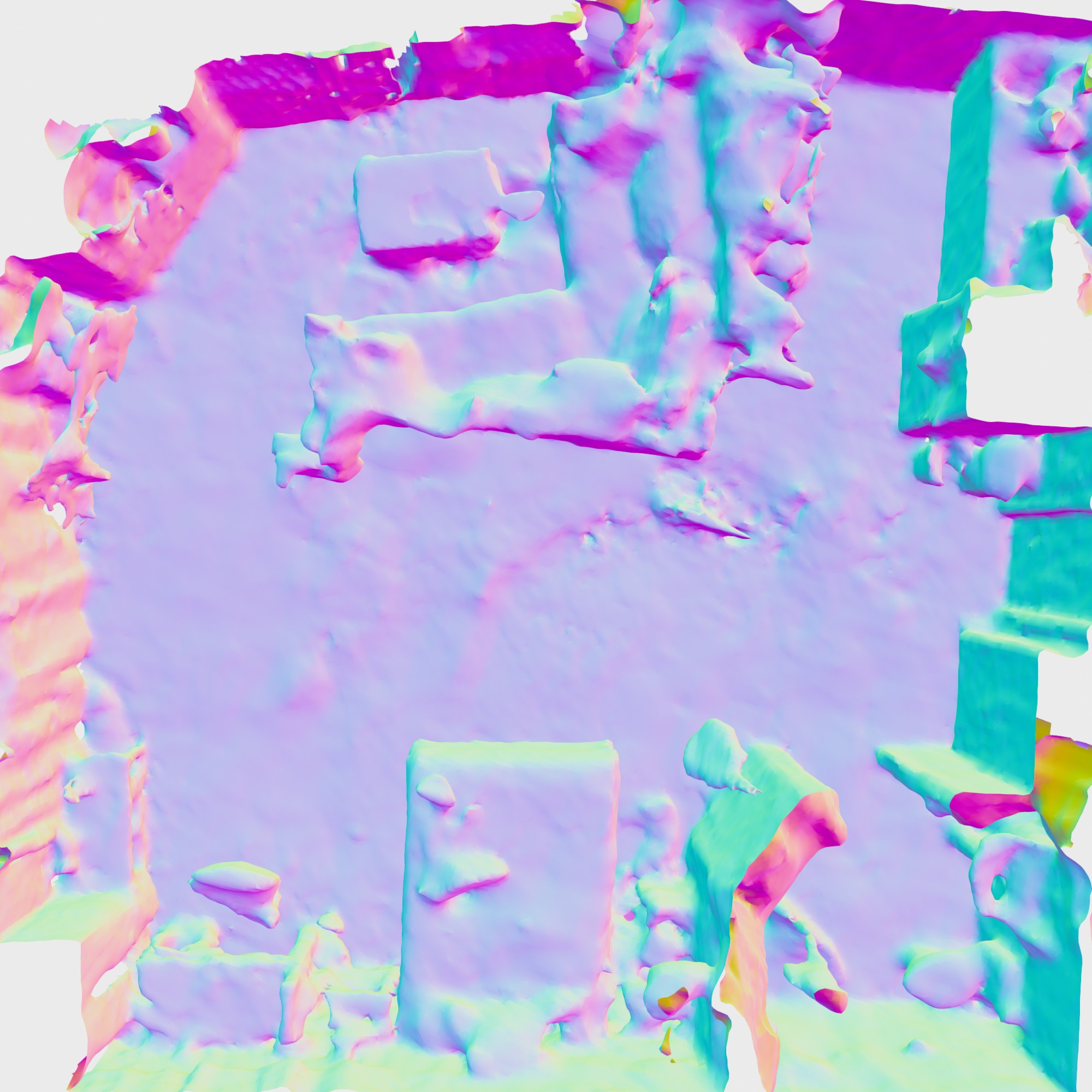} \\
  \end{tabular} 
  \caption{\textbf{3D Reconstruction Process on ScanNet~\cite{dai2017scannet}.} Due to our local map updates the resulting geometry is temporally more stable and often less noisy compared to iMAP$^*$~\cite{imap}. 
 }
  \label{fig:scannet}
\end{figure*}

We show the 3D reconstruction process of iMAP$^*$ and \ours{} on ScanNet {\tt{scene0000}} in \figref{fig:scannet}.

\end{document}